\RequirePackage{fix-cm}

\documentclass[twocolumn, natbib]{svjour3}

\smartqed  % flush right qed marks, e.g. at end of proof

\usepackage[ngerman,english]{babel}
\usepackage{graphicx}
\usepackage[caption=false]{subfig}
\usepackage{amsmath}
\usepackage{amssymb}
\usepackage{dsfont}
\usepackage[misc]{ifsym}
\usepackage{tikz}
\usepackage{booktabs}
\usepackage{multirow}
\usepackage{silence}
\journalname{PFG – Journal of Photogrammetry, Remote Sensing and Geoinformation Science}

\begin{document}

\title{Semi-Supervised Hyperspectral Image Classification with Edge-Aware
Superpixel Label Propagation and Adaptive Pseudo-Labeling}

\author{
        Yunfei Qiu\textsuperscript{1}
        \and
        Qiqiong Ma\textsuperscript{1,2,3}
        \and
        Tianhua Lv\textsuperscript{1,2,3}
        \and
        Li Fang\textsuperscript{2,3}
        \and
        Shudong Zhou\textsuperscript{2,3}
        \and
        Wei Yao\textsuperscript{2,3}
}

% \author{
%         Yunfei Qiu
%         \and
%         Qiqiong Ma
%         \and
%         Tianhua Lv
%         \and
%         Li Fang\thanks{Corresponding author:lfang@iue.ac.cn}
%         \and
%         Shudong Zhou
%         \and
%         Wei Yao
% }

\authorrunning{Yunfei Qiu et al.}

\institute{
\Letter\ Li Fang\\
% \email{lfang@iue.ac.cn}
\phantom{\Letter}\enspace lfang@iue.ac.cn
\\[12pt]
\textsuperscript{1}\,
School of Software Engineering,
Liaoning Technical University,
Huludao, Liaoning, China
\\[6pt]
\textsuperscript{2}\,
Spatial Intelligence and Urban Computing,
Institute of Urban Environment,
Chinese Academy of Sciences,
Xiamen, China
\\[6pt]
\textsuperscript{3}\,
The State Key Lab for Ecological Security of Regions and Cities, 
Institute of Urban Environment, 
Chinese Academy of Sciences, 
Xiamen, China
\\[6pt]
% \textsuperscript{*}\,
% Corresponding author: \email{lfang@iue.ac.cn}
}

% \institute{
%         Yunfei Qiu \at
%             School of Software Engineering,
%             Liaoning Technical University,
%             Huludao, Liaoning, China
%         \and
%         Qiqiong Ma \at
%         School of Software Engineering,
%         Liaoning Technical University,
%         Huludao, Liaoning, China
%         \and
%         Tianhua Lv \at
%         School of Software Engineering,
%         Liaoning Technical University,
%         Huludao, Liaoning, China
%         \and
%         Li Fang, Shudong Zhou, Wei Yao \at
%         Spatial Intelligence and Urban Computing,
%         Institute of Urban Environment,
%         Chinese Academy of Sciences,
%         Xiamen, China
% }
% \institute{F. Author \at
%               first address \\
%               Tel.: +123-45-678910\\
%               Fax: +123-45-678910\\
%               \email{fauthor@example.com}           %  \\
% %             \emph{Present address:} of F. Author  %  if needed
%            \and
%            S. Author \at
%               second address
% }
\date{Received: date / Accepted: date}

\maketitle

\begin{abstract}

Hyperspectral image (HSI) contain rich spectral information. However, owing to factors such as limited spatial resolution, complex spatial structures, and differences in spectral variation, unlabeled samples exhibit significant differences in classification difficulty. Therefore, samples located at land-cover boundaries and class-intermixed regions, as well as samples with large intra-class spectral variations, are more likely to generate erroneous supervisory signals. Existing methods mainly focus on enhancing spectral–spatial representations during the feature extraction stage, while paying relatively little attention to the reliability of the class information of unlabeled samples. To address the above issues, this paper proposes a semi-supervised hyperspectral image classification method that combines edge-aware superpixel label propagation with adaptive pseudo-label optimization. First, an Edge-Aware Superpixel Label Propagation (EASLP) module is proposed, which adjusts neighborhood similarity relationships by introducing boundary constraints, thereby suppressing cross-class label propagation. Second, a Dynamic Reliability-Enhanced Pseudo-Labeling (DREPL) strategy is proposed, which enhances pseudo-label consistency by fusing historical prediction information with current prediction results and adaptively divides unlabeled samples into easy, ambiguous, and hard categories according to their prediction states to alleviate the accumulation of erroneous pseudo-labels. Experimental evaluations on five benchmark datasets show that the proposed method improves the overall accuracy by 7.00\%. Even when 50\% fewer labeled samples are used, our method outperforms competing methods by 9.93\%.

% Although semi-supervised hyperspectral image (HSI) classification has achieved significant progress in feature extraction and classification performance, existing methods still face challenges such as boundary label diffusion and pseudo-label instability during the utilization of unlabeled samples. To address these issues, this paper proposes a semi-supervised hyperspectral classification framework that integrates edge-aware modeling with a dynamic reliable pseudo-labeling strategy. To mitigate cross-boundary label propagation caused by superpixel segmentation, we introduce an Edge-Aware Superpixel Label Propagation (EASLP) module, incorporating boundary constraints into similarity matrix modeling and adaptively adjusting neighbor contributions according to boundary strength. For pseudo-label optimization, a Dynamic Reliability-Enhanced Pseudo-Labeling (DREPL) strategy is adopted, which introduces historical predictions and fuses them with current predictions in a dynamic weighted manner to enhance prediction consistency. By jointly evaluating confidence and the Count-Gap metric, unlabeled samples are dynamically divided into easy, ambiguous, and hard categories, followed by class-specific optimization strategies, leading to improved pseudo-label stability from both spatial and temporal perspectives. Experimental evaluations on five benchmark datasets show that the proposed method improves the overall accuracy by 7.00\%. Even when 50\% fewer labeled samples are used, our method outperforms competing methods by 9.93\%.

\end{abstract}

\section{Introduction}
\label{intro}
Hyperspectral image (HSI) provides more continuous and detailed spectral band information, enabling it to capture subtle spectral differences in terrestrial materials. As a crucial application, hyperspectral image classification (HSIC) can assign precise land cover labels to each observed pixel, and thus has gained increasing attention \citep{ref1,ref2,ref6}. However, it is confronted with two major challenges: on one hand, the complex high-dimensional structure of HSI complicates feature extraction, which can be addressed through deep learning techniques \citep{ref8}; on the other hand, the labor-intensive and time-consuming nature of manual annotation makes it difficult to obtain large labeled datasets \citep{ref9,ref44,ref45}, presenting a significant challenge for deep learning models that typically require substantial training data, thereby driving the need for techniques that enhance classification performance under constrained supervision. 

In the early stages of HSIC research, fully supervised convolutional neural network (CNN)-based methods achieved remarkable performance \citep{ref10,ref41,ref11,ref13}. Representative approaches explored multi-scale feature learning \citep{ref12,ref14}. A2S2K \citep{ref15} improves local spectral–spatial feature extraction through adaptive spectral–spatial kernels. SSTN \citep{ref16} further enhances global dependency modeling with self-attention. The strong dependence on large-scale labeled samples limits the applicability of these methods in hyperspectral scenarios with scarce annotations.

% In the early stages of HSIC research, fully supervised methods based on convolutional neural networks (CNNs) achieved significant progress \citep{ref10,ref41,ref11,ref13}. \citep{ref12} effectively captured multi-granularity structural information of land cover by integrating features across different scales through dense connectivity mechanisms. \citep{ref14} further constructed a hierarchical feature structure, enabling the modeling of structural characteristics of land cover across multiple scales and levels. Subsequent approaches, such as A2S2K \citep{ref15}, introduced an adaptive spectral–spatial kernel mechanism, allowing convolutional kernels to dynamically adjust their receptive fields according to input features; SSTN \citep{ref16} designed a spectral–spatial transformation module, leveraging self-attention to model long-range dependencies. Nevertheless, these methods remain fundamentally constrained by their heavy reliance on large-scale high-quality annotations, making them impractical for hyperspectral scenarios with prohibitive labeling costs.

Unsupervised learning methods have been explored to reduce the reliance on labeled samples \citep{ref17}. Traditional approaches mainly learn representations through reconstruction or clustering \citep{ref19,ref21}. Due to the high-dimensional spectral characteristics of HSIs, these objectives may preserve data structures but fail to capture sufficiently discriminative class information.

% Recently, self-supervised methods based on contrastive learning and masked reconstruction have become increasingly popular \citep{ref22,ref39,ref40,ref43}. 
Recently, self-supervised methods based on contrastive learning and masked reconstruction, such as DEMAE \citep{ref23} and RMAE \citep{ref24}, have become increasingly popular \citep{ref39,ref40}. However, the absence of explicit class supervision may limit their ability to learn discriminative features for downstream classification, especially in HSI scenarios with complex spectral variations.
% To alleviate this limitation, research has gradually shifted toward unsupervised learning \citep{ref17}. Autoencoders (AEs) and their variants have been widely employed for hyperspectral feature learning, such as Denoising Autoencoders (DAE) \citep{ref18}, Stacked Sparse Autoencoders (SSAE) \citep{ref19}, and approaches combining autoencoders with random forest classifiers \citep{ref20,ref21}. However, these methods still rely on the structural information of the original data and struggle to fully capture higher-level semantic features. To address this, self-supervised pretraining methods have been proposed \citep{ref22,ref39,ref40,ref43}. For instance, DEMAE \citep{ref23} utilizes masked reconstruction and diffusion denoising auxiliary tasks for feature initialization, while RMAE \citep{ref24} further integrates limited labeled data for joint optimization, demonstrating feature transfer potential in low-label scenarios. Nevertheless, due to the lack of explicit class supervision, the features generated by these methods remain limited in class discriminability, making it challenging to ensure stable and reliable performance in downstream classification tasks.

In this context, semi-supervised learning has progressively emerged as a focal point for HSIC research \citep{ref25, ref26, ref27,ref47}. It combines limited labeled data with abundant unlabeled data, simultaneously leveraging label information from supervised learning and self-learning capabilities from unsupervised methodologies \citep{ref28}. The graph-based semi-supervised learning framework was first proposed by \citet{ref29}, where both labeled and unlabeled data are represented as vertices in a weighted graph and classification is achieved using harmonic functions. CTF-SSCL \citep{ref33} integrates CNN and Transformer architectures, leveraging a semi-supervised contrastive learning mechanism to enhance feature learning under few-shot conditions.

A significant advancement emerged with consistency regularization methods \citep{ref30,ref46}, FixMatch \citep{ref37} performs consistency training by matching high-confidence pseudo-labels generated from weakly augmented samples with strongly augmented samples. These methods generally rely on a fixed confidence threshold for pseudo-label selection. In low-label scenarios, incorrect pseudo-labels may be introduced, leading to error accumulation.
% However, this method relies on the model's confidence to assess the quality of pseudo-labels, which can lead to misclassification of pseudo-labels when supervised information is limited. 
To address this issue, CGMatch \citep{ref51} utilizes dynamic sample exploitation, finely dividing the unlabeled samples to achieve differentiated utilization based on the learning state of the samples. Although this mechanism is effective in natural images, in hyperspectral images, the correlation between spectrally similar pixels makes it difficult to distinguish between categories, and during the early stages of training, the model tends to oscillate between categories, exacerbating the instability of pseudo-labels \citep{ref50}.

Moreover, traditional consistency learning methods struggle to fully exploit the spatial continuity and complex land-cover boundary structures inherent in hyperspectral data. To address this, DSCA \citep{ref49} introduces a superpixel segmentation framework, where a majority voting strategy is used to assign labels to unlabeled pixels, and label propagation is achieved based on the similarity between adjacent regions. However, this approach tends to suffer from label diffusion, particularly in areas of category transitions and blurred boundaries. DMSGer \citep{ref32} further combines superpixels with graph convolutional networks, constructing region-level graph structures and introducing a dynamic pixel-level feature update strategy, which enhances the ability to extract spatial information through multi-scale learning \citep{ref52}. Conventional superpixel propagation generally relies on spectral or feature similarity and lacks explicit boundary constraints. Superpixels or neighborhoods may contain pixels from different classes. This can cause label diffusion across class boundaries and introduce erroneous supervisory information.

The above analysis reveals two key limitations in existing semi-supervised HSIC methods. First, spatial label propagation may cross heterogeneous land-cover boundaries because boundary information is not sufficiently modeled. Second, pseudo-label selection is often based on current predictions and static confidence thresholds, which cannot adequately reflect the dynamic reliability of unlabeled samples. These problems are particularly serious for boundary samples, class-intermixed samples, and samples with large intra-class spectral variations.

Based on the above analysis, we propose a semi-supervised hyperspectral image classification framework that integrates edge-aware spatial priors with dynamic reliable pseudo-labeling, improving pseudo-label quality under extremely limited labeled data. Unlike conventional static selection methods that rely on a single confidence threshold \citep{ref36,ref48}, the proposed framework adaptively adjusts pseudo-label utilization according to the training state, effectively mitigating label diffusion at class boundaries while enhancing the robustness and reliability of hyperspectral image classification. Its efficacy has been extensively validated across multiple benchmark datasets.

In summary, the key contributions of our work are stated as follows.

\begin{enumerate}
  \item Building upon the semi-supervised consistency learning paradigm, we develop a novel hyperspectral image classification method that jointly optimizes the label propagation and pseudo-label generation processes by incorporating edge awareness and dynamic reliability. The approach directly addresses cross-boundary misclassification and pseudo-label instability issues through these core design innovations.
  % We propose a semi-supervised hyperspectral classification framework that integrates edge awareness and dynamic reliability. To address the issues of cross-boundary label mis-propagation and instability in pseudo-label prediction, the processes of label propagation and pseudo-label generation are jointly modeled, thereby enhancing classification stability in complex regions under small-sample conditions.
  % We propose a semi-supervised hyperspectral classification framework that integrates edge-aware modeling with a dynamic reliable pseudo-labeling strategy. The framework leverages superpixel-based spatial structure priors and employs edge-aware constraints on neighborhood similarity, combined with weighted neighbor voting to refine boundary labels. In the pseudo-label optimization stage, it innovatively fuses historical and current predictions to stratify sample reliability and enable differentiated training, thereby providing a high-precision and robust pseudo-label generation pipeline for semi-supervised hyperspectral classification. 
  
  \item We design the Edge-Aware Superpixel Label Propagation (EASLP) algorithm, which is the first to construct a two-stage label propagation mechanism in the field of HSIC. This method directly applies edge strength information during the label propagation stage. Subsequently, edge-guided neighborhood-weighted voting is employed for secondary refinement, effectively preventing label mispropagation in complex boundary regions.
  % We design an Edge-Aware Superpixel Label Propagation (EASLP) algorithm. Within the spatial neighborhood, an edge operator is employed to construct label similarity relationships, enabling the differentiation between interior regions and boundary locations. The edge strength is then dynamically weighted and incorporated into the label correction process, thereby suppressing cross-boundary mis-propagation.
  % We propose an Edge-Aware Superpixel Label Propagation (EASLP) algorithm. The method leverages spatial structure information derived from superpixel segmentation to construct a similarity matrix for label distribution within local neighborhoods, incorporating edge operators. Pseudo-labels are then refined through a neighbor voting mechanism guided by dynamic edge strength weighting. This approach effectively suppresses cross-boundary label errors while preserving label consistency within homogeneous regions.
  
  \item We construct a Dynamic Reliability-Enhanced Pseudo-Labeling (DREPL) generation strategy. This strategy dynamically fuses historical and current predictions to alleviate pseudo-label instability caused by relying solely on current predictions. Based on the resulting confidence fusion, unlabeled samples are divided into three categories to reduce the interference of unreliable pseudo-labels.
  % We construct a Dynamic Reliability-Enhanced Pseudo-Labeling (DREPL) generation strategy. This strategy designs a Dynamic History-Fused Prediction (DHP) module to dynamically fuse historical predictions and current predictions, thereby alleviating the pseudo-label fluctuations caused by relying solely on current predictions. On this basis, the unlabeled samples are divided into three categories using the fused confidence to reduce the interference of unreliable pseudo-labels.

  % We propose a Dynamic Reliability-Enhanced Pseudo-Labeling (DREPL) generation strategy. In this strategy, the Dynamic History-Fused Prediction (DHP) module first generates a historical fused confidence by integrating historical and current predictions, and dynamically adjusts the historical weight according to training stage characteristics, thereby mitigating pseudo-label fluctuations caused by relying solely on current predictions. Subsequently, the Adaptive Tripartite Sample Categorization (ATSC) mechanism is introduced, which uses historical fused confidence and prediction consistency as joint evaluation metrics to dynamically categorize unlabeled samples into easy, ambiguous, and difficult classes. Differentiated optimization strategies are then applied for each class, strengthening the training of easy and ambiguous pseudo-labels while suppressing the noise introduced by difficult samples.
  
\end{enumerate}

This paper begins with a detailed presentation of the proposed framework in (see Sect.~\ref{method}). (see Sect.~\ref{experiments}) continues with a comprehensive analysis of the experimental results obtained from several public datasets. The paper ends with conclusions and future research direction in (see Sect.~\ref{conclusion}).

\section{Methodology}\label{method}

\begin{figure*}[t]
  \centering
  \includegraphics[width=\textwidth]{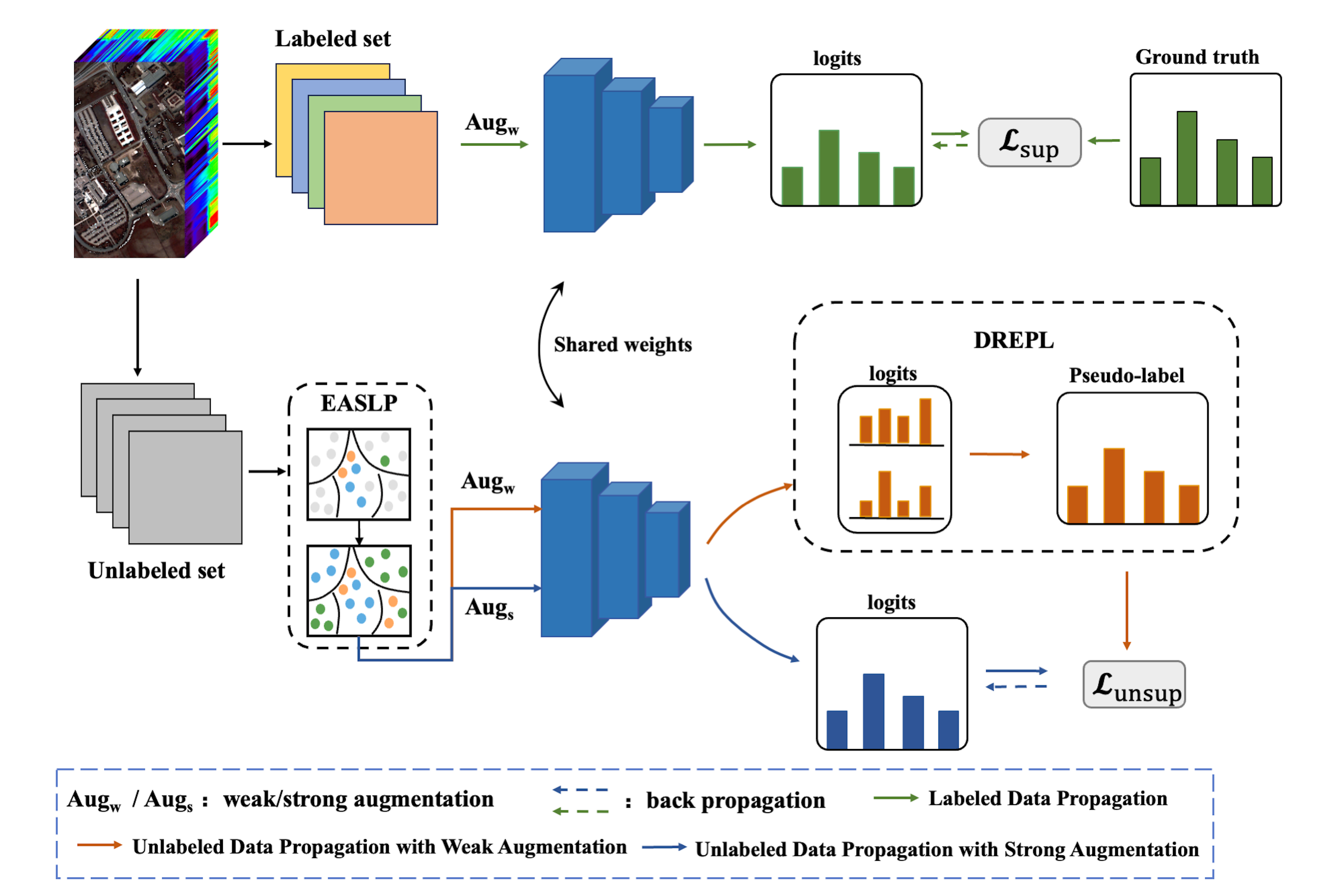}
  \caption{The overall framework of the proposed semi-supervised HSIC method.}
  \label{fig:1}
\end{figure*}

The overall workflow of the proposed semi-supervised HSIC method is illustrated in Fig~\ref{fig:1}. 

For labeled data, weak augmentation \(Aug_w\) is first applied before being fed into a shared feature extraction network consisting of three convolutional blocks, each containing a $3 \times 3$ convolution layer followed by batch normalization, ReLU activation, and a $2 \times 2$ max-pooling layer. The processed features then pass through a classification head, and the predictions are computed with the ground truth labels to calculate the supervised loss $\mathcal{L}_{\text{sup}}$. 

Unlabeled data first acquires spatial constraints via EASLP to address label diffusion along boundaries, then follows a dual-path augmentation scheme (\(Aug_w/Aug_s\))  feeding into shared feature extraction. Specifically, the weak augmentation \(Aug_w\) introduces mild perturbations (e.g., random flipping), while the strong augmentation \(Aug_s\) applies more aggressive transformations (e.g., noise disturbances). The weak augmentation \(Aug_w\) path flows into DHP and ATSC-based pseudo-label generation, whereas those from strong augmentation \(Aug_s\) path are used to compute the unsupervised loss $\mathcal{L}_{\text{unsup}}$ under pseudo-label guidance. This integrated DHP-ATSC framework establishes the proposed DREPL framework.  

\subsection{Edge-Aware Superpixel Label Propagation (EASLP)}\label{EPSLP}

\begin{figure*}[t]
  \centering
  \includegraphics[width=\textwidth]{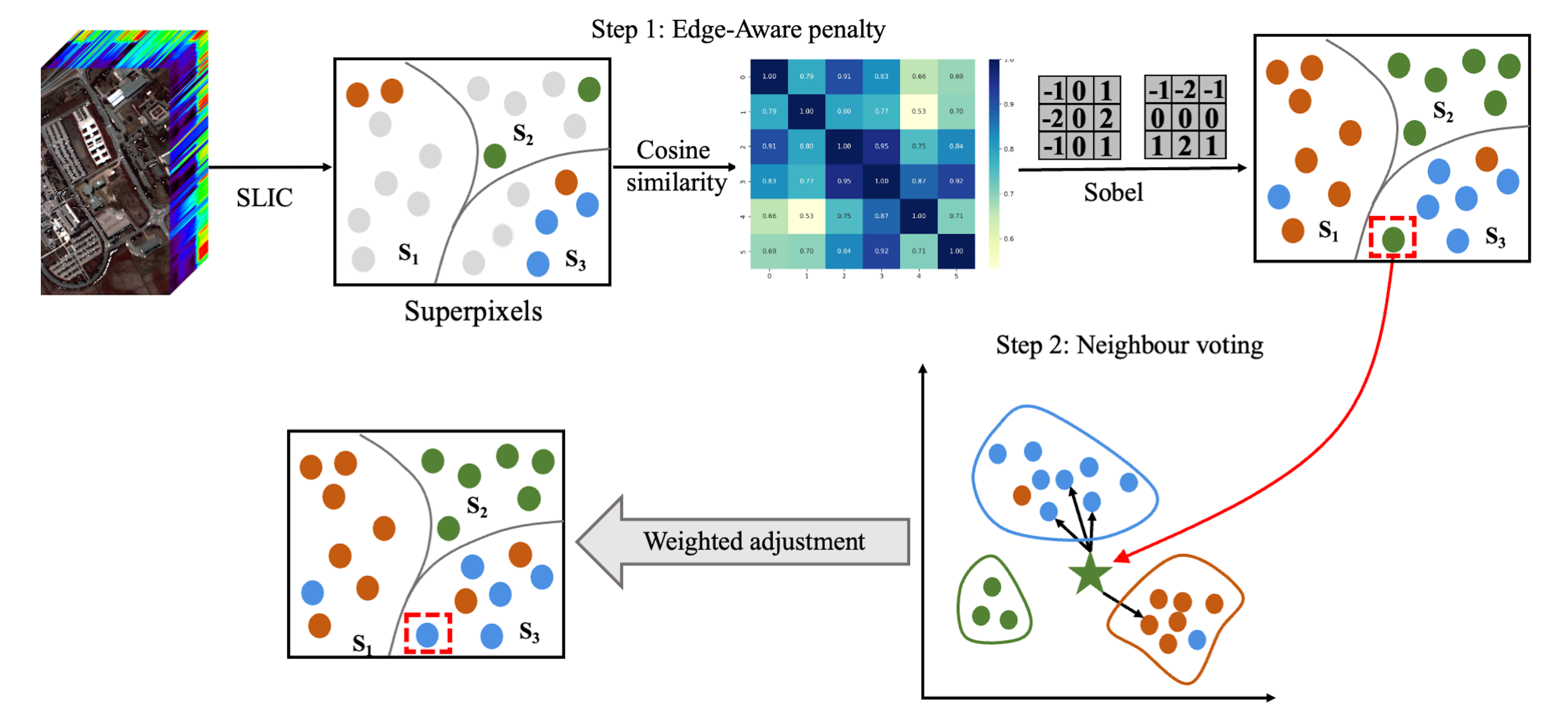}
  \caption{Structure of the Edge-Aware Superpixel Label Propagation (EASLP) module.}
  \label{fig:EASLP}
\end{figure*}

To address label propagation issues at the boundaries, the EASLP module incorporates an edge intensity penalty mechanism during the composition stage, adaptively suppressing label diffusion across the boundaries. Following propagation, an edge-sensitive neighborhood correction strategy is employed to further refine the labels of boundary pixels. This module can be seamlessly integrated with the backbone framework for structural enhancement in pseudo-label generation.

To reduce spectral redundancy while preserving the dominant spectral-spatial structure, PCA is first applied to the input hyperspectral image, followed by superpixel segmentation using the Simple Linear Iterative Clustering (SLIC) algorithm \cite{ref38}. It should be noted that the data after PCA dimensionality reduction are only used for superpixel generation, while the original full-band HSI data are retained for subsequent training and classification.

Specifically, the SLIC algorithm partitions the HSI into several non-overlapping superpixel regions with strong spectral–spatial consistency, as illustrated in Fig.~\ref{fig:1}. These regions are further grouped into multiple superpixel clusters (e.g., $S_1$, $S_2$, and $S_3$), each containing both labeled and unlabeled samples, thereby facilitating the propagation of local structural information during semi-supervised learning.

% The process begins by applying the Simple Linear Iterative Clustering (SLIC) algorithm \cite{ref38} to perform superpixel segmentation on the input HSI, yielding several non-overlapping regions with strong spatial–spectral consistency (as shown in Fig~\ref{fig:1}). These regions are divided into multiple superpixel clusters (e.g., $S_1$, $S_2$, $S_3$,  etc.). It contains labeled data as well as unlabeled data. 

For superpixels containing labeled samples, their categories are determined by the annotations within the region and propagated to the unlabeled samples. If a superpixel contains conflicting labeled classes or does not include any labeled samples, it is treated as an unlabeled region, where the cosine similarity between its spectral mean and the spectral mean of each training sample category is computed, serving as a matching score for label prediction:

\begin{equation}
\text{Sim}_{ij} = \frac{\mathbf{v}_i \cdot \mathbf{v}_j}{\|\mathbf{v}_i\| \cdot \|\mathbf{v}_j\|}
\end{equation}

\noindent where $\mathbf{v}_i$ denotes the spectral mean of the training samples in the $i$-th class; $\mathbf{v}_j$ denotes the spectral mean of the $j$-th unlabeled superpixel. The notation $\mathbf{v}_i \cdot \mathbf{v}_j$ represents the inner product between the two vectors.

% \noindent where $\mathbf{v}_i$ denotes the spectral mean of the training samples in the $i$-th class; $\mathbf{v}_j$ denotes the spectral mean of the unlabeled superpixels $j$.

To prevent the propagation of erroneous labels in boundary regions, we introduce an edge intensity map based on the Sobel filter to explicitly incorporate spatial structural constraints during similarity propagation. Specifically, the Sobel operator is a first-order gradient operator that calculates the rate of grayscale variation in both horizontal and vertical directions within an image, thereby detecting edge structures of features or objects.

Firstly, we take the mean along the spectral dimension of the hyperspectral data $\mathbf{X} \in \mathbb{R}^{H \times W \times B}$ to obtain a 2-D grayscale image $I$:

\begin{equation}
I(x, y) = \frac{1}{B} \sum_{b=1}^{B} \mathbf{X}(x, y, b)
\label{eq:grayscale}
\end{equation}

Subsequently, a Sobel filter is applied to $I$ to compute the gradient magnitude for each pixel:

\begin{equation}
E(x, y) = \sqrt{(G_x * I)^2 + (G_y * I)^2}
\label{eq:edge_magnitude}
\end{equation}

\noindent where $G_x$ and $G_y$ represent Sobel's fixed convolution kernels in the horizontal and vertical directions, respectively:

\begin{equation}
G_x =
\begin{bmatrix}
-1 & 0 & +1 \\
-2 & 0 & +2 \\
-1 & 0 & +1
\end{bmatrix},
\qquad
G_y =
\begin{bmatrix}
-1 & -2 & -1 \\
0 & 0 & 0 \\
+1 & +2 & +1
\end{bmatrix}.
\label{eq:sobel_kernels}
\end{equation}

The resulting $E(x, y)$ reflects the local intensity of gray-scale variation at each pixel, i.e., the edge significance. 
This edge intensity map is normalized and used to evaluate the average edge response for each superpixel region:

\begin{equation}
E_j = \frac{1}{|S_j|} \sum_{(x,y) \in S_j} E(x, y),
\label{eq:avg_edge_response}
\end{equation}

\noindent where \( E_j \) denotes the average edge intensity of the \( j \)-th superpixel; \( |S_j| \) represents the number of pixels within the superpixel region \( S_j \).

During the similarity propagation stage, the edge intensity serves as a penalty factor to dynamically weight the similarity scores derived from cosine similarity. The weight is calculated as 

\begin{equation}
\tilde{\text{Sim}}_{ij} = \frac{\text{Sim}_{ij}}{1 + E_j}
\end{equation}

Compared to weight functions in logarithmic or exponential forms, this linear decay strategy maintains higher similarity scores when $E_j \to 0$, while automatically attenuating the propagation effect when ${E}_j$ becomes large. 
This mechanism achieves a more robust balance between edge and non-edge regions. The first similarity matrix is obtained by computing the cosine similarity and weighting it with the edge intensity derived from the Sobel filter. 

Based on this matrix, each unlabeled superpixel region is assigned the class of the training sample with the highest cosine similarity to its spectral mean, as illustrated in Fig~\ref{fig:EASLP}.

Furthermore, to enhance the stability of pseudo-labels in boundary regions, we employ a neighborhood-based edge-weighted correction mechanism. Specifically, each unlabeled superpixel collects label votes from its adjacent superpixels, with weights assigned according to the edge intensity of its neighbors. In this way, the label $\hat{y}_j$ for each superpixel $j$ is obtained. The correction formula is defined as

\begin{equation}
\hat{y}_j = \arg \max_{c} \sum_{k \in \mathcal{N}(j)} \frac{1}{E_k + \epsilon} \cdot \mathbb{I}[y_k = c]
\label{eq:formula3}
\end{equation}

\noindent where \(\mathcal{N}(j)\) denotes the neighborhood set of superpixel \(j\), \( E_k \) represents the edge intensity of adjacent superpixel \( k \), \( \mathbb{I}[y_k = c] \) is the indicator function that equals 1 when the label of \( k \) is \( c \) and 0 otherwise, and \( \epsilon = 10^{-6} \) is a small constant to prevent division by zero.

% \textcolor{red}{When the edge intensities of neighboring superpixels are identical, Eq.~(\ref{eq:formula3}) reduces to a standard majority voting strategy, while uniformly high edge intensities do not alter the relative voting relationships among neighboring classes.} Therefore, superpixels with weaker edge responses contribute more strongly to the voting process, whereas those separated by strong edges have limited influence, reducing the risk of incorrect label correction across category boundaries.
% Therefore, regions with weaker edges exert greater influence on label propagation, thereby guiding pseudo-labels toward areas of strong internal consistency and preventing their propagation across category boundaries, ultimately correcting errors in the initial classification

Upon completion of the entire process, we generate a structure-aware pseudo-label map and perform balanced sampling of unlabeled training samples from each pseudo-label category to support semi-supervised training.

\subsection{Dynamic History-Fused Prediction (DHP)}

To address the issue of temporal fluctuation in pseudo-labels, we propose a DHP method that leverages historical information to improve the pseudo-label stability, with its overall structure illustrated in Fig~\ref{fig:DREPL}.

\subsubsection{Historical Prediction Queue Modeling}

Initially, a historical prediction queue is maintained for each unlabeled sample, recording its categorical prediction frequencies over the most recent T iterations.

Specifically, the historical prediction queue is given by:

\begin{equation}
Q_i^{(t)} = \{ \text{cnt}_i^{(t)}(C_1), \text{cnt}_i^{(t)}(C_2), \dots, \text{cnt}_i^{(t)}(C_K) \}
\end{equation}

\noindent where $Q_i^{(t)}$ denotes the historical prediction statistics of sample $u_i$ at the $t$-th training epoch,  
$\text{cnt}_i^{(t)}(C_k) (k=1,\cdots,K)$ represents the frequency with which sample $u_i$ is predicted as category $C_k$ over the most recent $T$ iterations. 

\subsubsection{Queue Update and Growth Strategy}

Subsequently, an exponential growth strategy is employed to dynamically adjust the historical prediction queue length \( L \). 

The queue length update rule is defined as

\begin{equation}
L(t) = L_{\min} \cdot \left( \frac{L_{\max}}{L_{\min}} \right)^{\tfrac{t}{T_{\text{maxEpoch}}}}
\end{equation}

\noindent where $L_{min}$ and $L_{max}$ represent the minimum and maximum queue length, respectively, $t$ denotes the current training epoch, and $T_{maxEpoch}$ indicates the total number of training epochs. The historical statistics are computed within a finite sliding window of length \(L(t)\), such that outdated predictions are gradually discarded once the queue reaches its current capacity.

\subsubsection{Dynamic Weighted Fusion Strategy}

\begin{figure*}[t]
  \centering
  \includegraphics[width=\textwidth]{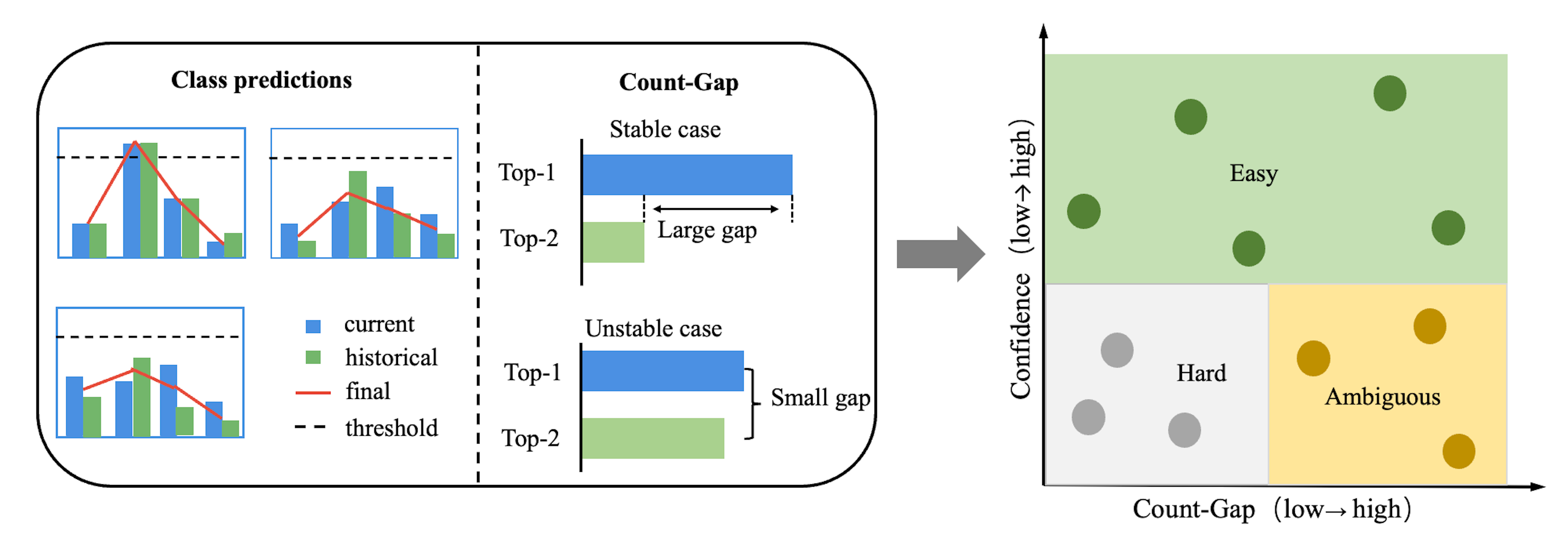}
  \caption{Structure of the Dynamic Reliability-Enhanced Pseudo-Labeling (DREPL) module.}
  \label{fig:DREPL}
\end{figure*}

We introduce a dynamic weighted fusion strategy to effective integration of historical and current predictions. Defining $P_{\text{cur}}$ and $P_{\text{hist}}$ as the current and historical prediction distributions, respectively, the final fused distribution is formulated as

\begin{equation}
P_{\text{fuse}}(u_i) = (1-\alpha)P_{\text{cur}}(u_i) + \alpha P_{\text{hist}}(u_i), 
\end{equation}

\begin{equation}
P_{\text{hist}}(u_i) = \frac{\text{cnt}_i^{(t)}(k)}{\sum_{k=1}^{K} \text{cnt}_i^{(t)}(k)}
\end{equation}

\noindent where $\alpha$ represents the weight assigned to historical predictions, being gradually incremented throughout training, with its evolution governed by:

\begin{equation}
\alpha_t = \min \left( \alpha_{\text{max}}, \, \alpha_{\text{min}} + (\alpha_{\text{max}} - \alpha_{\text{min}}) 
\cdot \frac{t - t_0}{T_{\text{maxEpoch}} - t_0} \right)
\end{equation}

\noindent where $\alpha_{min}$ and $\alpha_{max}$ denote the minimum and maximum weight values, respectively, \(t_0\) represents the epoch at which historical predictions are first incorporated. As illustrated in Fig~\ref{fig:DREPL}, the category prediction is obtained through a weighted fusion of historical and current predictions, thereby effectively smoothing the prediction distribution and yielding more stable and reliable pseudo-labels.

\subsection{Adaptive Tripartite Sample Categorization (ATSC)}\label{ATSC}

\subsubsection{Confidence-Driven Sample Selection}

% For quantitative assessment of prediction stability, the Count-Gap (CG) metric is introduced, utilizing temporally accumulated prediction records from unlabeled sample. The metric calculates the frequency disparity between the top-two most frequently predicted categories:
A short warm-up stage is adopted before activating ATSC, during which only labeled samples are used for model optimization. In the latter part of the warm-up stage, predictions on unlabeled samples are accumulated only to initialize and maintain the historical prediction queue for each sample, without contributing to the training loss. This design reduces the influence of potentially incorrect pseudo-labels and alleviates the risk of confirmation bias and progressive error accumulation during training.

For quantitative assessment of prediction stability, the Count-Gap (CG) metric proposed by CGMatch\cite{ref51} is introduced, which utilizes temporally accumulated prediction records from unlabeled samples. The metric calculates the frequency disparity between the top-two most frequently predicted categories:

\begin{equation}
CG(u_i) = f_{\max}(u_i) - f_{\text{second}}(u_i)
\end{equation}

\noindent where \( f_{\text{max}}(u_i) \) denotes the count of times the sample \( u_i \) is predicted as the most frequent category, and \( f_{\text{second}}(u_i) \) represents the corresponding count for the second most frequent category.

\subsubsection{Adaptive Sample Classification}

Utilizing  the historical consistency measure, the framework employs the fused prediction distribution $P_{\text{fuse}}(u_i)$ for adaptive classification of unlabeled samples. The procedure commences with calculating the prediction confidence per sample:

\begin{equation}
Conf(u_i) = \max \left( P_{\text{fuse}}(u_i) \right)
\end{equation}

Based on the confidence and CG values, we adopt a dual-threshold strategy, as illustrated in Fig~\ref{fig:DREPL}, to categorize unlabeled samples into three distinct classes:

\textbf{Easy samples}: Confidence score $\text{Conf}(u_i) > \tau_c$. The pseudo-labels are reliable and can be directly incorporated into training.

\textbf{Ambiguous samples}: Confidence score $\text{Conf}(u_i) \leq \tau_c$ and $CG > \tau_a$, indicating unstable predictions that require cautious utilization.

\textbf{Hard samples}: Confidence score $\text{Conf}(u_i) < \tau_c$ and $CG < \tau_a$, indicating significant prediction fluctuations. Such samples are temporarily discarded to avoid introducing noise.

\subsubsection{Adaptive Threshold Update Mechanism}

To further enhance the flexibility of sample categorization, we dynamically update thresholds $\tau_a$ and $\tau_c$ using an exponential moving average (EMA) mechanism:

\begin{equation}
\tau_{a}^{(t)} = m \cdot \tau_{a}^{(t-1)} + (1-m)\cdot \mu_{a}^{(t)}, 
\quad 
\mu_{a}^{(t)} = \frac{1}{B_u} \sum_{i=1}^{B_u} CG_i
\end{equation}

\begin{equation}
\tau_{c}^{(t)} = m \cdot \tau_{c}^{(t-1)} + (1-m)\cdot \mu_{c}^{(t)}, 
\quad 
\mu_{c}^{(t)} = \frac{1}{B_u} \sum_{i=1}^{B_u} Conf(u_i)
\end{equation}

\noindent where $m$ indicates the smoothing coefficient and $\tau^{(t)}$ denotes the candidate threshold computed for the current batch $B_u$. 

This adaptive partitioning method overcomes the limitations of binary approaches and enhances the utilization efficiency of pseudo-labels.

\subsection{Loss Function}

\subsubsection{Supervised Loss}

The labeled data objective minimizes the standard cross-entropy loss:

\begin{equation}
\mathcal{L}_{sup} = -\frac{1}{N_l} \sum_{i=1}^{N_l} \log p(y_i \mid x_i)
\end{equation}

\noindent where $N_l$ denotes the number of labeled samples,$x_i$ denotes the input data of the $i$-th labeled sample, $y_i$ represents the true label, and $p(\cdot)$ is the predicted probability distribution.

\subsubsection{Self-training Loss for Unlabeled Samples}

A differentiated loss scheme is implemented for unlabeled samples according to pseudo-label confidence and CG metric, following a tripartite division into easy, ambiguous, and hard categories.

\textbf{Easy samples}: High confidence and strong consistency indicate reliable pseudo-labels, permitting the direct application of cross-entropy loss:

\begin{equation}
\mathcal{L}_{easy} = -\frac{1}{N_e} \sum_{i \in \mathcal{E}} \log p(\hat{y}_i \mid u_i^{strong})
\end{equation}

\noindent where $\hat{y}_i$ indicates the generated pseudo-label and $u_i^{\text{strong}}$ denotes strongly augmented samples.

\textbf{Ambiguous samples}: Samples characterized by intermediate or fluctuating confidence levels are supervised using temperature-scaled soft prediction distribution $p(\cdot)$ rather than hard pseudo-labels. The consistency is regularized through KL divergence as

\begin{equation}
\mathcal{L}_{amb} = \frac{1}{N_a} \sum_{i \in \mathcal{A}} 
D_{KL} \left( \tilde{p}(u_i^{weak}) \, \| \, p(u_i^{strong}) \right)
\end{equation}

\noindent where $\mathcal{A}$ is the ambiguous sample set, while $\tilde{p}(u_i^{\text{weak}})$ and $p(u_i^{\text{strong}})$ correspond to the soft prediction distributions under weak and strong augmentation, respectively.

\textbf{Hard samples}: Samples with excessively low confidence and significant prediction fluctuations are directly discarded to avoid the detrimental effects of misleading noisy labels. 

The overall self-training loss for unlabeled samples is formulated below:

\begin{equation}
\mathcal{L}_{unsup} = \mathcal{L}_{easy} + \lambda_a \cdot \mathcal{L}_{amb}
\end{equation}

\noindent where $\lambda_a$ is a dynamically adjusted weight during training that controls the contribution of ambiguous samples.

\subsubsection{Overall Objective}

In summary, the model's final optimization objective is expressed by:

\begin{equation}
\mathcal{L} = \mathcal{L}_{\text{sup}} + \mathcal{L}_{\text{unsup}},
\end{equation}

Specifically, it jointly optimizes the supervised classification loss on labeled samples and the self-training loss on unlabeled samples. 

\section{Experiments and analysis}\label{experiments}

\subsection{Dataset details}\label{subsec1}

A comprehensive evaluation of the proposed method is conducted employing five benchmark HSI datasets: Pavia University (PaviaU), Houston2013, Salinas, Kennedy Space Center (KSC), and Botswana. Spanning urban, forests, wetlands, and agricultural scenarios with distinct spectral-spatial features, these datasets enable rigorous evaluation of methodological robustness and generalization in complex settings. Detailed parameters (image size, spectral bands, and per-class labeled samples) for all datasets are summarized in Table~\ref{tab:division}.

\begin{table*}[!t]
\centering
\caption{Detailed parameter information for different datasets.}
\label{tab:division}
\renewcommand{\arraystretch}{1.2}
\setlength{\tabcolsep}{4pt}

\resizebox{\textwidth}{!}{%
\begin{tabular}{c|lcc|lcc|lcc|lcc|lcc}
\hline
\textbf{Dataset} &
\multicolumn{3}{c|}{\textbf{PaviaU}} &
\multicolumn{3}{c|}{\textbf{Houston2013}} &
\multicolumn{3}{c|}{\textbf{KSC}} &
\multicolumn{3}{c|}{\textbf{Botswana}} &
\multicolumn{3}{c}{\textbf{Salinas}} \\
\hline
Sensor &
\multicolumn{3}{c|}{ROSIS} &
\multicolumn{3}{c|}{CASI-1500} &
\multicolumn{3}{c|}{AVIRIS} &
\multicolumn{3}{c|}{Hyperion} &
\multicolumn{3}{c}{AVIRIS} \\
\hline
Size \& Bands &
\multicolumn{3}{c|}{$610\times340\times103$} &
\multicolumn{3}{c|}{$349\times1905\times144$} &
\multicolumn{3}{c|}{$512\times614\times176$} &
\multicolumn{3}{c|}{$1476\times256\times145$} &
\multicolumn{3}{c}{$512\times217\times204$} \\
\hline
\textbf{Class} & Name & Training & Test &
Name & Training & Test &
Name & Training & Test &
Name & Training & Test &
Name & Training & Test \\
\hline
1  & Asphalt                & 10 & 6621 &
Healthy Grass        & 10 & 1241 &
Scrub                & 10 & 751 &
Water                & 10 & 260 &
Brocoli green weeds 1 & 10 & 1999 \\
2  & Meadows                & 10 & 18639 &
Stressed Grass       & 10 & 1244 &
Willow Swamp         & 10 & 233 &
Hippo Grass          & 10 & 91 &
Brocoli green weeds 2 & 10 & 3716 \\
3  & Gravel                 & 10 & 2089 &
Synthetic Grass      & 10 & 687  &
Cabbage Palm Hammock & 10 & 246 &
Floodplain Grasses 1 & 10 & 241 &
Fallow                & 10 & 1966 \\
4  & Trees                  & 10 & 3054 &
Trees                & 10 & 1234 &
Cabbage Palm/Oak     & 10 & 242 &
Floodplain Grasses 2 & 10 & 205 &
Fallow rough plow     & 10 & 1384 \\
5  & Painted Metal Sheets   & 10 & 1335 &
Soil                 & 10 & 1232 &
Slash Pine           & 10 & 151 &
Reeds                & 10 & 259 &
Fallow smooth         & 10 & 2668 \\
6  & Bare Soil              & 10 & 5019 &
Water                & 10 & 315  &
Oak/Broadleaf        & 10 & 219 &
Riparian             & 10 & 259 &
Stubble               & 10 & 3949 \\
7  & Bitumen                & 10 & 1320 &
Residential          & 10 & 1258 &
Hardwood Swamp       & 10 & 95  &
Firescar             & 10 & 249 &
Celery                & 10 & 3569 \\
8  & Self-blocking Bricks   & 10 & 3672 &
Commercial           & 10 & 1234 &
Graminoid Marsh      & 10 & 421 &
Island Interior      & 10 & 193 &
Grapes untrained      & 10 & 11261 \\
9  & Shadows                & 10 & 937  &
Road                 & 10 & 1242 &
Spartina Marsh       & 10 & 510 &
Acacia Woodlands     & 10 & 304 &
Soil vineyard develop & 10 & 6193 \\
10 & --                     & -- & --   &
Highway              & 10 & 1217 &
Cattail Marsh        & 10 & 394 &
Acacia Shrublands    & 10 & 238 &
Corn senesced green weeds & 10 & 3268 \\
11 & --                     & -- & --   &
Railway              & 10 & 1225 &
Salt Marsh           & 10 & 409 &
Acacia Grasslands    & 10 & 295 &
Lettuce romaine 4wk   & 10 & 1058 \\
12 & --                     & -- & --   &
Parking Lot 1        & 10 & 1223 &
Mud Flats            & 10 & 493 &
Short Mopane         & 10 & 171 &
Lettuce romaine 5wk   & 10 & 1917 \\
13 & --                     & -- & --   &
Parking Lot 2        & 10 & 459  &
Water                & 10 & 917 &
Mixed Mopane         & 10 & 258 &
Lettuce romaine 6wk   & 10 & 906 \\
14 & --                     & -- & --   &
Tennis Court         & 10 & 418  &
--                   & -- & --  &
Exposed Soils        & 10 & 85 &
Lettuce romaine 7wk   & 10 & 1060 \\
15 & --                     & -- & --   &
Running Track        & 10 & 650  &
--                   & -- & --  &
--                   & -- & -- &
Vineyard untrained    & 10 & 7258 \\
16 & --                     & -- & --   &
--                   & -- & --  &
--                   & -- & --  &
--                   & -- & -- &
Vineyard vertical trellis & 10 & 1797 \\
\hline
Total &  & 90 & 42686 &
 & 150 & 14879 &
 & 130 & 5081 &
 & 140 & 3108 &
 & 160 & 53969 \\
\hline
\end{tabular}%
}
\end{table*}

\begin{figure*}[!t]
\centering

\newcommand{\colw}{0.18\linewidth}
\newcommand{\colsep}{-0.01\linewidth}

%==================== 第一行 ====================%
\begin{minipage}[t]{\colw}\centering
    \includegraphics[width=0.7\linewidth]{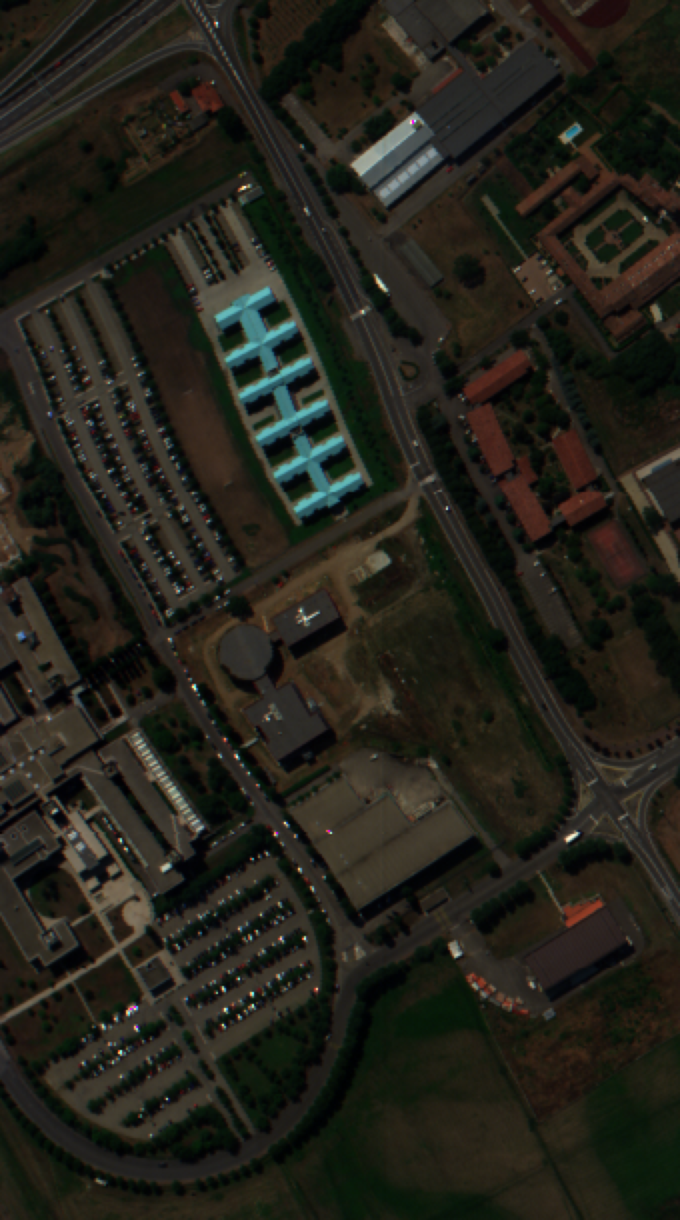}\\[0.2em]
    \scriptsize(a)
\end{minipage}\hspace{\colsep}
\begin{minipage}[t]{\colw}\centering
    \includegraphics[width=0.7\linewidth]{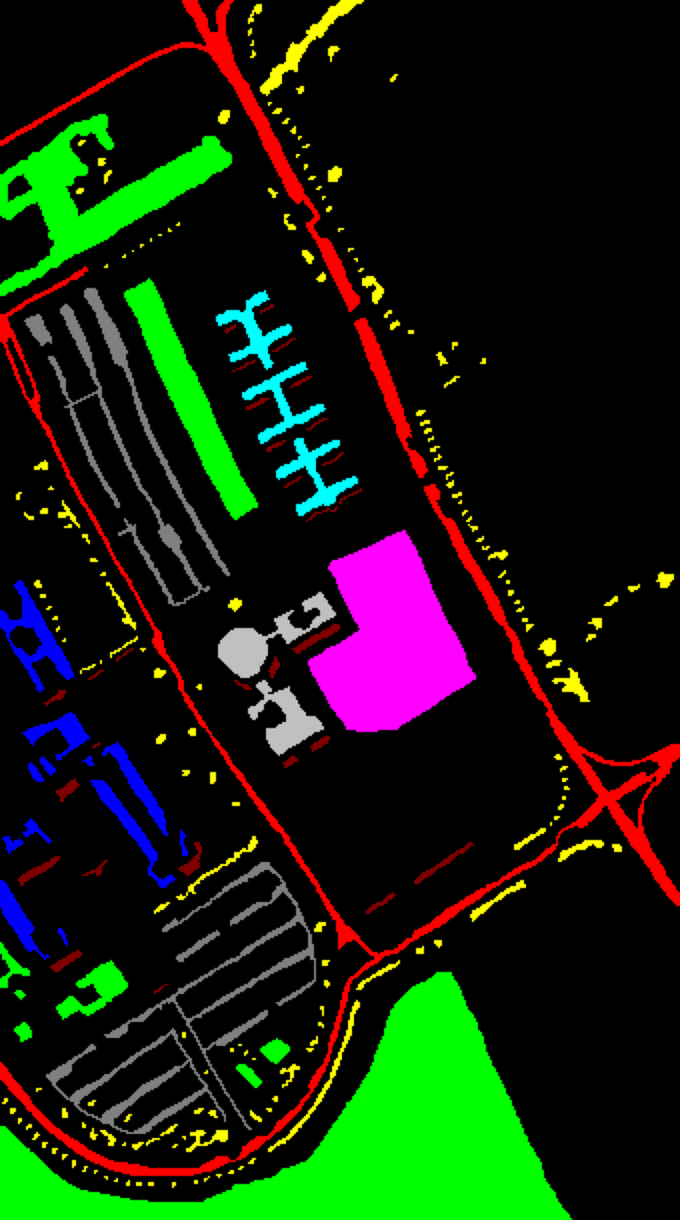}\\[0.2em]
    \scriptsize(b)
\end{minipage}\hspace{\colsep}
\begin{minipage}[t]{\colw}\centering
    \includegraphics[width=0.7\linewidth]{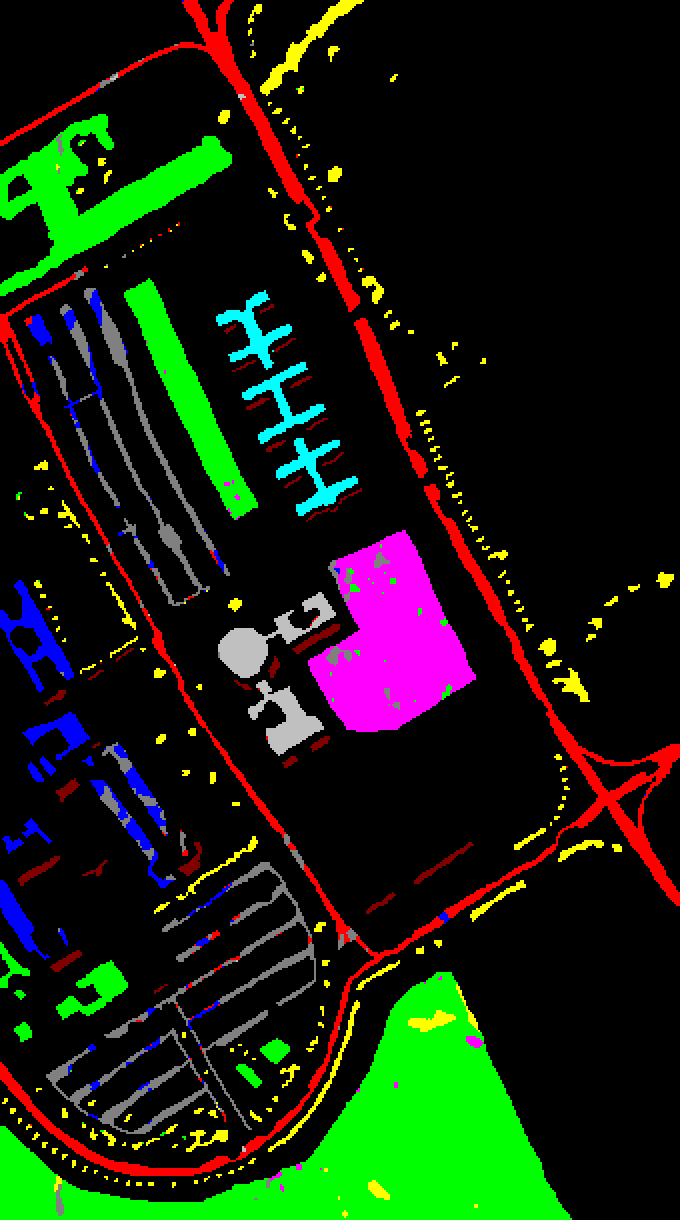}\\[0.2em]
    \scriptsize(c)
\end{minipage}\hspace{\colsep}
\begin{minipage}[t]{\colw}\centering
    \includegraphics[width=0.7\linewidth]{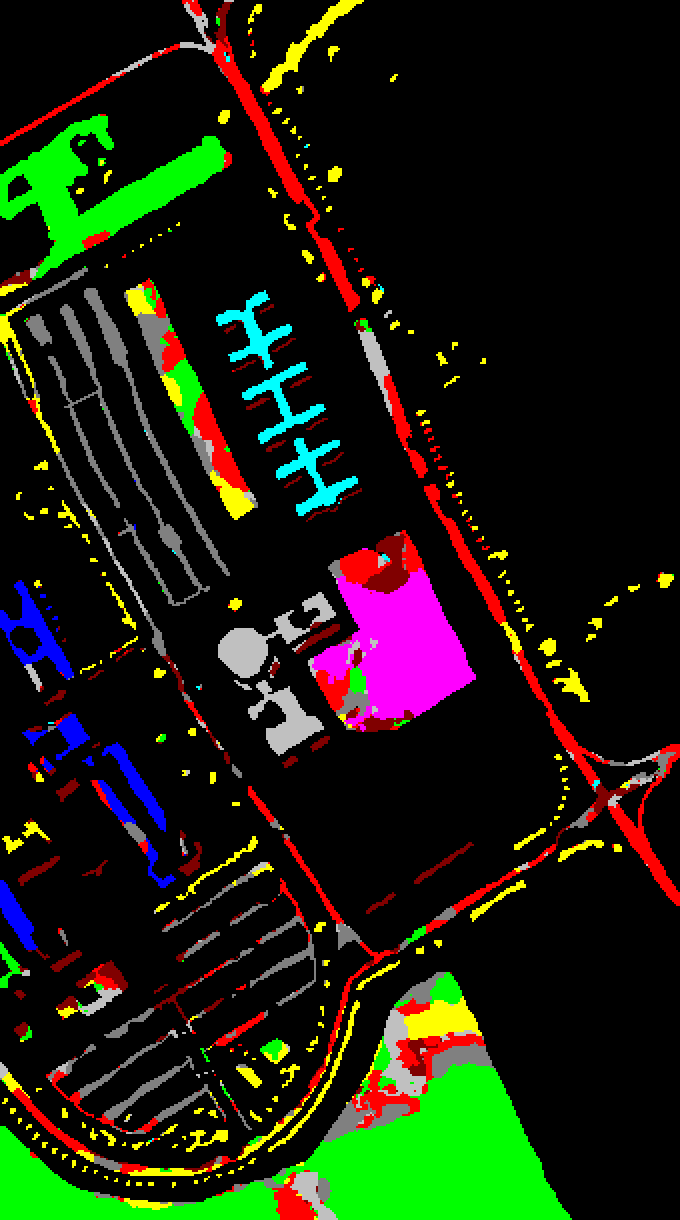}\\[0.2em]
    \scriptsize(d)
\end{minipage}\hspace{\colsep}
\begin{minipage}[t]{\colw}\centering
    \includegraphics[width=0.7\linewidth]{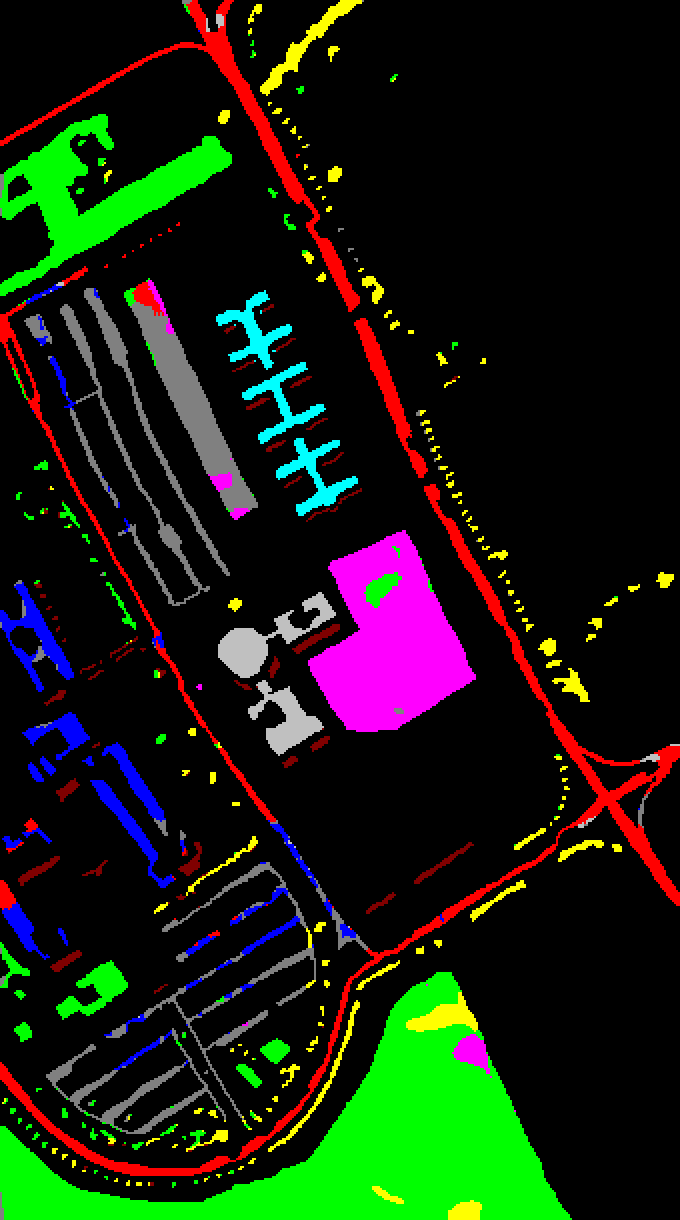}\\[0.2em]
    \scriptsize(e)
\end{minipage}

\vspace{0.8em}

%==================== 第二行 ====================%
\begin{minipage}[t]{\colw}\centering
    \includegraphics[width=0.7\linewidth]{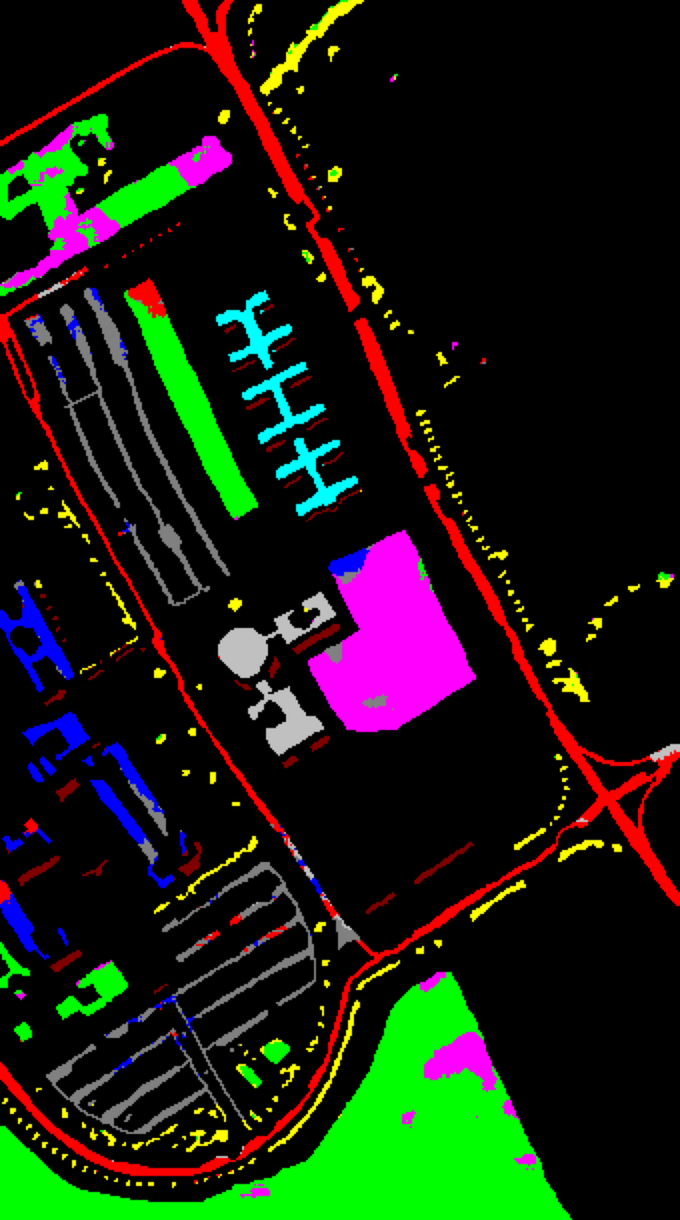}\\[0.2em]
    \scriptsize(f)
\end{minipage}\hspace{\colsep}
\begin{minipage}[t]{\colw}\centering
    \includegraphics[width=0.7\linewidth]{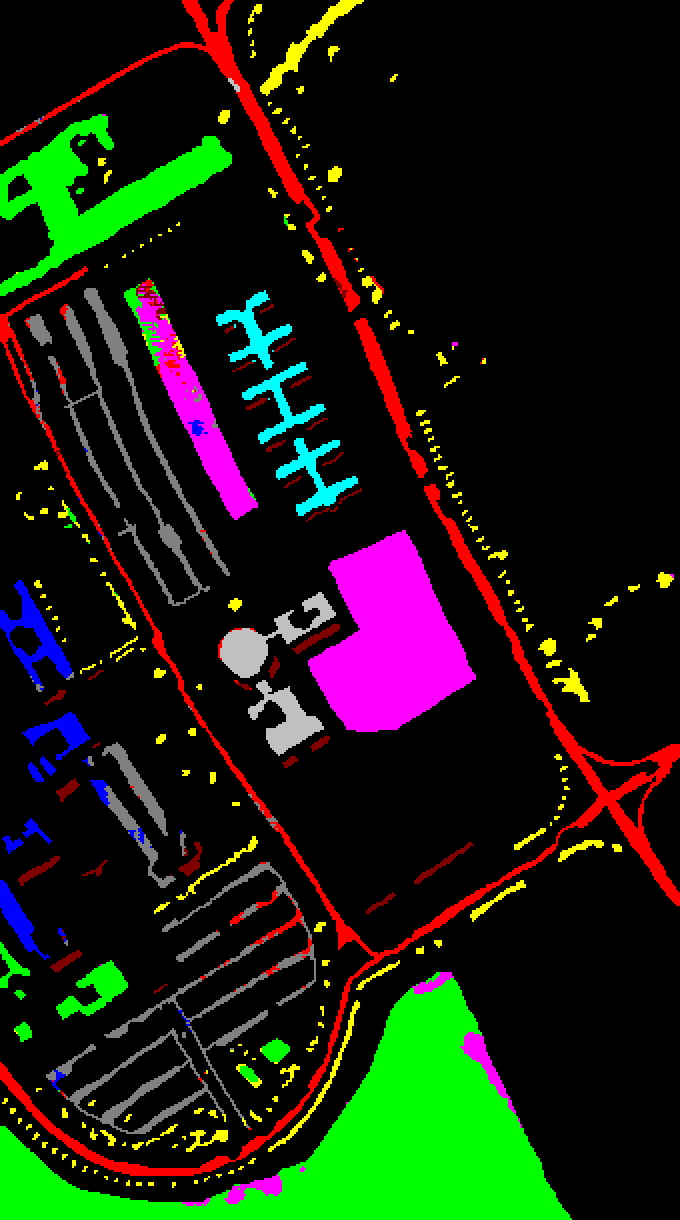}\\[0.2em]
    \scriptsize(g)
\end{minipage}\hspace{\colsep}
\begin{minipage}[t]{\colw}\centering
    \includegraphics[width=0.7\linewidth]{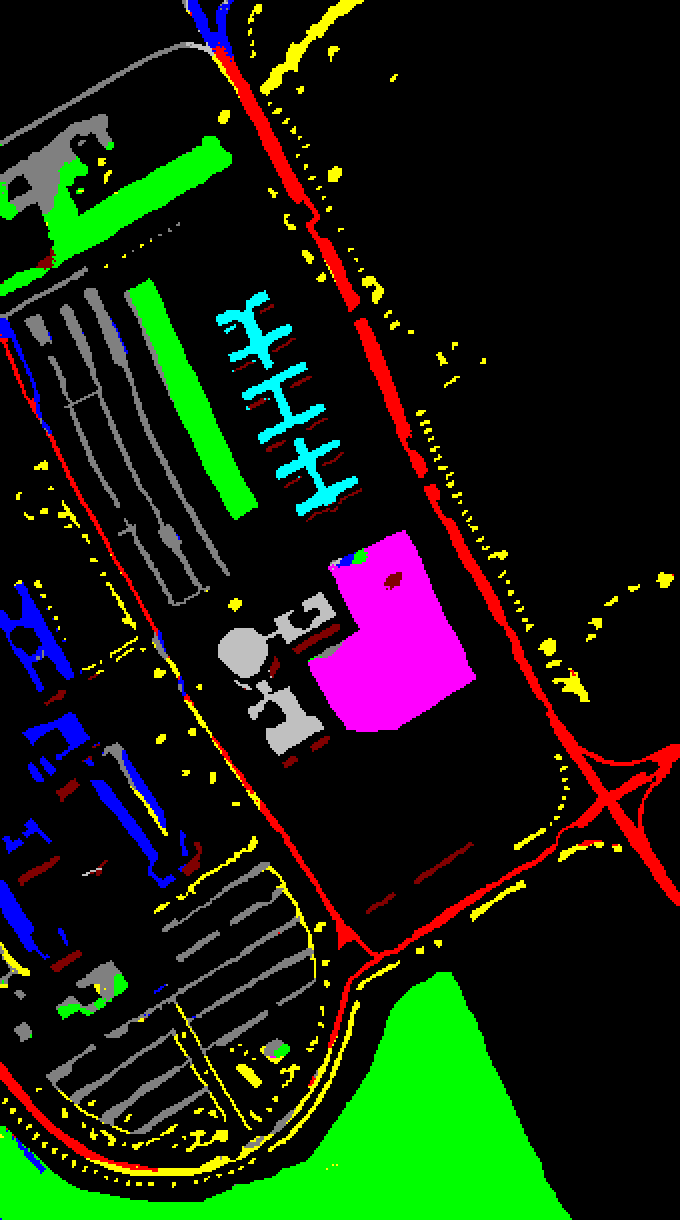}\\[0.2em]
    \scriptsize(h)
\end{minipage}\hspace{\colsep}
\begin{minipage}[t]{\colw}\centering
    \includegraphics[width=0.7\linewidth]{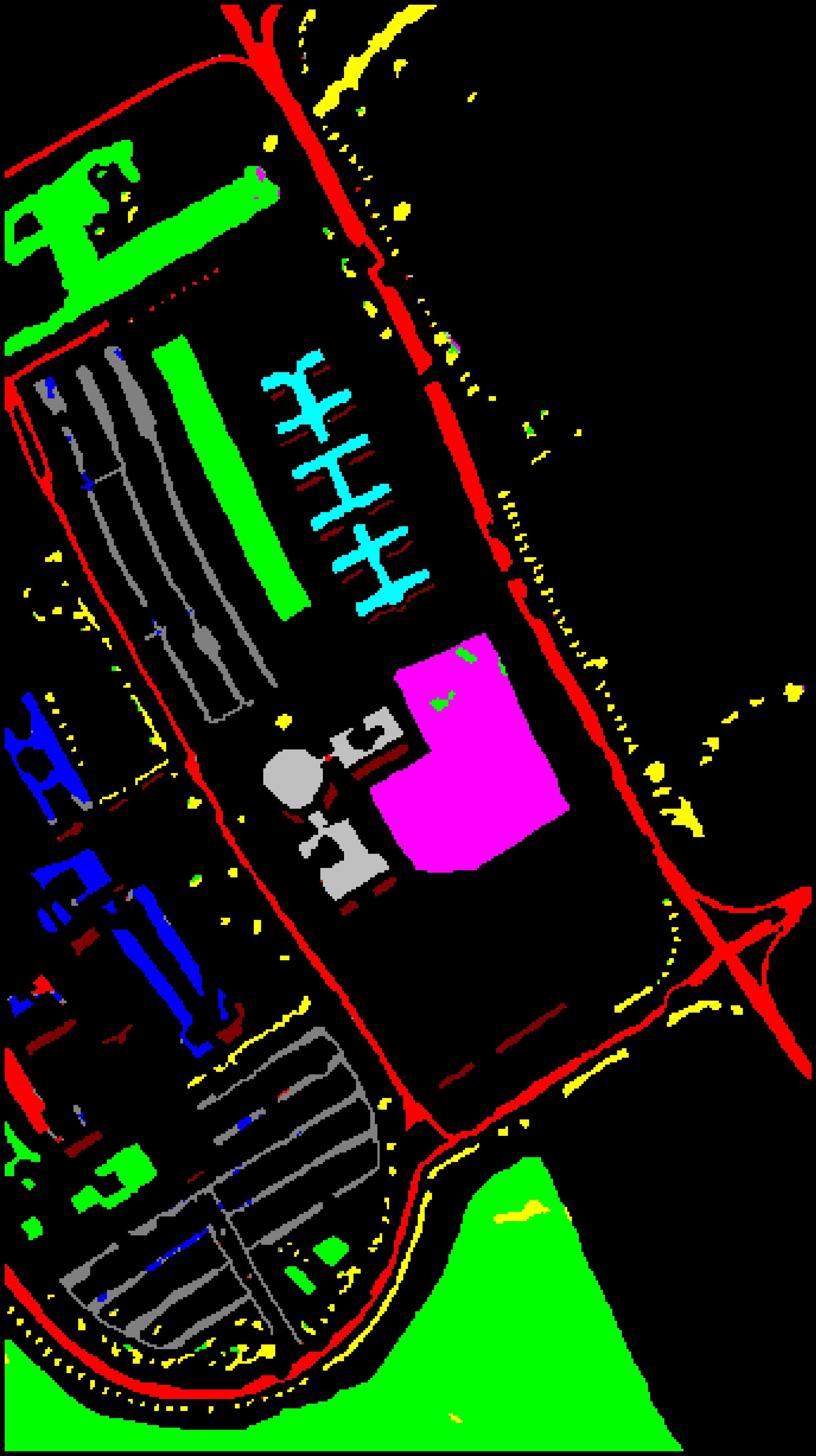}\\[0.2em]
    \scriptsize(i)
\end{minipage}\hspace{\colsep}
\begin{minipage}[t]{\colw}\centering
    \includegraphics[width=0.7\linewidth]{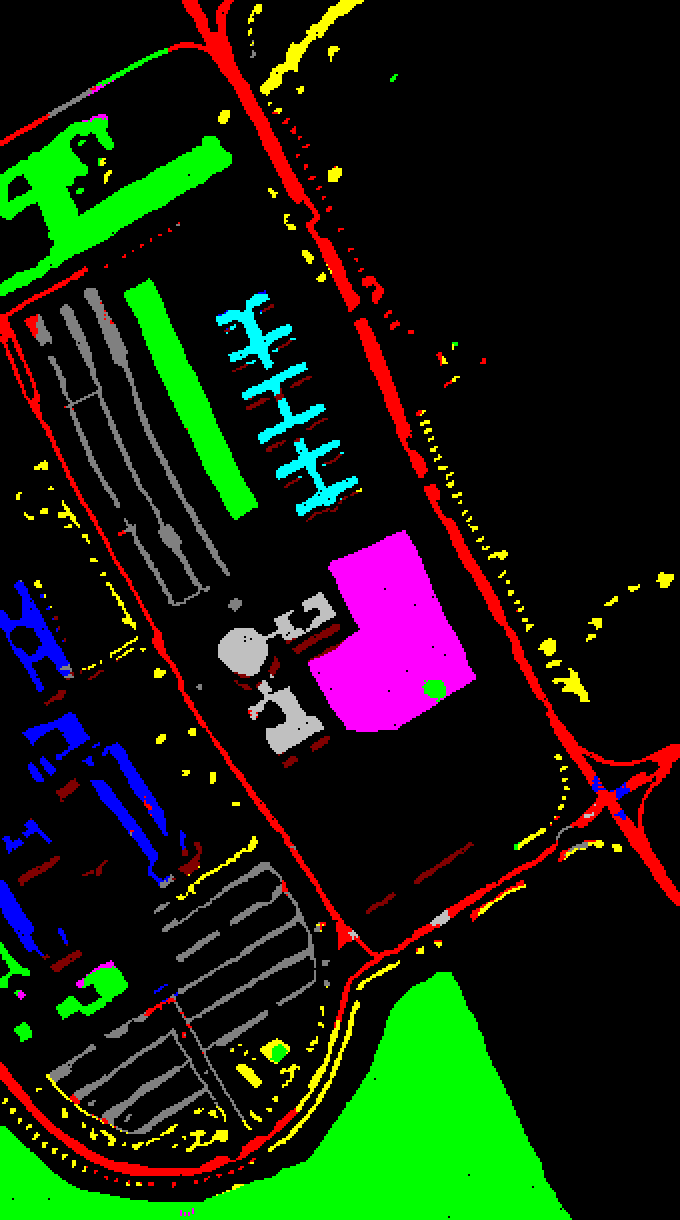}\\[0.2em]
    \scriptsize(j)
\end{minipage}

\vspace{0.1em}

%==================== 横向颜色条 ====================%
% \scriptsize
% \renewcommand{\arraystretch}{1.1}

% \begin{tabular}{cccccccccc}
% \textcolor[rgb]{0.0,0.0,0.0}{\rule{4mm}{4mm}} &
% \textcolor[rgb]{1.0,0.0,0.0}{\rule{4mm}{4mm}} &
% \textcolor[rgb]{0.0,1.0,0.0}{\rule{4mm}{4mm}} &
% \textcolor[rgb]{0.0,0.0,1.0}{\rule{4mm}{4mm}} &
% \textcolor[rgb]{1.0,1.0,0.0}{\rule{4mm}{4mm}} &
% \textcolor[rgb]{0.0,1.0,1.0}{\rule{4mm}{4mm}} &
% \textcolor[rgb]{1.0,0.0,1.0}{\rule{4mm}{4mm}} &
% \textcolor[rgb]{0.75,0.75,0.75}{\rule{4mm}{4mm}} &
% \textcolor[rgb]{0.5,0.5,0.5}{\rule{4mm}{4mm}} &
% \textcolor[rgb]{0.5,0.0,0.0}{\rule{4mm}{4mm}} \\

% Unlabelled &
% Asphalt &
% Meadows &
% Gravel &
% Trees &
% Metal &
% Bare Soil &
% Bitumen &
% Bricks &
% Shadows \\
% \end{tabular}

\begin{center}
\begin{tikzpicture}[
every node/.style={
    minimum width=2.8cm,
    minimum height=0.6cm,
    font=\small,
    inner sep=0pt,
    outer sep=0pt
}
]

% 第一行
\node[fill=black,text=white] at (1.4,0) {\textbf{Unlabelled}};
\node[fill=red,text=white] at (4.2,0) {\textbf{Asphalt}};
\node[fill=green,text=black] at (7.0,0) {\textbf{Meadows}};
\node[fill=blue,text=white] at (9.8,0) {\textbf{Gravel}};
\node[fill=yellow,text=black] at (12.6,0) {\textbf{Trees}};

% 第二行（紧贴第一行）
\node[fill=cyan,text=black] at (1.4,-0.6) {\textbf{Metal}};
\node[fill=magenta,text=white] at (4.2,-0.6) {\textbf{Bare Soil}};
\node[fill={rgb,1:red,0.75;green,0.75;blue,0.75},text=black] at (7.0,-0.6) {\textbf{Bitumen}};
\node[fill={rgb,1:red,0.5;green,0.5;blue,0.5},text=white] at (9.8,-0.6) {\textbf{Bricks}};
\node[fill={rgb,1:red,0.5;green,0;blue,0},text=white] at (12.6,-0.6) {\textbf{Shadows}};

\end{tikzpicture}
\end{center}

\vspace{0.1em}

% \scriptsize(k)

\caption{Classification maps for the PaviaU dataset. 
(a) False color image. 
(b) Ground-truth. 
(c) A2S2K. 
(d) DMSGer. 
(e) SSTN. 
(f) CTF-SSCL. 
(g) DEMAE. 
(h) RMAE. 
(i) Semi-HCN. 
(j) Ours.} 
% (k) Color labels.}

\label{fig:PaviaU_classification_maps}

\end{figure*}

\begin{figure*}[!t] 
\centering

%==================== 参数 ====================%
\newcommand{\colw}{0.089\textwidth}
\newcommand{\colsep}{0.003\textwidth}

% \hspace*{-0.8em}

%==================== 分类图 ====================%
\begin{minipage}[t]{\colw}\centering
    \rotatebox{-90}{\includegraphics[width=5\linewidth]{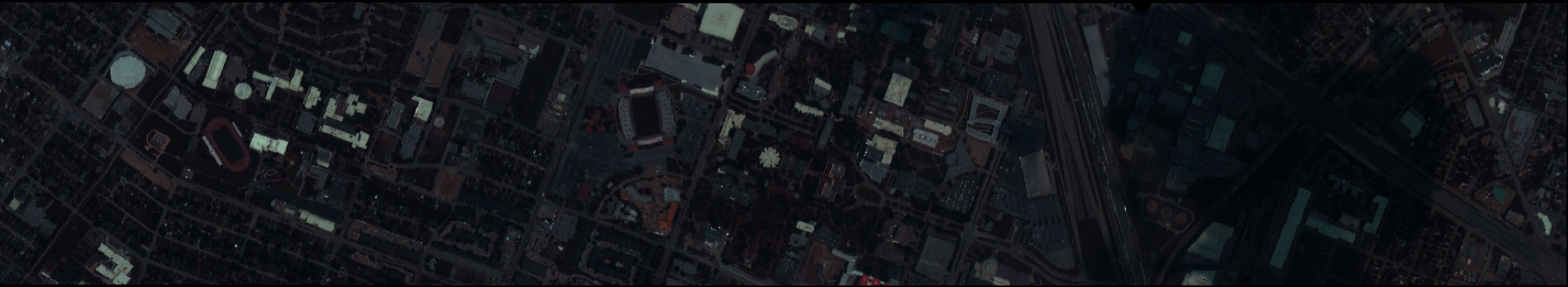}}\\[0.2em]
    \small(a)
\end{minipage}\hspace{\colsep}
\begin{minipage}[t]{\colw}\centering
    \rotatebox{-90}{\includegraphics[width=5\linewidth]{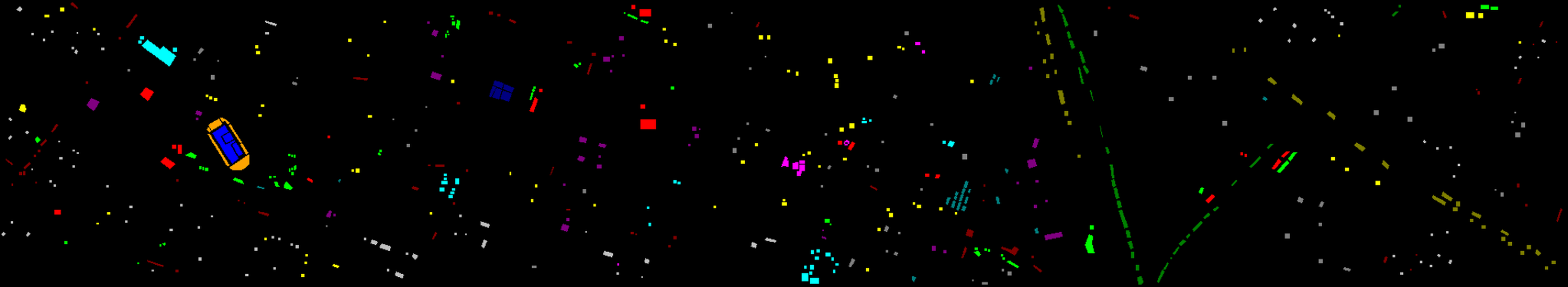}}\\[0.2em]
    \small(b)
\end{minipage}\hspace{\colsep}
\begin{minipage}[t]{\colw}\centering
    \rotatebox{-90}{\includegraphics[width=5\linewidth]{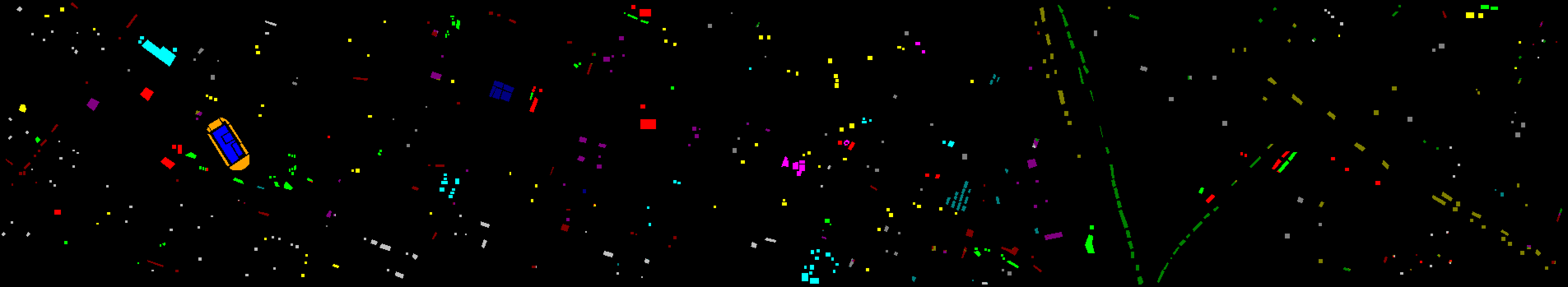}}\\[0.2em]
    \small(c)
\end{minipage}\hspace{\colsep}
\begin{minipage}[t]{\colw}\centering
    \rotatebox{-90}{\includegraphics[width=5\linewidth]{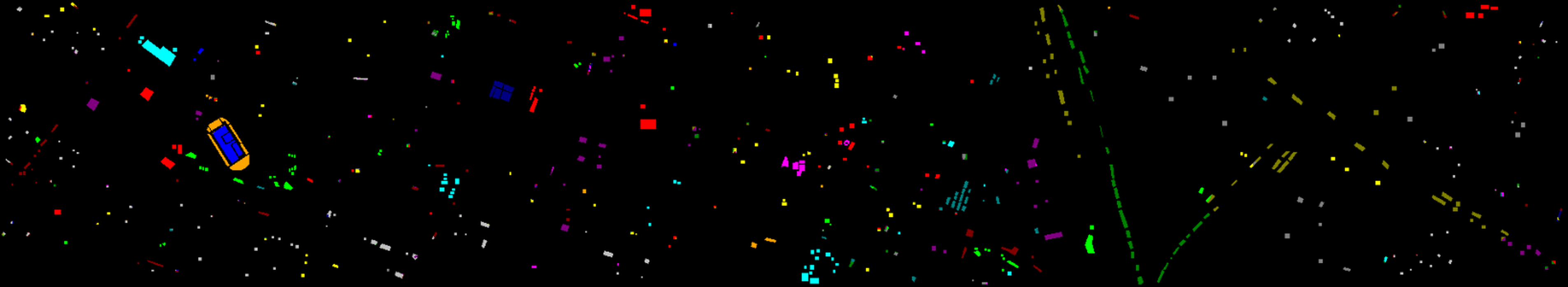}}\\[0.2em]
    \small(d)
\end{minipage}\hspace{\colsep}
\begin{minipage}[t]{\colw}\centering
    \rotatebox{-90}{\includegraphics[width=5\linewidth]{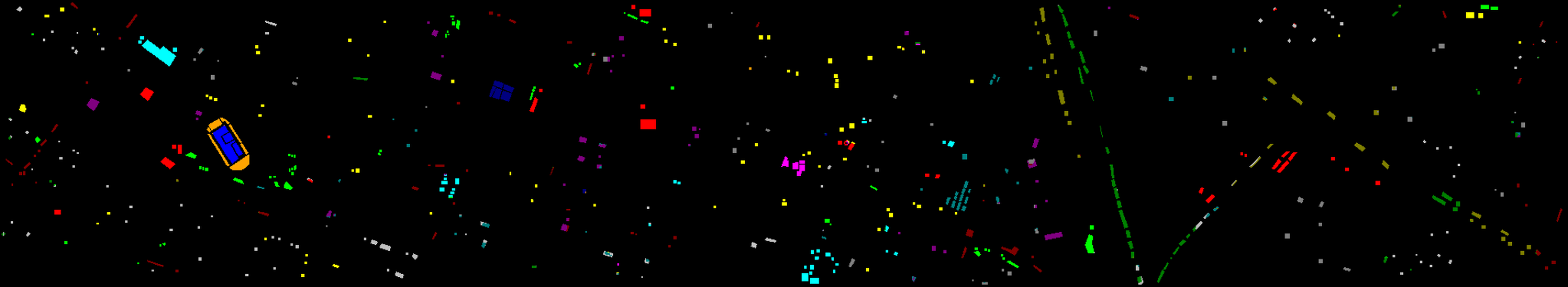}}\\[0.2em]
    \small(e)
\end{minipage}\hspace{\colsep}
\begin{minipage}[t]{\colw}\centering
    \rotatebox{-90}{\includegraphics[width=5\linewidth]{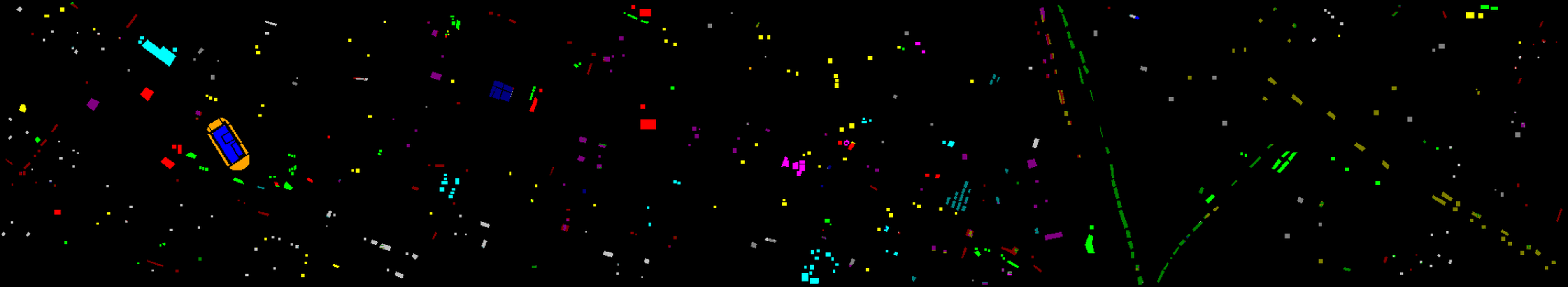}}\\[0.2em]
    \small(f)
\end{minipage}\hspace{\colsep}
\begin{minipage}[t]{\colw}\centering
    \rotatebox{-90}{\includegraphics[width=5\linewidth]{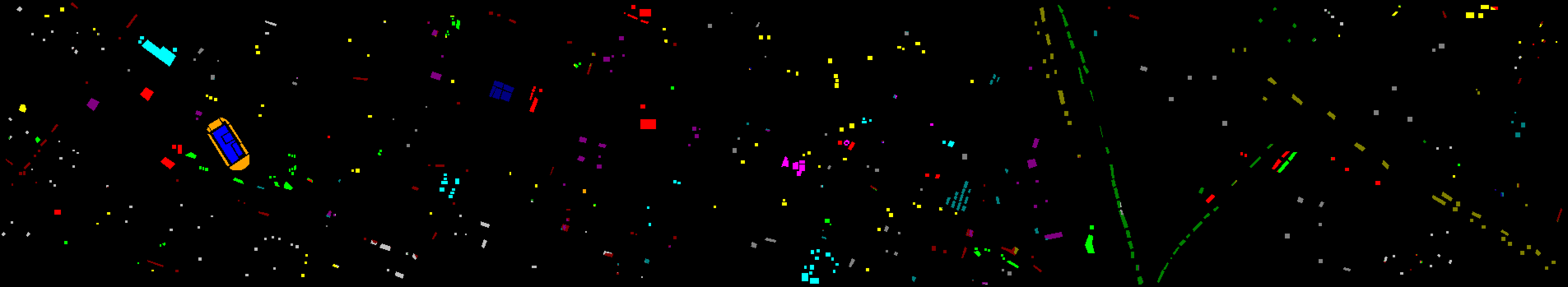}}\\[0.2em]
    \small(g)
\end{minipage}\hspace{\colsep}
\begin{minipage}[t]{\colw}\centering
    \rotatebox{-90}{\includegraphics[width=5\linewidth]{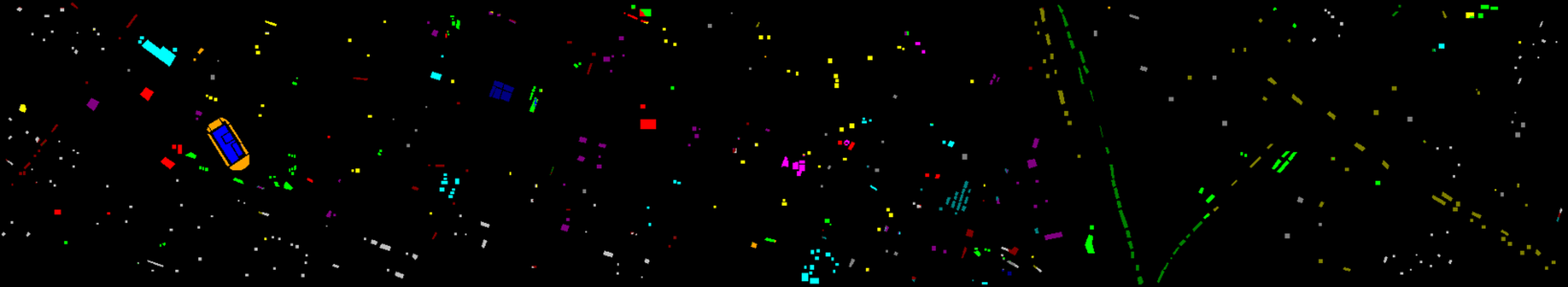}}\\[0.2em]
    \small(h)
\end{minipage}\hspace{\colsep}
\begin{minipage}[t]{\colw}\centering
    \rotatebox{-90}{\includegraphics[width=5\linewidth]{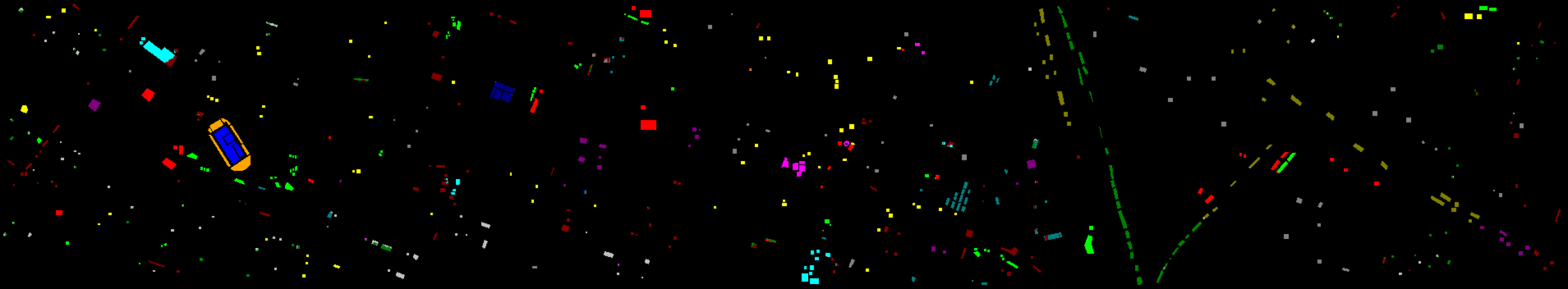}}\\[0.2em]
    \small(i)
\end{minipage}\hspace{\colsep}
\begin{minipage}[t]{\colw}\centering
    \rotatebox{-90}{\includegraphics[width=5\linewidth]{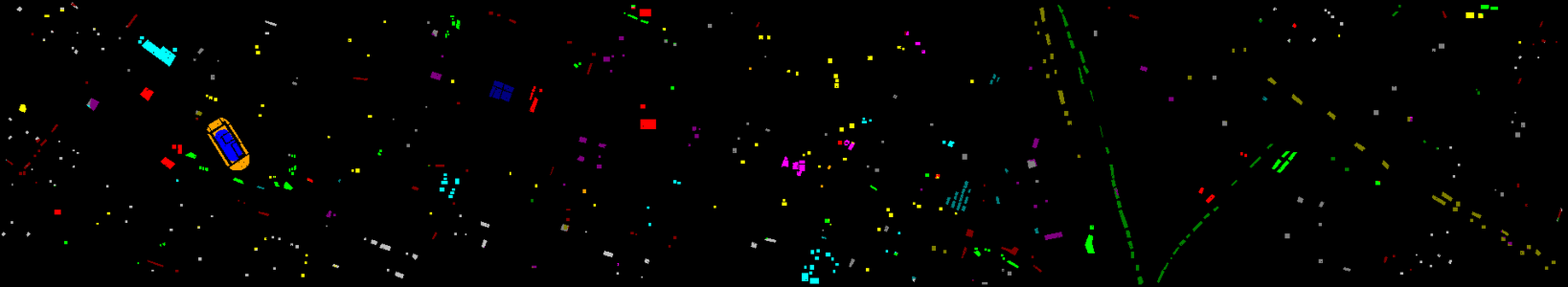}}\\[0.2em]
    \small(j)
\end{minipage}
%==================== 横向颜色条 ====================%
\vspace{0.1em}

% \centering
% \normalsize
% \scriptsize
% \setlength{\tabcolsep}{1pt}

% \begin{tabular}{cccccccccccccccc}

% \textcolor[rgb]{0.0,0.0,0.0}{\rule{4mm}{4mm}} &
% \textcolor[rgb]{1.0,0.0,0.0}{\rule{4mm}{4mm}} &
% \textcolor[rgb]{0.0,1.0,0.0}{\rule{4mm}{4mm}} &
% \textcolor[rgb]{0.0,0.0,1.0}{\rule{4mm}{4mm}} &
% \textcolor[rgb]{1.0,1.0,0.0}{\rule{4mm}{4mm}} &
% \textcolor[rgb]{0.0,1.0,1.0}{\rule{4mm}{4mm}} &
% \textcolor[rgb]{1.0,0.0,1.0}{\rule{4mm}{4mm}} &
% \textcolor[rgb]{0.75,0.75,0.75}{\rule{4mm}{4mm}} &
% \textcolor[rgb]{0.5,0.5,0.5}{\rule{4mm}{4mm}} &
% \textcolor[rgb]{0.5,0.0,0.0}{\rule{4mm}{4mm}} &
% \textcolor[rgb]{0.5,0.5,0.0}{\rule{4mm}{4mm}} &
% \textcolor[rgb]{0.0,0.5,0.0}{\rule{4mm}{4mm}} &
% \textcolor[rgb]{0.5,0.0,0.5}{\rule{4mm}{4mm}} &
% \textcolor[rgb]{0.0,0.5,0.5}{\rule{4mm}{4mm}} &
% \textcolor[rgb]{0.0,0.0,0.5}{\rule{4mm}{4mm}} &
% \textcolor[rgb]{1.0,0.65,0.0}{\rule{4mm}{4mm}} \\

% Unlabelled &
% Health grass &
% Stressed grass &
% Synthetic grass &
% Trees &
% Soil &
% Water &
% Residential &
% Commercial &
% Road &
% Highway &
% Railway &
% Parking lot 1 &
% Parking lot 2 &
% Tennis court &
% Running track \\

% \end{tabular}

\begin{center}
\normalsize
\begin{tikzpicture}[
every node/.style={
    minimum width=2cm,
    minimum height=0.7cm,
    font=\footnotesize,
    inner sep=0pt,
    outer sep=0pt,
    text width=2cm,
    align=center
}
]

% 第一行
\node[fill=black,text=white] at (2,0) {\textbf{Unlabelled}};
\node[fill=red,text=white] at (4,0) {\textbf{Health\\grass}};
\node[fill=green,text=black] at (6,0) {\textbf{Stressed\\grass}};
\node[fill=blue,text=white] at (8,0) {\textbf{Synthetic\\grass}};
\node[fill=yellow,text=black] at (10,0) {\textbf{Trees}};
\node[fill=cyan,text=black] at (12,0) {\textbf{Soil}};
\node[fill=magenta,text=white] at (14,0) {\textbf{Water}};
\node[fill={rgb,1:red,0.75;green,0.75;blue,0.75},text=black]
     at (16,0) {\textbf{Residential}};
% 第二行（与第一行无缝拼接）

\node[fill={rgb,1:red,0.5;green,0.5;blue,0.5},text=white]
     at (2,-0.7) {\textbf{Commercial}};

\node[fill={rgb,1:red,0.5;green,0;blue,0},text=white]
     at (4,-0.7) {\textbf{Road}};

\node[fill={rgb,1:red,0.5;green,0.5;blue,0},text=white]
     at (6,-0.7) {\textbf{Highway}};

\node[fill={rgb,1:red,0;green,0.5;blue,0},text=white]
     at (8,-0.7) {\textbf{Railway}};

\node[fill={rgb,1:red,0.5;green,0;blue,0.5},text=white]
     at (10,-0.7) {\textbf{Parking\\lot 1}};

\node[fill={rgb,1:red,0;green,0.5;blue,0.5},text=white]
     at (12,-0.7) {\textbf{Parking\\lot 2}};

\node[fill={rgb,1:red,0;green,0;blue,0.5},text=white]
     at (14,-0.7) {\textbf{Tennis\\court}};

\node[fill={rgb,1:red,1;green,0.65;blue,0},text=black]
     at (16,-0.7) {\textbf{Running\\track}};

\end{tikzpicture}
\end{center}

\vspace{0.1em}
% \tiny(k)

\caption{Classification maps for the Houston2013 dataset.
(a) False color image.
(b) Ground-truth.
(c) A2S2K.
(d) DMSGer.
(e) SSTN.
(f) CTF-SSCL.
(g) DEMAE.
(h) RMAE.
(i) Semi-HCN.
(j) Ours.}
% (k) Color labels.}

\label{fig:Houston_classification_maps}

\end{figure*}

%PaviaU
\begin{table*}[!t]
\centering
\caption{Classification results for the PaviaU dataset with 10 labeled samples per class.}
\label{tab:PaviaU_results}
\renewcommand{\arraystretch}{1.15}
\setlength{\tabcolsep}{3pt}
\resizebox{\linewidth}{!}{%
\begin{tabular}{c|cccccccc}
\hline
No. & A2S2K & DMSGer & SSTN & CTF-SSCL & DEMAE & RMAE & Semi-HCN & Ours \\
\hline
1 & 73.06 $\pm$ 6.43 & 69.71 $\pm$ 4.60 & 91.83 $\pm$ 1.26 & 91.95 $\pm$ 3.64 & 87.03 $\pm$ 5.36 & 79.18 $\pm$ 1.61 & \textbf{98.13 $\pm$ 1.38} & 88.28 $\pm$ 5.45 \\
2 & 88.29 $\pm$ 4.58 & 87.27 $\pm$ 5.47 & 79.59 $\pm$ 3.44 & 80.52 $\pm$ 8.72 & 90.98 $\pm$ 8.75 & 93.76 $\pm$ 2.22 & 88.38 $\pm$ 2.77 & \textbf{97.47 $\pm$ 1.59} \\
3 & 81.58 $\pm$ 2.14 & 91.46 $\pm$ 7.06 & 91.87 $\pm$ 4.35 & 84.32 $\pm$ 4.86 & 83.32 $\pm$ 9.11 & 90.37 $\pm$ 4.41 & 85.57 $\pm$ 13.12 & \textbf{94.04 $\pm$ 1.36} \\
4 & 96.80 $\pm$ 0.97 & 84.92 $\pm$ 5.49 & 79.13 $\pm$ 4.80 & 92.69 $\pm$ 3.42 & 94.55 $\pm$ 2.57 & \textbf{98.72 $\pm$ 0.49} & 94.13 $\pm$ 2.87 & 91.63 $\pm$ 4.43 \\
5 & \textbf{100.00 $\pm$ 0.00} & 99.80 $\pm$ 0.19 & 99.98 $\pm$ 0.03 & 99.81 $\pm$ 0.24 & 99.75 $\pm$ 0.29 & 99.42 $\pm$ 0.15 & 99.98 $\pm$ 0.05 & 98.15 $\pm$ 2.40 \\
6 & 91.27 $\pm$ 4.39 & 95.75 $\pm$ 1.55 & 97.36 $\pm$ 2.97 & 88.29 $\pm$ 8.39 & 95.33 $\pm$ 4.97 & \textbf{99.94 $\pm$ 0.06} & 94.63 $\pm$ 2.04 & 97.66 $\pm$ 2.77 \\
7 & 98.98 $\pm$ 1.11 & 98.88 $\pm$ 3.44 & 98.14 $\pm$ 4.04 & 97.66 $\pm$ 1.90 & 99.20 $\pm$ 0.72 & \textbf{99.97 $\pm$ 0.06} & 99.15 $\pm$ 0.76 & 98.36 $\pm$ 2.86 \\
8 & 60.79 $\pm$ 6.03 & 92.47 $\pm$ 3.59 & 83.91 $\pm$ 4.76 & 88.42 $\pm$ 6.91 & 85.53 $\pm$ 5.74 & 92.48 $\pm$ 0.79 & \textbf{98.10 $\pm$ 3.1} & 95.57 $\pm$ 3.07 \\
9 & 99.59 $\pm$ 0.32 & 98.85 $\pm$ 0.52 & 99.53 $\pm$ 0.40 & 98.32 $\pm$ 1.18 & 98.09 $\pm$ 1.14 & 94.13 $\pm$ 0.65 & \textbf{99.74 $\pm$ 0.23} & 91.47 $\pm$ 7.74 \\
\hline
\textbf{OA (\%)} & 86.94 $\pm$ 1.32 & 87.00 $\pm$ 2.62 & 86.15 $\pm$ 1.26 & 86.47 $\pm$ 4.24 & 90.98 $\pm$ 3.87 & 92.68 $\pm$ 0.73 & 91.74 $\pm$ 1.03 & \textbf{95.21 $\pm$ 1.17} \\
\textbf{AA (\%)} & 87.26 $\pm$ 1.28 & 90.90 $\pm$ 1.43 & 91.26 $\pm$ 0.92 & 91.33 $\pm$ 1.94 & 92.64 $\pm$ 1.53 & 94.22 $\pm$ 0.32 & 94.07 $\pm$ 1.26 & \textbf{94.46 $\pm$ 1.48} \\
\textbf{$\kappa\times100$} & 85.30 $\pm$ 1.48 & 83.30 $\pm$ 1.17 & 82.27 $\pm$ 1.54 & 82.74 $\pm$ 5.11 & 88.32 $\pm$ 4.78 & 90.46 $\pm$ 0.90 & 89.23 $\pm$ 1.31 & \textbf{93.67 $\pm$ 1.54} \\
\hline
\end{tabular}%
}
\end{table*}

%Huston
\begin{table*}[!t]
\centering
\caption{Classification results for the Houston2013 dataset with 10 labeled samples per class.}
\label{tab:Houston2013_results}
\renewcommand{\arraystretch}{1.15}
\setlength{\tabcolsep}{3pt}
\resizebox{\textwidth}{!}{%
\begin{tabular}{c|cccccccc}
\hline
No. & A2S2K & DMSGer & SSTN & CTF-SSCL & DEMAE & RMAE & Semi-HCN & Ours \\
\hline
1  & 84.69 $\pm$ 1.14 & 85.37 $\pm$ 8.00 & \textbf{97.13 $\pm$ 2.71} & 93.17 $\pm$ 5.76 & 90.09 $\pm$ 13.03 & 76.51 $\pm$ 5.24 & 94.22 $\pm$ 8.24 & 86.67 $\pm$ 6.13 \\
2  & \textbf{97.99 $\pm$ 1.09} & 78.91 $\pm$ 4.96 & 91.77 $\pm$ 5.21 & 89.55 $\pm$ 7.96 & 87.52 $\pm$ 7.22 & 80.55 $\pm$ 3.69 & 94.25 $\pm$ 6.07 & 88.64 $\pm$ 4.60 \\
3  & \textbf{99.94 $\pm$ 0.07} & 97.51 $\pm$ 0.77 & 99.46 $\pm$ 1.03 & 98.92 $\pm$ 1.33 & 99.23 $\pm$ 0.58 & 99.91 $\pm$ 0.12 & 97.92 $\pm$ 3.87 & 95.79 $\pm$ 4.52 \\
4  & 92.88 $\pm$ 0.94 & 66.65 $\pm$ 6.03 & 94.02 $\pm$ 2.55 & \textbf{95.44 $\pm$ 2.93} & 92.01 $\pm$ 5.53 & 74.02 $\pm$ 2.57 & 93.28 $\pm$ 1.27 & 87.60 $\pm$ 5.00 \\
5  & \textbf{100.00 $\pm$ 0.00} & 99.86 $\pm$ 0.22 & 99.15 $\pm$ 0.73 & 98.69 $\pm$ 1.47 & 99.89 $\pm$ 0.25 & 99.55 $\pm$ 0.92 & 85.81 $\pm$ 8.69 & 99.33 $\pm$ 0.83 \\
6  & 98.95 $\pm$ 0.94 & \textbf{99.91 $\pm$ 0.29} & 83.93 $\pm$ 2.94 & 90.79 $\pm$ 6.17 & 94.83 $\pm$ 5.75 & 92.16 $\pm$ 3.16 & 92.56 $\pm$ 6.90 & 89.91 $\pm$ 5.74 \\
7  & 78.86 $\pm$ 3.62 & 71.87 $\pm$ 3.41 & 77.49 $\pm$ 5.55 & 79.17 $\pm$ 6.23 & 80.14 $\pm$ 11.03 & 85.37 $\pm$ 1.06 & 75.06 $\pm$ 13.88 & \textbf{87.84 $\pm$ 3.26} \\
8  & 56.13 $\pm$ 2.64 & 51.26 $\pm$ 4.51 & 53.64 $\pm$ 6.40 & 62.75 $\pm$ 7.70 & 67.06 $\pm$ 7.95 & 67.58 $\pm$ 5.25 & \textbf{69.68 $\pm$ 12.08} & 69.55 $\pm$ 8.87 \\
9  & 76.89 $\pm$ 4.16 & 70.79 $\pm$ 5.16 & 84.75 $\pm$ 4.00 & 77.79 $\pm$ 6.68 & 78.86 $\pm$ 5.75 & 59.51 $\pm$ 3.73 & 79.16 $\pm$ 9.16 & \textbf{88.80 $\pm$ 3.56} \\
10 & 77.18 $\pm$ 5.01 & 89.31 $\pm$ 4.42 & 86.45 $\pm$ 8.24 & 88.05 $\pm$ 10.09 & 93.45 $\pm$ 9.22 & \textbf{99.52 $\pm$ 0.61} & 83.11 $\pm$ 4.66 & 98.54 $\pm$ 1.53 \\
11 & 84.57 $\pm$ 5.41 & 75.14 $\pm$ 6.20 & 87.68 $\pm$ 6.28 & 79.97 $\pm$ 12.42 & 85.82 $\pm$ 14.58 & 86.16 $\pm$ 2.07 & 81.79 $\pm$ 3.82 & \textbf{90.73 $\pm$ 7.21} \\
12 & 75.65 $\pm$ 5.88 & 86.23 $\pm$ 3.45 & 51.14 $\pm$ 10.19 & 83.18 $\pm$ 10.32 & 77.45 $\pm$ 8.00 & 80.10 $\pm$ 4.82 & 72.61 $\pm$ 4.96 & \textbf{87.46 $\pm$ 4.24} \\
13 & 91.85 $\pm$ 0.75 & 89.85 $\pm$ 2.57 & 82.57 $\pm$ 6.82 & 92.92 $\pm$ 4.47 & 87.63 $\pm$ 7.22 & 73.83 $\pm$ 3.85 & 76.17 $\pm$ 7.45 & \textbf{95.95 $\pm$ 2.27} \\
14 & \textbf{100.00 $\pm$ 0.00} & \textbf{100.00 $\pm$ 0.00} & 99.77 $\pm$ 0.46 & 99.45 $\pm$ 0.89 & \textbf{100.00 $\pm$ 0.00} & 99.67 $\pm$ 0.62 & 97.99 $\pm$ 5.62 & 99.04 $\pm$ 1.98 \\
15 & \textbf{100.00 $\pm$ 0.00} & \textbf{100.00 $\pm$ 0.00} & \textbf{100.00 $\pm$ 0.00} & 99.57 $\pm$ 0.82 & \textbf{100.00 $\pm$ 0.00} & 99.71 $\pm$ 0.55 & 99.00 $\pm$ 2.95 & 98.28 $\pm$ 2.48 \\
\hline
\textbf{OA (\%)} & 85.21 $\pm$ 0.97 & 80.93 $\pm$ 1.24 & 84.41 $\pm$ 0.84 & 86.86 $\pm$ 1.79 & 87.20 $\pm$ 2.41 & 83.10 $\pm$ 0.64 & 84.70 $\pm$ 4.77 & \textbf{89.77 $\pm$ 1.17} \\
\textbf{AA (\%)} & 87.71 $\pm$ 0.77 & 84.20 $\pm$ 1.03 & 85.93 $\pm$ 0.82 & 88.63 $\pm$ 1.66 & 88.93 $\pm$ 2.26 & 84.94 $\pm$ 0.43 & 84.77 $\pm$ 4.27 & \textbf{90.87 $\pm$ 0.98} \\
\textbf{$\kappa\times100$} & 84.02 $\pm$ 1.05 & 79.40 $\pm$ 1.34 & 83.15 $\pm$ 0.90 & 85.80 $\pm$ 1.93 & 86.17 $\pm$ 2.60 & 81.73 $\pm$ 0.69 & 83.46 $\pm$ 5.16 & \textbf{88.94 $\pm$ 1.26} \\
\hline
\end{tabular}%
}
\end{table*}

\begin{figure*}[!htbp]
\centering

\newcommand{\colw}{0.18\linewidth}
\newcommand{\colsep}{-0.01\linewidth}

%==================== 第一行 ====================%
\begin{minipage}[t]{\colw}\centering
    \includegraphics[width=0.7\linewidth]{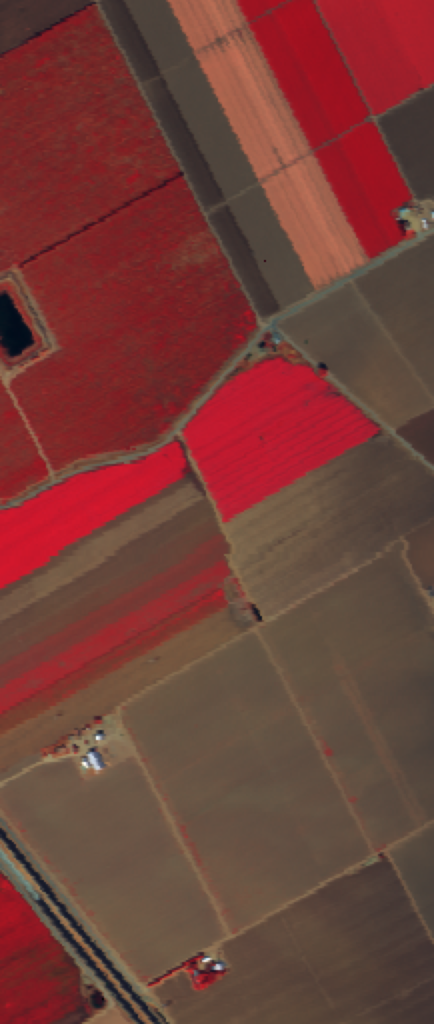}\\[0.2em]
    \scriptsize(a)
\end{minipage}\hspace{\colsep}
\begin{minipage}[t]{\colw}\centering
    \includegraphics[width=0.7\linewidth]{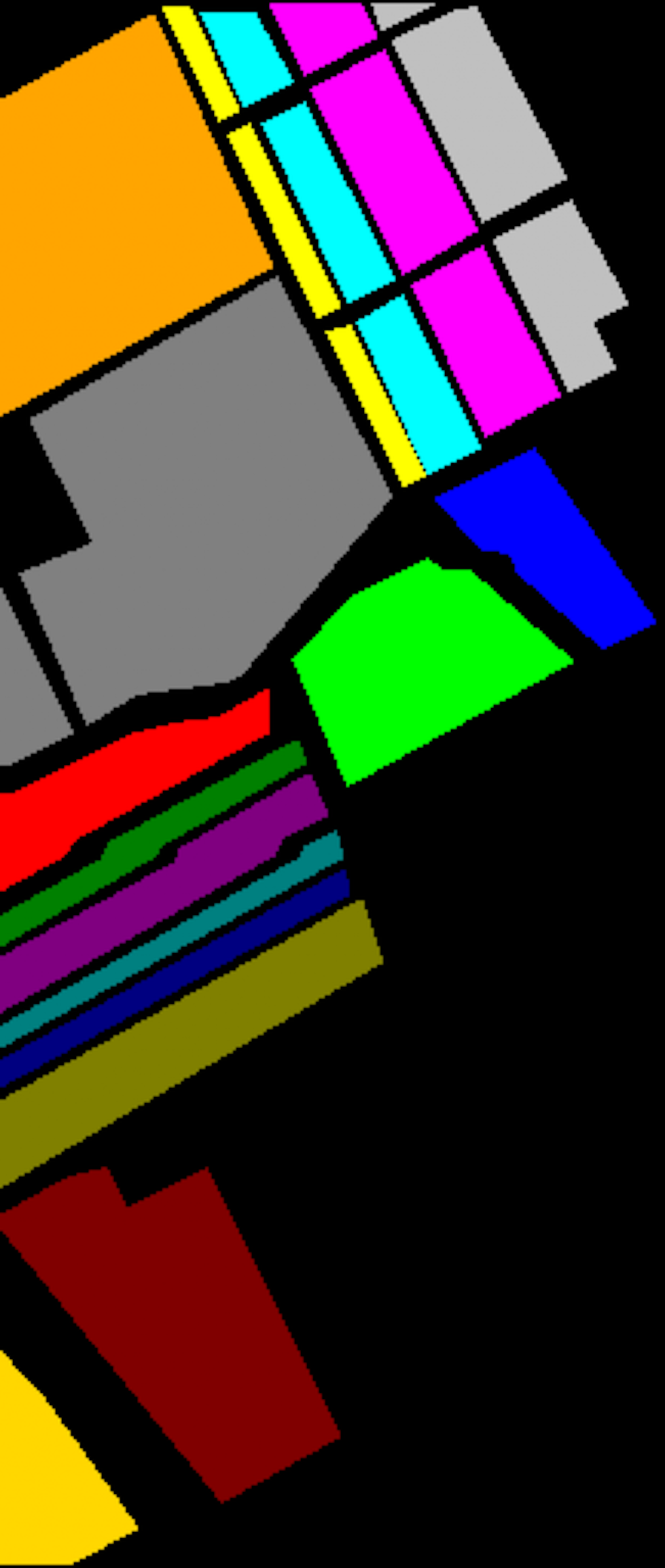}\\[0.2em]
    \scriptsize(b)
\end{minipage}\hspace{\colsep}
\begin{minipage}[t]{\colw}\centering
    \includegraphics[width=0.7\linewidth]{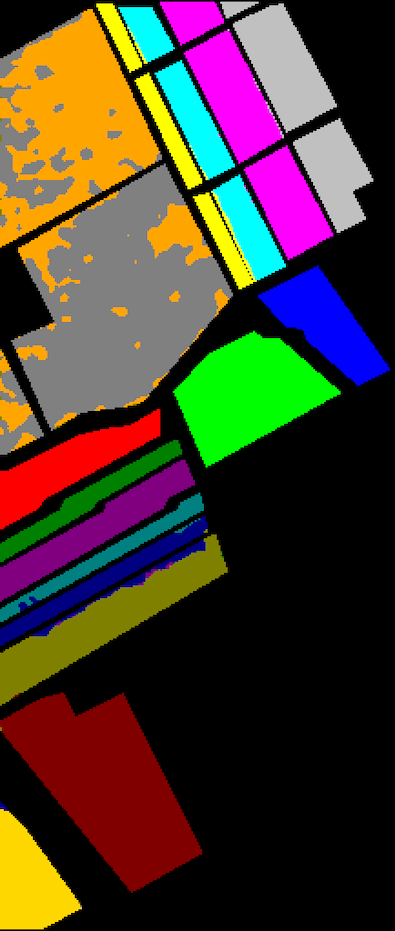}\\[0.2em]
    \scriptsize(c)
\end{minipage}\hspace{\colsep}
\begin{minipage}[t]{\colw}\centering
    \includegraphics[width=0.7\linewidth]{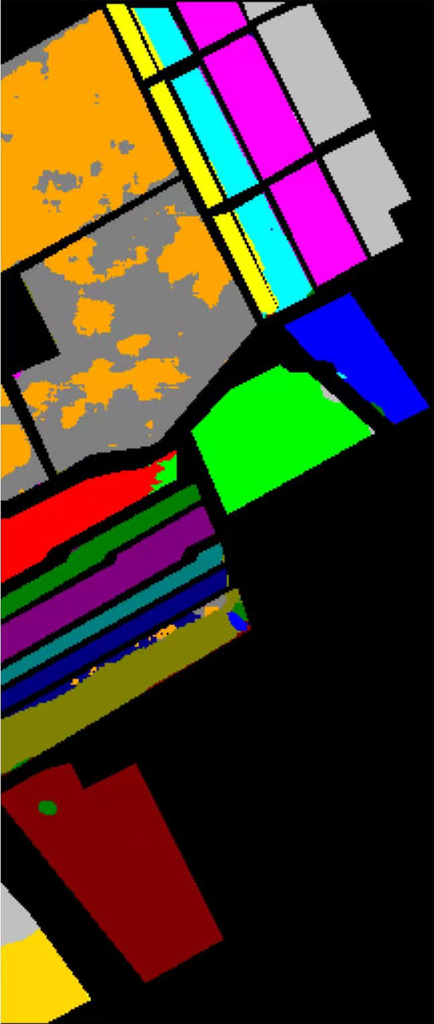}\\[0.2em]
    \scriptsize(d)
\end{minipage}\hspace{\colsep}
\begin{minipage}[t]{\colw}\centering
    \includegraphics[width=0.7\linewidth]{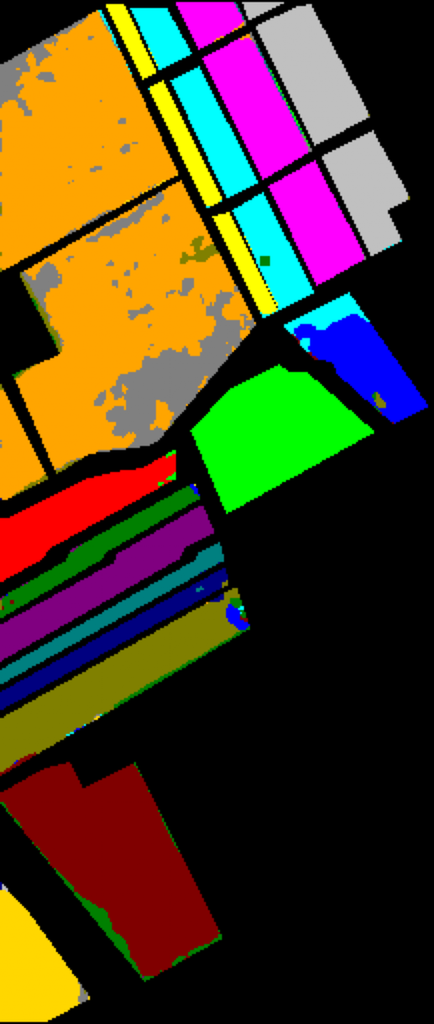}\\[0.2em]
    \scriptsize(e)
\end{minipage}

\vspace{0.8em}

%==================== 第二行 ====================%
\begin{minipage}[t]{\colw}\centering
    \includegraphics[width=0.7\linewidth]{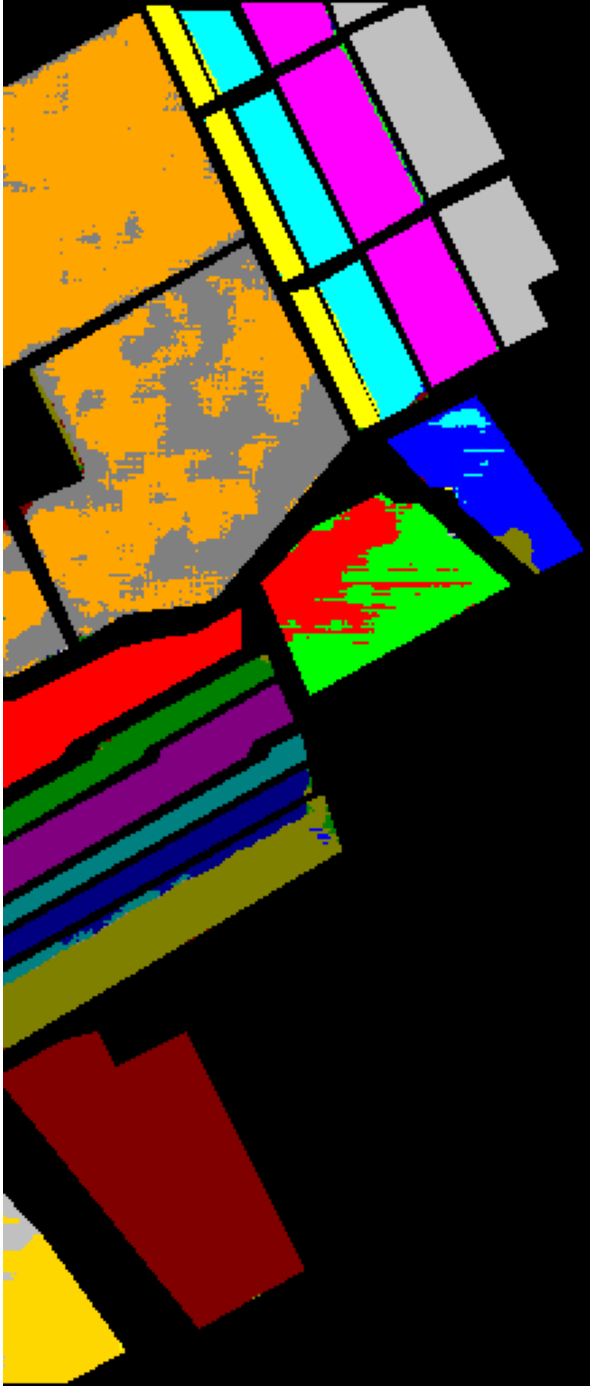}\\[0.2em]
    \scriptsize(f)
\end{minipage}\hspace{\colsep}
\begin{minipage}[t]{\colw}\centering
    \includegraphics[width=0.7\linewidth]{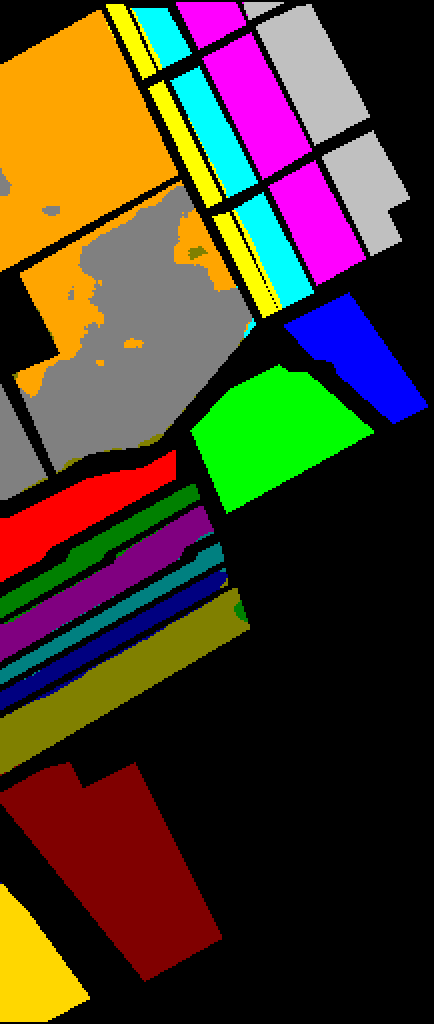}\\[0.2em]
    \scriptsize(g)
\end{minipage}\hspace{\colsep}
\begin{minipage}[t]{\colw}\centering
    \includegraphics[width=0.7\linewidth]{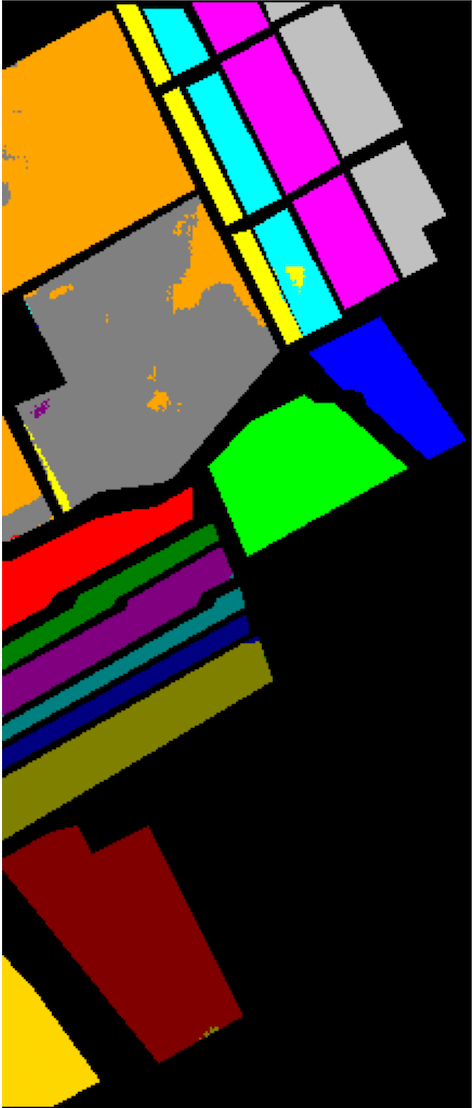}\\[0.2em]
    \scriptsize(h)
\end{minipage}\hspace{\colsep}
\begin{minipage}[t]{\colw}\centering
    \includegraphics[width=0.7\linewidth]{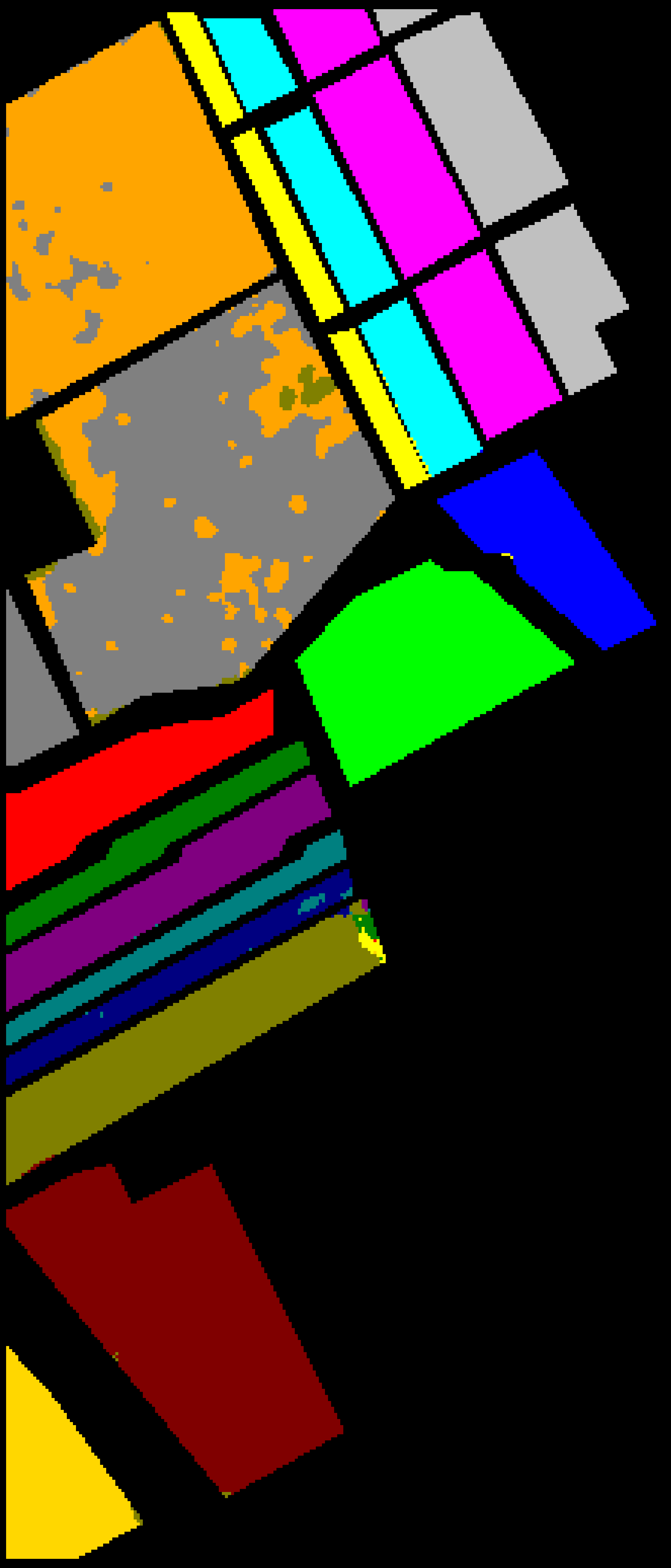}\\[0.2em]
    \scriptsize(i)
\end{minipage}\hspace{\colsep}
\begin{minipage}[t]{\colw}\centering
    \includegraphics[width=0.7\linewidth]{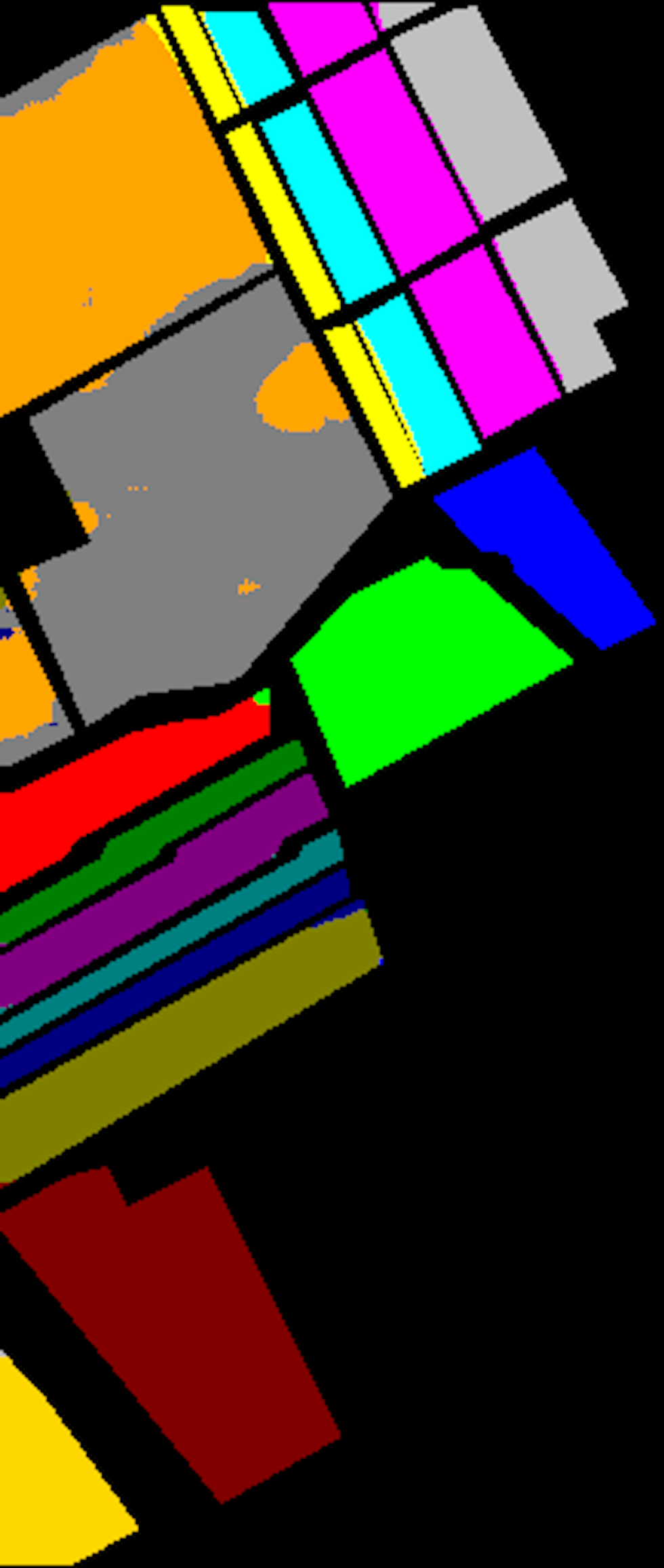}\\[0.2em]
    \scriptsize(j)
\end{minipage}

\vspace{0.1em}

%==================== 横向颜色条 ====================%
% \centering

% \scriptsize

% % 第一行更紧凑
% \setlength{\tabcolsep}{3pt}

% \begin{tabular}{ccccccccc}

% % 第一行颜色块
% \textcolor[rgb]{0.0,0.0,0.0}{\rule{4mm}{4mm}} &
% \textcolor[rgb]{1.0,0.0,0.0}{\rule{4mm}{4mm}} &
% \textcolor[rgb]{0.0,1.0,0.0}{\rule{4mm}{4mm}} &
% \textcolor[rgb]{0.0,0.0,1.0}{\rule{4mm}{4mm}} &
% \textcolor[rgb]{1.0,1.0,0.0}{\rule{4mm}{4mm}} &
% \textcolor[rgb]{0.0,1.0,1.0}{\rule{4mm}{4mm}} &
% \textcolor[rgb]{1.0,0.0,1.0}{\rule{4mm}{4mm}} &
% \textcolor[rgb]{0.75,0.75,0.75}{\rule{4mm}{4mm}} &
% \textcolor[rgb]{0.5,0.5,0.5}{\rule{4mm}{4mm}}  \\

% % 第一行文字
% Unlabelled &
% Brocoli green weeds 1 &
% Brocoli green weeds 2 &
% Fallow &
% Fallow rough plow &
% Fallow smooth &
% Stubble &
% Celery &
% Grapes untrained \\[2mm]

% \end{tabular}

% \vspace{1mm}

% % 第二行保持原样
% \begin{tabular}{cccccccc}

% \textcolor[rgb]{0.5,0.0,0.0}{\rule{4mm}{4mm}} &
% \textcolor[rgb]{0.5,0.5,0.0}{\rule{4mm}{4mm}} &
% \textcolor[rgb]{0.0,0.5,0.0}{\rule{4mm}{4mm}} &
% \textcolor[rgb]{0.5,0.0,0.5}{\rule{4mm}{4mm}} &
% \textcolor[rgb]{0.0,0.5,0.5}{\rule{4mm}{4mm}} &
% \textcolor[rgb]{0.0,0.0,0.5}{\rule{4mm}{4mm}} &
% \textcolor[rgb]{1.0,0.65,0.0}{\rule{4mm}{4mm}} &
% \textcolor[rgb]{1.0,0.84,0.0}{\rule{4mm}{4mm}} \\

% Soil develop &
% Corn weeds &
% Lettuce 4wk &
% Lettuce 5wk &
% Lettuce 6wk &
% Lettuce 7wk &
% Vinyard untrained &
% Vinyard trellis

% \end{tabular}

\begin{center}
\begin{tikzpicture}[
every node/.style={
    minimum width=2.2cm,
    minimum height=0.7cm,
    font=\footnotesize,
    inner sep=0pt,
    outer sep=0pt,
    align=center,
    text width=2.2cm
}
]

% -------------------------
% 第一行颜色块
\node[fill={rgb,1:red,0.0;green,0.0;blue,0.0},text=white] at (1.4,0) {\textbf{Unlabelled}};
\node[fill={rgb,1:red,1.0;green,0.0;blue,0.0},text=white] at (3.6,0) {\textbf{Brocoli\\green weeds1}};
\node[fill={rgb,1:red,0.0;green,1.0;blue,0.0},text=black] at (5.8,0) {\textbf{Brocoli\\green weeds2}};
\node[fill={rgb,1:red,0.0;green,0.0;blue,1.0},text=white] at (8,0) {\textbf{Fallow}};
\node[fill={rgb,1:red,1.0;green,1.0;blue,0.0},text=black] at (10.2,0) {\textbf{Fallow\\rough plow}};
\node[fill={rgb,1:red,0.0;green,1.0;blue,1.0},text=black] at (12.4,0) {\textbf{Fallow\\smooth}};
\node[fill={rgb,1:red,1.0;green,0.0;blue,1.0},text=white] at (14.6,0) {\textbf{Stubble}};
\node[fill={rgb,1:red,0.75;green,0.75;blue,0.75},text=black] at (16.8,0) {\textbf{Celery}};
% \node[fill={rgb,1:red,0.5;green,0.5;blue,0.5},text=white] at (17.4,0) {\textbf{Grapes\\untrained}};
\end{tikzpicture}
\begin{tikzpicture}[
every node/.style={
    minimum width=2cm,
    minimum height=0.7cm,
    font=\footnotesize,
    inner sep=0pt,
    outer sep=0pt,
    align=center,
    text width=2cm
}
]
% -------------------------
% 第二行颜色块（紧贴第一行）
\node[fill={rgb,1:red,0.5;green,0.5;blue,0.5},text=white] at (1.4,-0.7) {\textbf{Grapes\\untrained}};
\node[fill={rgb,1:red,0.5;green,0.0;blue,0.0},text=white] at (3.4,-0.7) {\textbf{Soil develop}};
\node[fill={rgb,1:red,0.5;green,0.5;blue,0.0},text=white] at (5.4,-0.7) {\textbf{Corn weeds}};
\node[fill={rgb,1:red,0.0;green,0.5;blue,0.0},text=white] at (7.4,-0.7) {\textbf{Lettuce 4wk}};
\node[fill={rgb,1:red,0.5;green,0.0;blue,0.5},text=white] at (9.4,-0.7) {\textbf{Lettuce 5wk}};
\node[fill={rgb,1:red,0.0;green,0.5;blue,0.5},text=white] at (11.4,-0.7){\textbf {Lettuce 6wk}};
\node[fill={rgb,1:red,0.0;green,0.0;blue,0.5},text=white] at (13.4,-0.7) {\textbf{Lettuce 7wk}};
\node[fill={rgb,1:red,1.0;green,0.65;blue,0.0},text=black] at (15.4,-0.7) {\textbf{Vinyard untrained}};
\node[fill={rgb,1:red,1.0;green,0.84;blue,0.0},text=black] at (17.4,-0.7) {\textbf{Vinyard trellis}};

\end{tikzpicture}
\end{center}
\vspace{0.1em}

% \scriptsize(k)

\caption{Classification maps for the Salinas dataset. 
(a) False color image. 
(b) Ground-truth. 
(c) A2S2K. 
(d) DMSGer. 
(e) SSTN. 
(f) CTF-SSCL. 
(g) DEMAE. 
(h) RMAE. 
(i) Semi-HCN. 
(j) Ours. }
% (k) Color labels.}

\label{fig:Salinas_classification_maps}

\end{figure*}

%Salinas
\begin{table*}[!t]
\centering
\caption{Classification results for the Salinas dataset with 10 labeled samples per class.}
\label{tab:Salinas_results}
\renewcommand{\arraystretch}{1.15}
\setlength{\tabcolsep}{3pt}
\resizebox{\textwidth}{!}{%
\begin{tabular}{c|cccccccc}
\hline
No. & A2S2K & DMSGer & SSTN & CTF-SSCL & DEMAE & RMAE & Semi-HCN & Ours \\
\hline
1  & \textbf{100 $\pm$ 0.00} & 99.38 $\pm$ 0.84 & 99.75 $\pm$ 0.36 & 98.01 $\pm$ 2.00 & \textbf{100 $\pm$ 0.00} & 99.73 $\pm$ 0.22 & 98.00 $\pm$ 2.41 & 99.65 $\pm$ 0.83 \\
2  & 99.91 $\pm$ 0.11 & 91.61 $\pm$ 1.62 & 97.81 $\pm$ 3.08 & 97.42 $\pm$ 2.63 & \textbf{100 $\pm$ 0.00} & \textbf{100 $\pm$ 0.00} & 99.60 $\pm$ 0.98 & 99.94 $\pm$ 0.17 \\
3  & \textbf{100 $\pm$ 0.00} & \textbf{100 $\pm$ 0.00} & 86.02 $\pm$ 4.75 & 91.36 $\pm$ 7.14 & \textbf{100 $\pm$ 0.00} & \textbf{100 $\pm$ 0.00} & \textbf{100 $\pm$ 0.00} & 99.79 $\pm$ 0.59 \\
4  & \textbf{100 $\pm$ 0.00} & 99.35 $\pm$ 0.42 & 99.49 $\pm$ 0.32 & 98.64 $\pm$ 1.03 & 99.88 $\pm$ 0.17 & 99.65 $\pm$ 0.23 & 98.70 $\pm$ 2.36 & 99.27 $\pm$ 1.13 \\
5  & 95.08 $\pm$ 0.47 & 90.76 $\pm$ 2.64 & 95.93 $\pm$ 0.84 & 94.54 $\pm$ 3.96 & 94.57 $\pm$ 2.45 & 96.98 $\pm$ 0.61 & \textbf{99.45 $\pm$ 0.75} & 98.09 $\pm$ 1.36 \\
6  & 99.86 $\pm$ 0.10 & 99.19 $\pm$ 1.08 & 97.56 $\pm$ 0.52 & 99.68 $\pm$ 0.56 & 98.91 $\pm$ 1.49 & \textbf{100 $\pm$ 0.00} & 99.94 $\pm$ 0.13 & 98.80 $\pm$ 1.67 \\
7  & \textbf{100 $\pm$ 0.00} & 98.11 $\pm$ 2.74 & 99.86 $\pm$ 0.06 & 97.73 $\pm$ 2.92 & 99.85 $\pm$ 0.17 & 95.10 $\pm$ 2.17 & 99.80 $\pm$ 0.29 & 99.91 $\pm$ 0.22 \\
8  & 80.83 $\pm$ 0.72 & 87.15 $\pm$ 4.17 & 84.50 $\pm$ 1.90 & 74.68 $\pm$ 7.15 & 73.84 $\pm$ 9.10 & 90.02 $\pm$ 1.20 & 87.50 $\pm$ 5.42 & \textbf{91.56 $\pm$ 3.23} \\
9  & 99.75 $\pm$ 0.29 & 95.00 $\pm$ 1.97 & 98.12 $\pm$ 1.67 & 98.62 $\pm$ 1.54 & \textbf{100 $\pm$ 0.00} & 98.16 $\pm$ 1.82 & 99.86 $\pm$ 0.22 & 99.99 $\pm$ 0.02 \\
10 & 90.08 $\pm$ 2.29 & 93.48 $\pm$ 1.53 & 89.23 $\pm$ 1.38 & 83.98 $\pm$ 6.85 & 97.49 $\pm$ 1.81 & 96.66 $\pm$ 2.05 & 94.60 $\pm$ 1.58 & \textbf{98.15 $\pm$ 3.37} \\
11 & 99.96 $\pm$ 0.04 & 98.86 $\pm$ 0.51 & 97.78 $\pm$ 0.44 & 93.97 $\pm$ 2.75 & \textbf{100 $\pm$ 0.00} & \textbf{100 $\pm$ 0.00} & 93.52 $\pm$ 11.35 & 99.28 $\pm$ 0.91 \\
12 & 98.97 $\pm$ 0.04 & 91.59 $\pm$ 2.88 & 94.39 $\pm$ 5.91 & 96.10 $\pm$ 4.95 & 98.44 $\pm$ 1.74 & 99.80 $\pm$ 0.10 & \textbf{99.85 $\pm$ 0.36} & 99.32 $\pm$ 0.68 \\
13 & 98.36 $\pm$ 1.84 & 97.03 $\pm$ 2.12 & 99.85 $\pm$ 0.21 & 97.72 $\pm$ 3.00 & 99.48 $\pm$ 0.73 & \textbf{99.97 $\pm$ 0.05} & 97.40 $\pm$ 5.94 & 99.66 $\pm$ 0.56 \\
14 & 96.57 $\pm$ 2.09 & 98.57 $\pm$ 1.06 & 93.43 $\pm$ 3.38 & 97.45 $\pm$ 2.23 & 99.43 $\pm$ 0.80 & \textbf{99.93 $\pm$ 0.17} & 83.81 $\pm$ 2.24 & 98.03 $\pm$ 2.46 \\
15 & 84.08 $\pm$ 1.83 & 78.92 $\pm$ 5.35 & 56.34 $\pm$ 7.32 & 71.49 $\pm$ 2.23 & \textbf{97.10 $\pm$ 1.94} & 87.35 $\pm$ 1.57 & 76.88 $\pm$ 11.60 & 92.94 $\pm$ 2.98 \\
16 & 98.58 $\pm$ 1.41 & 99.50 $\pm$ 0.67 & 96.68 $\pm$ 1.66 & 91.44 $\pm$ 5.30 & 99.83 $\pm$ 0.09 & \textbf{100 $\pm$ 0.00} & 98.86 $\pm$ 1.94 & 99.83 $\pm$ 0.27 \\
\hline
\textbf{OA (\%)} & 96.31 $\pm$ 0.42 & 91.84 $\pm$ 0.79 & 88.44 $\pm$ 1.87 & 88.08 $\pm$ 1.93 & 93.56 $\pm$ 1.63 & 95.30 $\pm$ 0.37 & 93.22 $\pm$ 0.94 & \textbf{96.85 $\pm$ 0.35} \\
\textbf{AA (\%)} & 96.37 $\pm$ 0.45 & 94.90 $\pm$ 0.40 & 92.92 $\pm$ 1.46 & 92.68 $\pm$ 1.27 & 97.43 $\pm$ 0.44 & 97.71 $\pm$ 0.19 & 95.48 $\pm$ 2.09 & \textbf{98.38 $\pm$ 0.23} \\
\textbf{$k$ (\%)} & 96.06 $\pm$ 0.45 & 90.95 $\pm$ 0.88 & 87.11 $\pm$ 2.08 & 86.76 $\pm$ 2.13 & 92.86 $\pm$ 1.79 & 94.77 $\pm$ 0.41 & 92.44 $\pm$ 1.05 & \textbf{96.50 $\pm$ 0.39} \\
\hline
\end{tabular}%
}
\end{table*}

%KSC
\begin{table*}[!t]
\centering
\caption{Classification results for the KSC dataset with 10 labeled samples per class.}
\label{tab:KSC_results}
\renewcommand{\arraystretch}{1.15}
\setlength{\tabcolsep}{3pt}
\resizebox{\textwidth}{!}{%
\begin{tabular}{c|cccccccc}
\hline
No. & A2S2K & DMSGer & SSTN & CTF-SSCL & DEMAE & RMAE & Semi-HCN & Ours \\
\hline
1  & 97.64 $\pm$ 2.06 & 93.59 $\pm$ 6.36 & 99.41 $\pm$ 0.54 & 98.12 $\pm$ 1.01 & 99.61 $\pm$ 0.80 & 89.32 $\pm$ 5.00 & 98.74 $\pm$ 0.56 & \textbf{99.84 $\pm$ 0.19} \\
2  & 98.88 $\pm$ 0.95 & 78.45 $\pm$ 0.26 & 81.13 $\pm$ 6.46 & 97.08 $\pm$ 1.88 & \textbf{99.87 $\pm$ 0.27} & 96.61 $\pm$ 2.42 & 98.43 $\pm$ 1.86 & 99.32 $\pm$ 1.52 \\
3  & 99.58 $\pm$ 0.16 & 98.90 $\pm$ 0.13 & 98.67 $\pm$ 2.32 & 97.80 $\pm$ 1.95 & 99.43 $\pm$ 1.71 & 91.02 $\pm$ 2.54 & 98.64 $\pm$ 1.86 & \textbf{99.95 $\pm$ 0.15} \\
4  & 91.77 $\pm$ 4.46 & \textbf{100.00 $\pm$ 0.00} & 87.97 $\pm$ 8.52 & 76.94 $\pm$ 12.44 & 94.96 $\pm$ 6.94 & 77.69 $\pm$ 7.11 & 95.43 $\pm$ 7.99 & 99.06 $\pm$ 1.15 \\
5  & 98.01 $\pm$ 2.22 & \textbf{100.00 $\pm$ 0.00} & 82.52 $\pm$ 12.16 & 91.32 $\pm$ 4.90 & 99.80 $\pm$ 0.30 & 92.58 $\pm$ 1.70 & 99.93 $\pm$ 0.21 & 96.10 $\pm$ 2.82 \\
6  & 99.43 $\pm$ 0.63 & \textbf{100.00 $\pm$ 0.00} & 96.13 $\pm$ 4.72 & 98.17 $\pm$ 4.30 & \textbf{100.00 $\pm$ 0.00} & 86.53 $\pm$ 3.49 & \textbf{100.00 $\pm$ 0.00} & \textbf{100.00 $\pm$ 0.00} \\
7  & \textbf{100.00 $\pm$ 0.00} & \textbf{100.00 $\pm$ 0.00} & 99.07 $\pm$ 1.69 & 96.63 $\pm$ 10.11 & 99.58 $\pm$ 1.26 & 97.85 $\pm$ 1.14 & 96.71 $\pm$ 6.48 & \textbf{100.00 $\pm$ 0.00} \\
8  & 90.67 $\pm$ 3.15 & 94.61 $\pm$ 0.42 & 92.19 $\pm$ 2.97 & 95.80 $\pm$ 3.23 & 98.24 $\pm$ 2.13 & 83.66 $\pm$ 2.78 & \textbf{99.95 $\pm$ 0.10} & 99.54 $\pm$ 0.77 \\
9  & 98.20 $\pm$ 2.02 & 83.53 $\pm$ 6.20 & 98.71 $\pm$ 2.15 & 98.45 $\pm$ 4.33 & 98.47 $\pm$ 4.33 & 96.96 $\pm$ 1.24 & \textbf{100.00 $\pm$ 0.00} & 99.76 $\pm$ 0.48 \\
10 & \textbf{100.00 $\pm$ 0.00} & \textbf{100.00 $\pm$ 0.00} & 99.28 $\pm$ 1.32 & 97.66 $\pm$ 1.53 & \textbf{100.00 $\pm$ 0.00} & 96.40 $\pm$ 3.17 & \textbf{100.00 $\pm$ 0.00} & \textbf{100.00 $\pm$ 0.00} \\
11 & 99.77 $\pm$ 0.34 & \textbf{100.00 $\pm$ 0.00} & 99.46 $\pm$ 0.61 & 98.80 $\pm$ 1.77 & 98.51 $\pm$ 2.03 & 99.88 $\pm$ 0.25 & 98.05 $\pm$ 2.00 & 99.93 $\pm$ 0.20 \\
12 & 99.05 $\pm$ 1.75 & 87.67 $\pm$ 0.18 & 95.41 $\pm$ 5.15 & 95.84 $\pm$ 1.77 & 99.84 $\pm$ 0.32 & 97.83 $\pm$ 1.48 & 92.92 $\pm$ 3.35 & \textbf{100.00 $\pm$ 0.00} \\
13 & \textbf{100.00 $\pm$ 0.00} & \textbf{100.00 $\pm$ 0.00} & 99.84 $\pm$ 0.47 & \textbf{100.00 $\pm$ 0.00} & \textbf{100.00 $\pm$ 0.00} & \textbf{100.00 $\pm$ 0.00} & \textbf{100.00 $\pm$ 0.00} & \textbf{100.00 $\pm$ 0.00} \\
\hline
\textbf{OA (\%)} & 98.04 $\pm$ 0.59 & 94.70 $\pm$ 1.08 & 96.46 $\pm$ 1.02 & 96.80 $\pm$ 0.98 & 99.22 $\pm$ 0.58 & 93.81 $\pm$ 1.05 & 98.55 $\pm$ 0.39 & \textbf{99.71 $\pm$ 0.17} \\
\textbf{AA (\%)} & 97.92 $\pm$ 0.59 & 95.12 $\pm$ 0.64 & 94.61 $\pm$ 1.45 & 95.59 $\pm$ 0.99 & 99.10 $\pm$ 0.71 & 92.96 $\pm$ 0.87 & 98.37 $\pm$ 0.62 & \textbf{99.48 $\pm$ 0.30} \\
\textbf{$\kappa\times100$} & 97.81 $\pm$ 0.66 & 94.11 $\pm$ 1.19 & 96.04 $\pm$ 1.14 & 96.43 $\pm$ 1.09 & 99.13 $\pm$ 0.64 & 93.11 $\pm$ 1.16 & 98.38 $\pm$ 0.43 & \textbf{99.68 $\pm$ 0.19} \\
\hline
\end{tabular}%
}
\end{table*}

%Botswana

\begin{table*}[!t]
\centering
\caption{Classification results for the Botswana dataset with 10 labeled samples per class.}
\label{tab:Botswana_results}
\renewcommand{\arraystretch}{1.15}
\setlength{\tabcolsep}{3pt}
\resizebox{\linewidth}{!}{%
\begin{tabular}{c|cccccccc}
\hline
No. & A2S2K & DMSGer & SSTN & CTF-SSCL & DEMAE & RMAE & Semi-HCN & Ours \\
\hline
1  & 99.04 $\pm$ 1.56 & 96.65 $\pm$ 2.86 & 99.21 $\pm$ 0.51 & 99.96 $\pm$ 0.12 & 99.31 $\pm$ 1.60 & \textbf{100.00 $\pm$ 0.00} & 99.92 $\pm$ 0.24 & 98.54 $\pm$ 1.59 \\
2  & \textbf{100.00 $\pm$ 0.00} & \textbf{100.00 $\pm$ 0.00} & \textbf{100.00 $\pm$ 0.00} & 99.56 $\pm$ 1.01 & \textbf{100.00 $\pm$ 0.00} & \textbf{100.00 $\pm$ 0.00} & 99.75 $\pm$ 0.74 & \textbf{100.00 $\pm$ 0.00} \\
3  & \textbf{100.00 $\pm$ 0.00} & 98.80 $\pm$ 0.46 & \textbf{100.00 $\pm$ 0.00} & 99.54 $\pm$ 0.51 & 94.85 $\pm$ 10.29 & 99.17 $\pm$ 1.26 & 96.62 $\pm$ 6.17 & 99.84 $\pm$ 0.29 \\
4  & \textbf{100.00 $\pm$ 0.00} & \textbf{100.00 $\pm$ 0.00} & \textbf{100.00 $\pm$ 0.00} & 99.56 $\pm$ 0.88 & \textbf{100.00 $\pm$ 0.00} & \textbf{100.00 $\pm$ 0.00} & 99.90 $\pm$ 0.31 & \textbf{100.00 $\pm$ 0.00} \\
5  & 91.73 $\pm$ 1.30 & 93.90 $\pm$ 0.24 & \textbf{99.16 $\pm$ 0.65} & 90.93 $\pm$ 4.81 & 96.06 $\pm$ 2.87 & 98.30 $\pm$ 2.02 & 97.43 $\pm$ 2.87 & 95.18 $\pm$ 3.92 \\
6  & 82.41 $\pm$ 9.82 & 93.86 $\pm$ 0.31 & 96.82 $\pm$ 0.86 & 98.11 $\pm$ 1.56 & 98.07 $\pm$ 2.25 & 94.90 $\pm$ 1.02 & 98.23 $\pm$ 1.45 & \textbf{99.81 $\pm$ 0.42} \\
7  & \textbf{100.00 $\pm$ 0.00} & 99.92 $\pm$ 0.07 & \textbf{100.00 $\pm$ 0.00} & 98.84 $\pm$ 2.98 & \textbf{100.00 $\pm$ 0.00} & \textbf{100.00 $\pm$ 0.00} & 99.96 $\pm$ 0.13 & 99.64 $\pm$ 0.77 \\
8  & 99.23 $\pm$ 0.95 & 99.90 $\pm$ 0.37 & \textbf{100.00 $\pm$ 0.00} & 94.87 $\pm$ 10.28 & 98.55 $\pm$ 4.02 & \textbf{100.00 $\pm$ 0.00} & \textbf{100.00 $\pm$ 0.00} & 99.58 $\pm$ 0.78 \\
9  & 91.77 $\pm$ 6.24 & \textbf{99.97 $\pm$ 0.60} & 99.93 $\pm$ 0.21 & 97.14 $\pm$ 4.08 & 99.01 $\pm$ 2.96 & 95.99 $\pm$ 3.18 & 99.22 $\pm$ 2.35 & \textbf{99.97 $\pm$ 0.10} \\
10 & \textbf{100.00 $\pm$ 0.00} & \textbf{100.00 $\pm$ 0.00} & \textbf{100.00 $\pm$ 0.00} & 99.16 $\pm$ 1.50 & \textbf{100.00 $\pm$ 0.00} & \textbf{100.00 $\pm$ 0.00} & \textbf{100.00 $\pm$ 0.00} & \textbf{100.00 $\pm$ 0.00} \\
11 & 99.54 $\pm$ 0.57 & \textbf{100.00 $\pm$ 0.00} & \textbf{100.00 $\pm$ 0.00} & 99.46 $\pm$ 0.71 & \textbf{100.00 $\pm$ 0.00} & \textbf{100.00 $\pm$ 0.00} & 83.12 $\pm$ 8.80 & 99.49 $\pm$ 1.18 \\
12 & 99.94 $\pm$ 0.19 & 79.83 $\pm$ 6.99 & \textbf{100.00 $\pm$ 0.00} & 99.88 $\pm$ 0.35 & 99.88 $\pm$ 0.23 & 99.53 $\pm$ 0.68 & \textbf{100.00 $\pm$ 0.00} & \textbf{100.00 $\pm$ 0.00} \\
13 & 99.44 $\pm$ 0.81 & \textbf{100.00 $\pm$ 0.00} & 99.71 $\pm$ 0.50 & 99.57 $\pm$ 0.91 & 98.37 $\pm$ 4.28 & 98.53 $\pm$ 0.62 & 99.40 $\pm$ 0.83 & \textbf{100.00 $\pm$ 0.00} \\
14 & 99.33 $\pm$ 1.23 & \textbf{100.00 $\pm$ 0.00} & 83.54 $\pm$ 5.42 & 99.41 $\pm$ 0.79 & \textbf{100.00 $\pm$ 0.00} & \textbf{100.00 $\pm$ 0.00} & 99.87 $\pm$ 0.40 & 98.12 $\pm$ 5.95 \\
\hline
\textbf{OA (\%)} & 96.78 $\pm$ 0.89 & 96.67 $\pm$ 0.42 & 99.11 $\pm$ 0.14 & 98.14 $\pm$ 0.83 & 98.73 $\pm$ 1.57 & 98.83 $\pm$ 0.35 & 97.60 $\pm$ 0.69 & \textbf{99.19 $\pm$ 0.40} \\
\textbf{AA (\%)} & 97.32 $\pm$ 0.74 & 97.33 $\pm$ 0.36 & 98.45 $\pm$ 0.38 & 98.29 $\pm$ 0.82 & 98.87 $\pm$ 1.42 & 99.03 $\pm$ 0.29 & 98.10 $\pm$ 0.58 & \textbf{99.22 $\pm$ 0.49} \\
\textbf{$\kappa\times100$} & 96.50 $\pm$ 0.96 & 96.39 $\pm$ 0.46 & 99.03 $\pm$ 0.15 & 97.98 $\pm$ 0.90 & 98.62 $\pm$ 1.70 & 98.73 $\pm$ 0.38 & 97.40 $\pm$ 0.75 & \textbf{99.12 $\pm$ 0.44} \\
\hline
\end{tabular}%
}
\end{table*}

\begin{figure*}[!t]
    \centering
    
    \newcommand{\colw}{0.18\textwidth}
    \newcommand{\colsep}{0.001\textwidth}

    %==================== 第一行 ====================%
    % \hspace*{-4.5em}
    \begin{minipage}[t]{\colw}\centering
        \includegraphics[width=\linewidth]{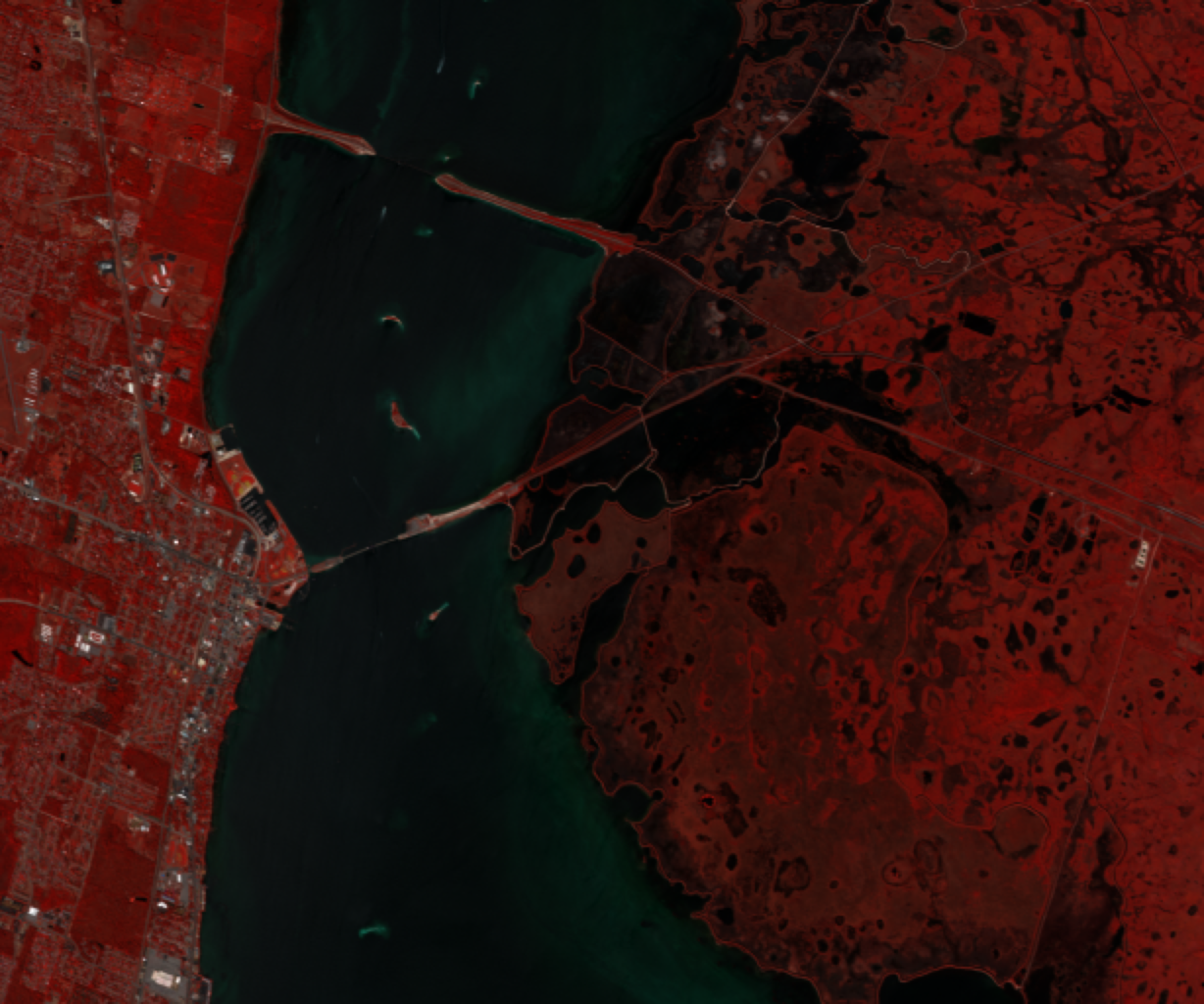}\\[0.2em]
        \scriptsize(a)
    \end{minipage}\hspace{\colsep}
    \begin{minipage}[t]{\colw}\centering
        \includegraphics[width=\linewidth]{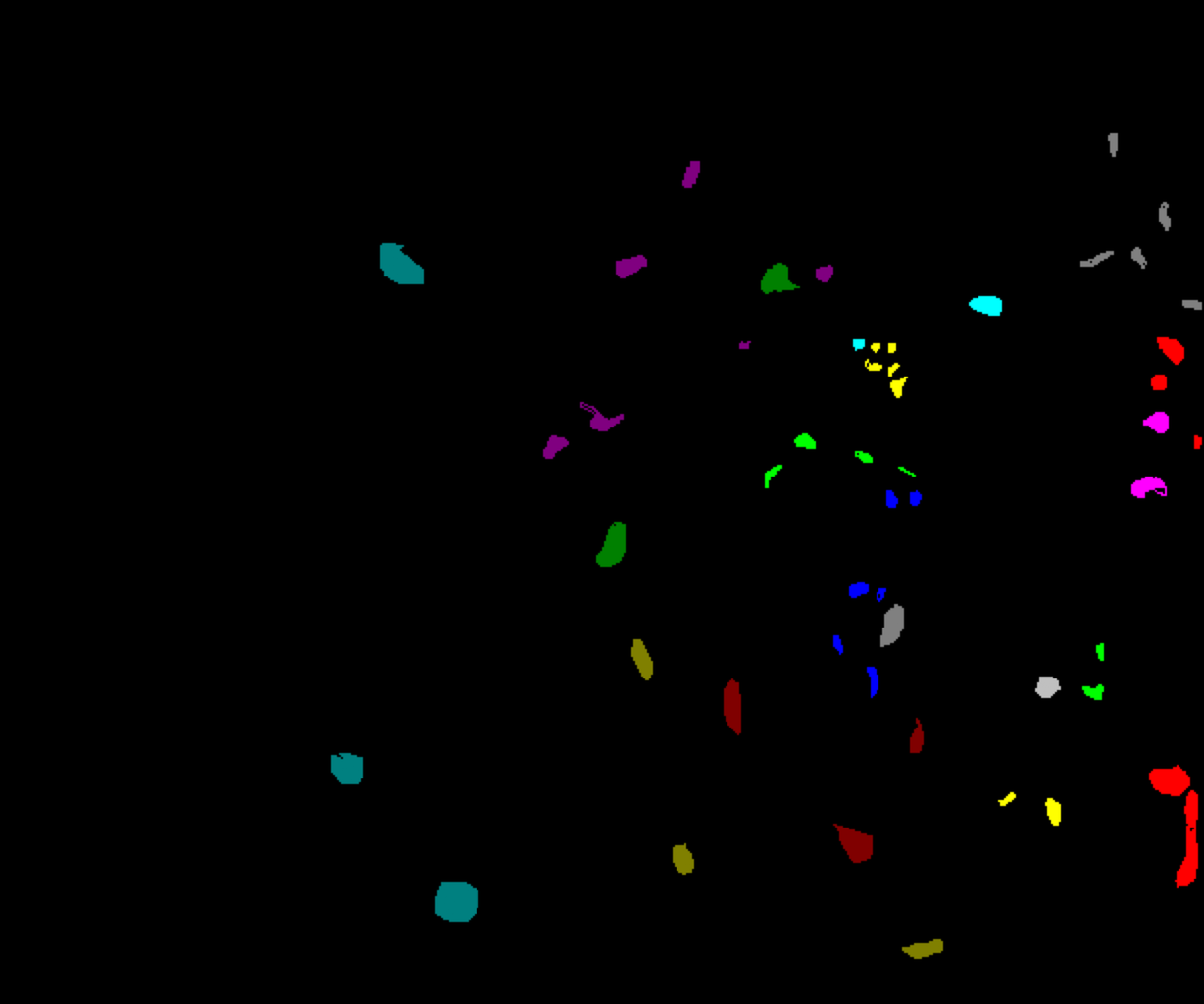}\\[0.2em]
        \scriptsize(b)
    \end{minipage}\hspace{\colsep}
    \begin{minipage}[t]{\colw}\centering
        \includegraphics[width=\linewidth]{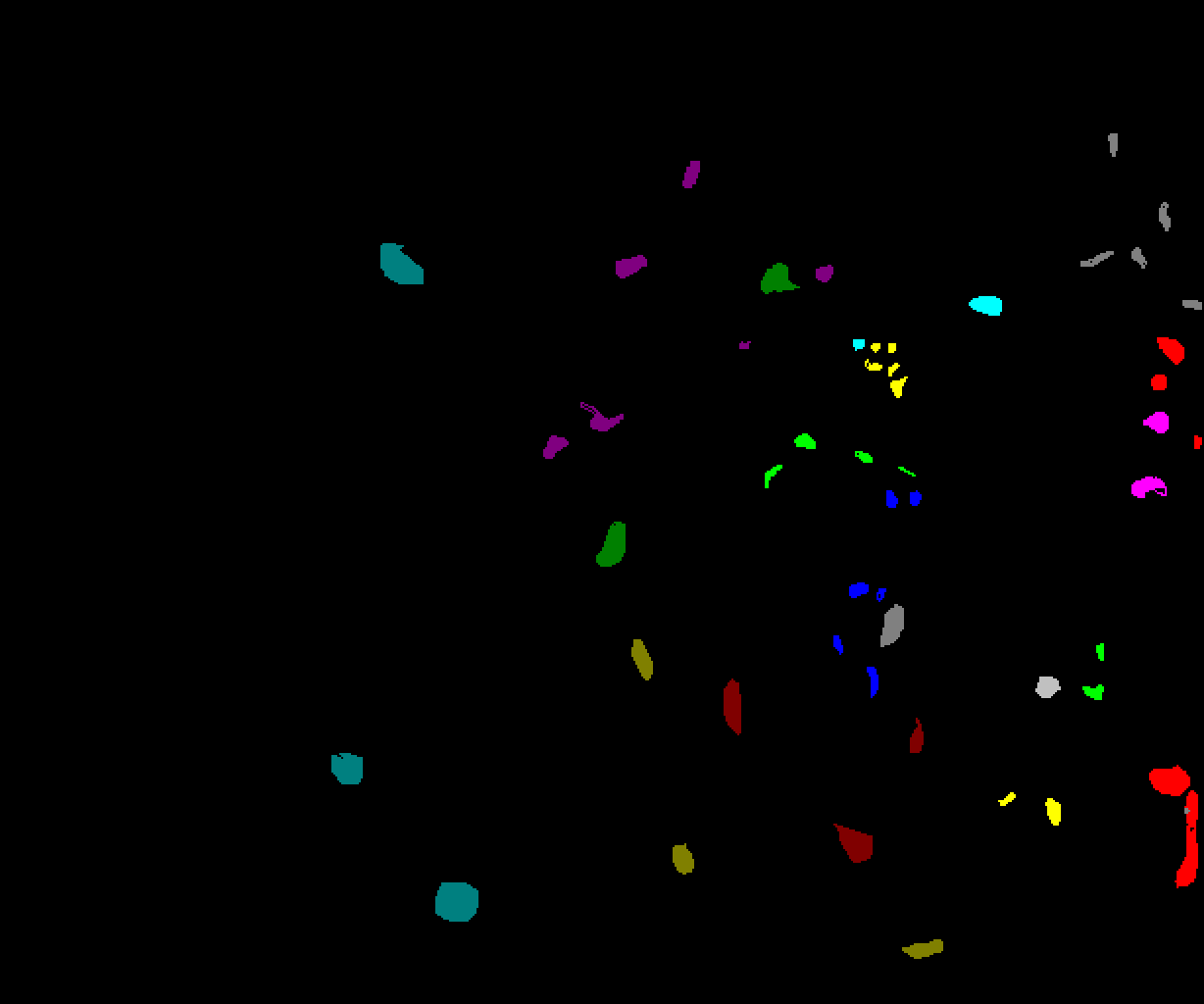}\\[0.2em]
        \scriptsize(c)
    \end{minipage}\hspace{\colsep}
    \begin{minipage}[t]{\colw}\centering
        \includegraphics[width=\linewidth]{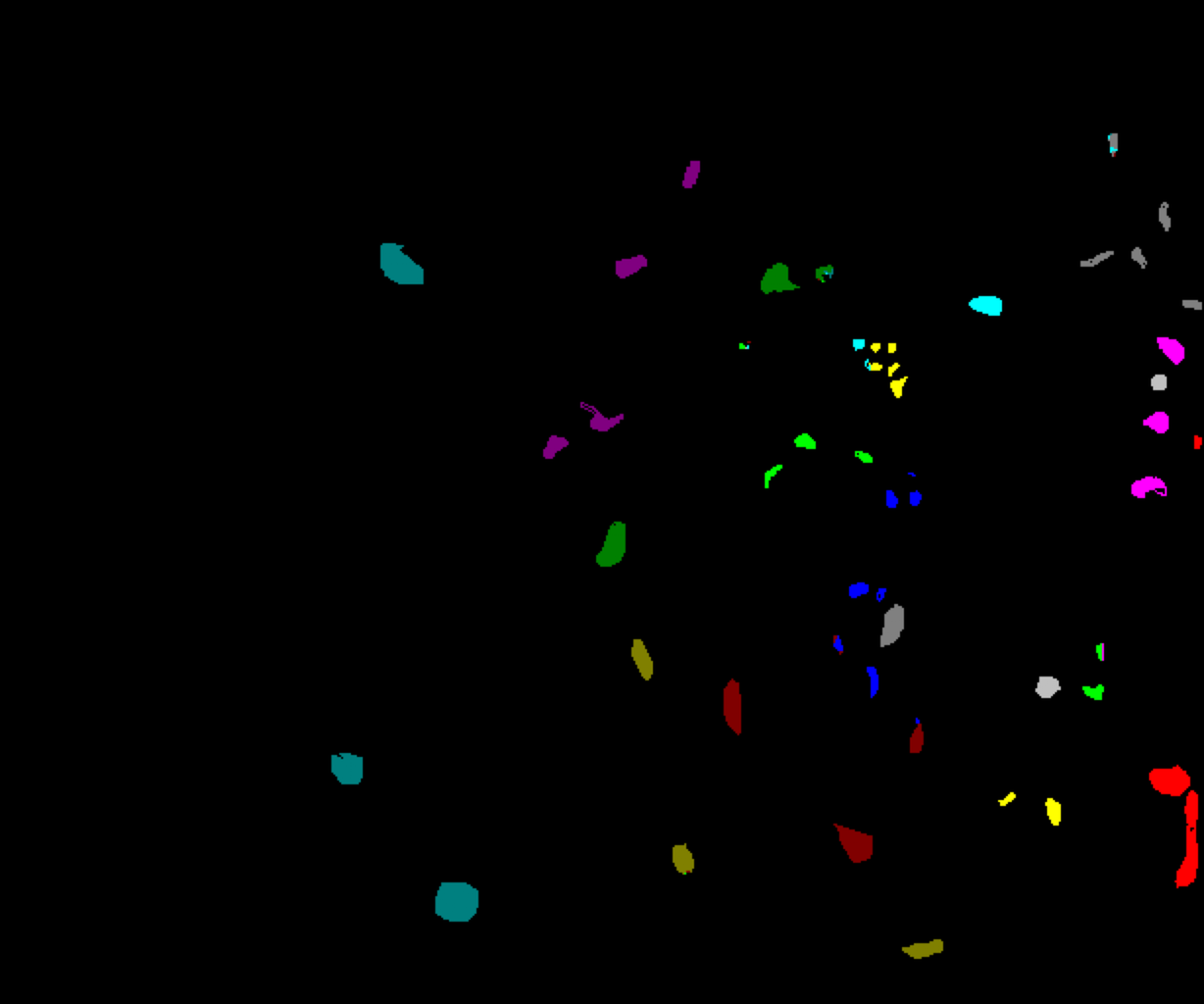}\\[0.2em]
        \scriptsize(d)
    \end{minipage}\hspace{\colsep}
    \begin{minipage}[t]{\colw}\centering
        \includegraphics[width=\linewidth]{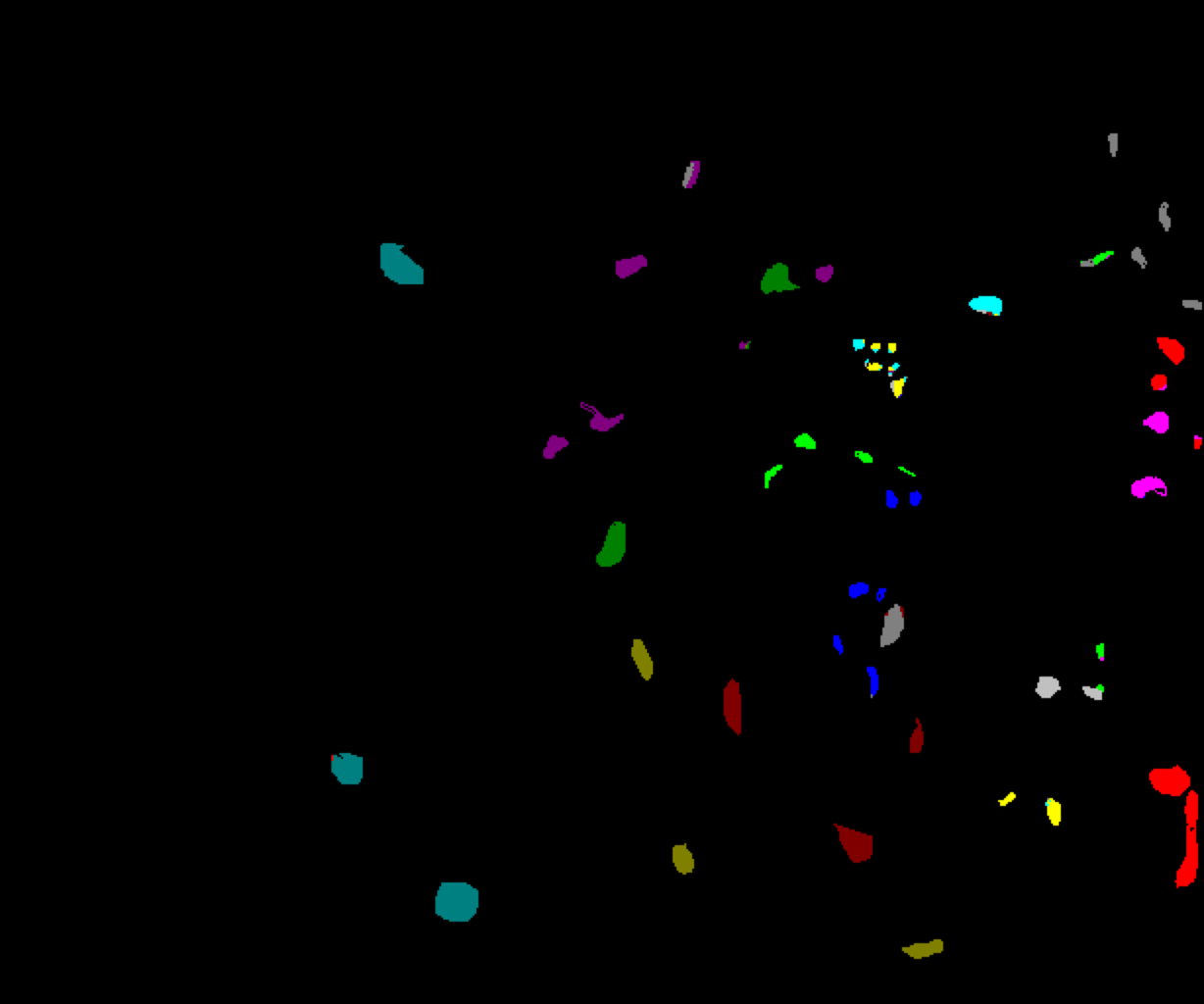}\\[0.2em]
        \scriptsize(e)
    \end{minipage}

    \vspace{0.8em}

    %==================== 第二行 ====================%
    % \hspace*{-4.5em}
    \begin{minipage}[t]{\colw}\centering
        \includegraphics[width=\linewidth]{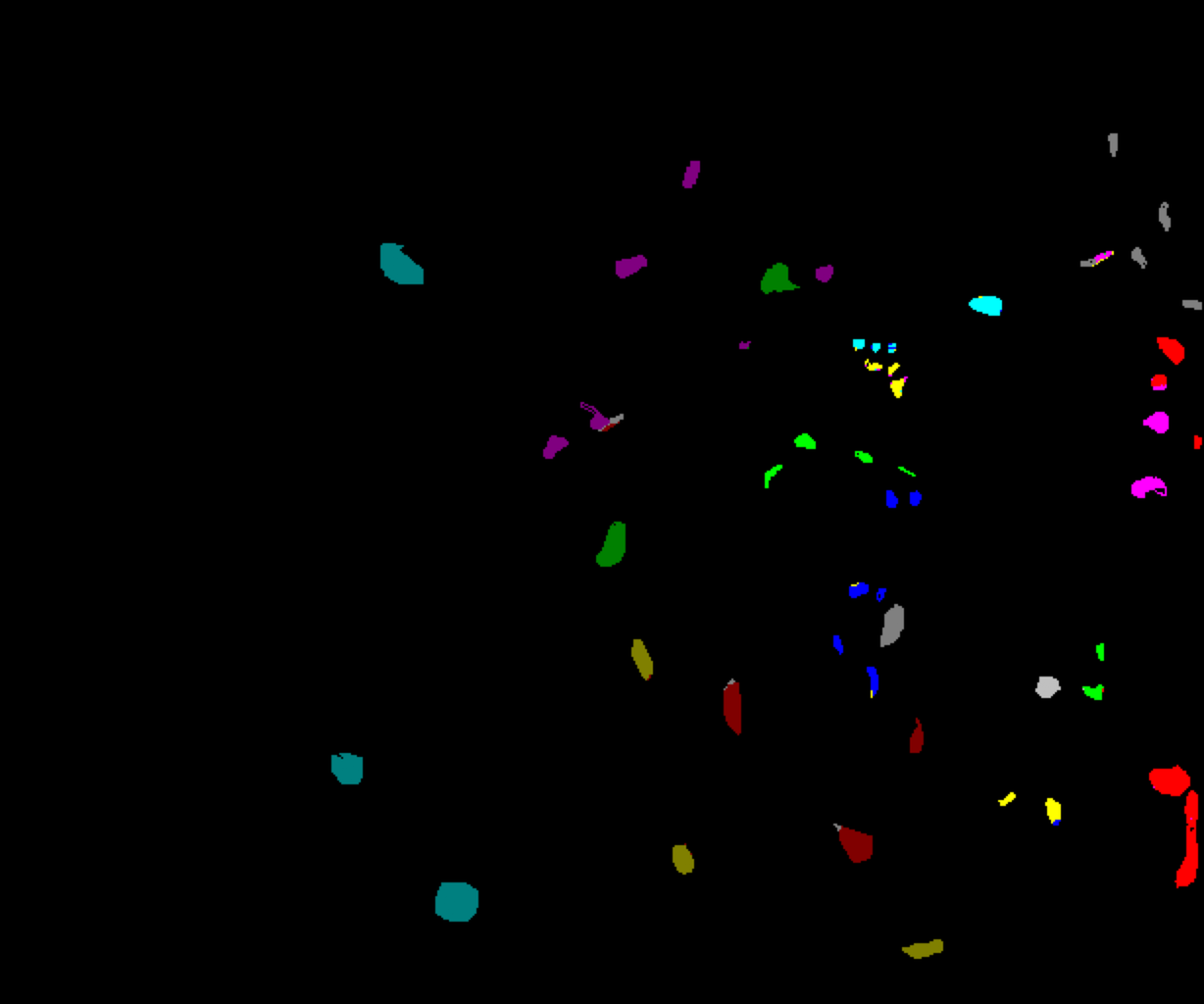}\\[0.2em]
        \scriptsize(f)
    \end{minipage}\hspace{\colsep}
    \begin{minipage}[t]{\colw}\centering
        \includegraphics[width=\linewidth]{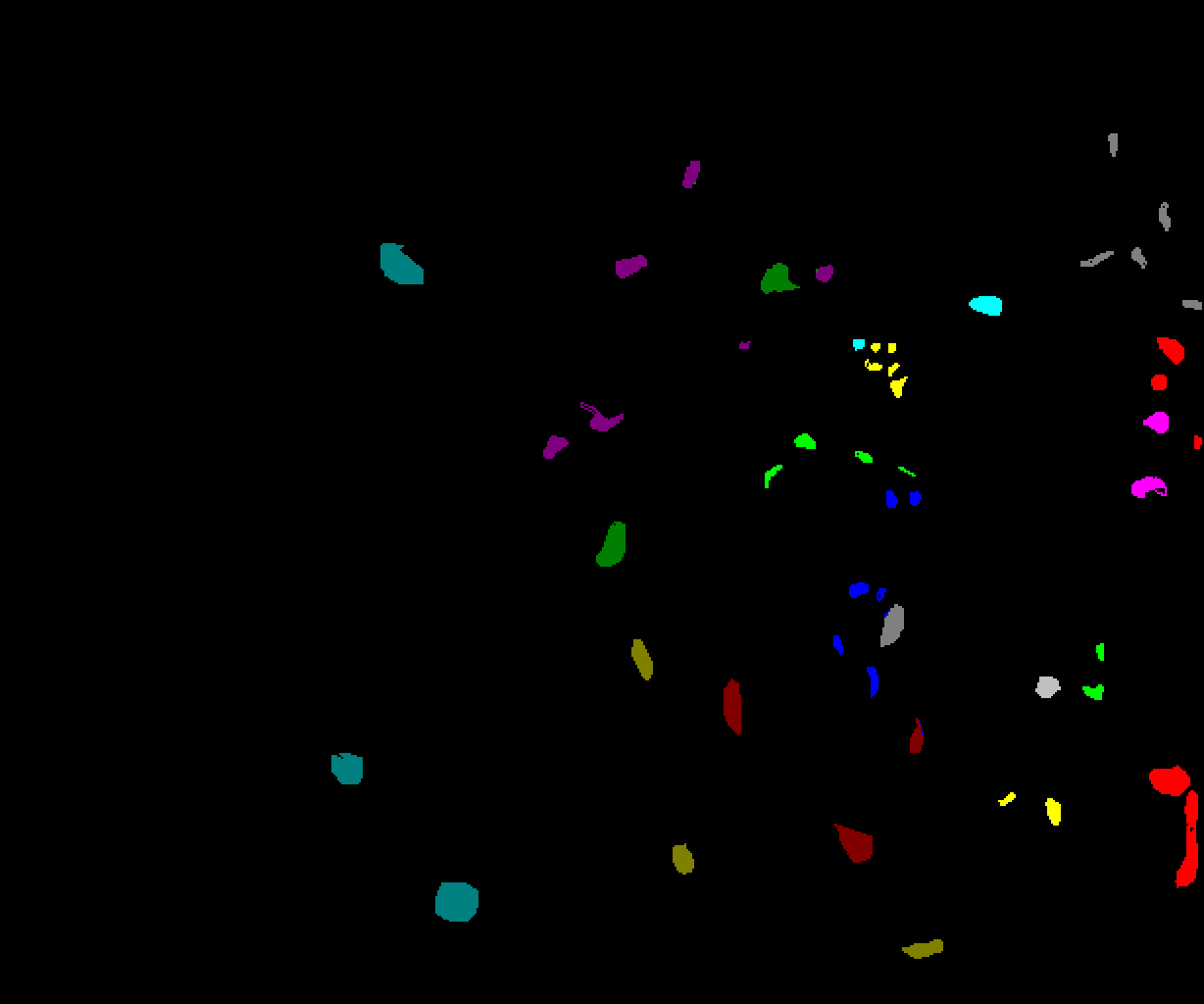}\\[0.2em]
        \scriptsize(g)
    \end{minipage}\hspace{\colsep}
    \begin{minipage}[t]{\colw}\centering
        \includegraphics[width=\linewidth]{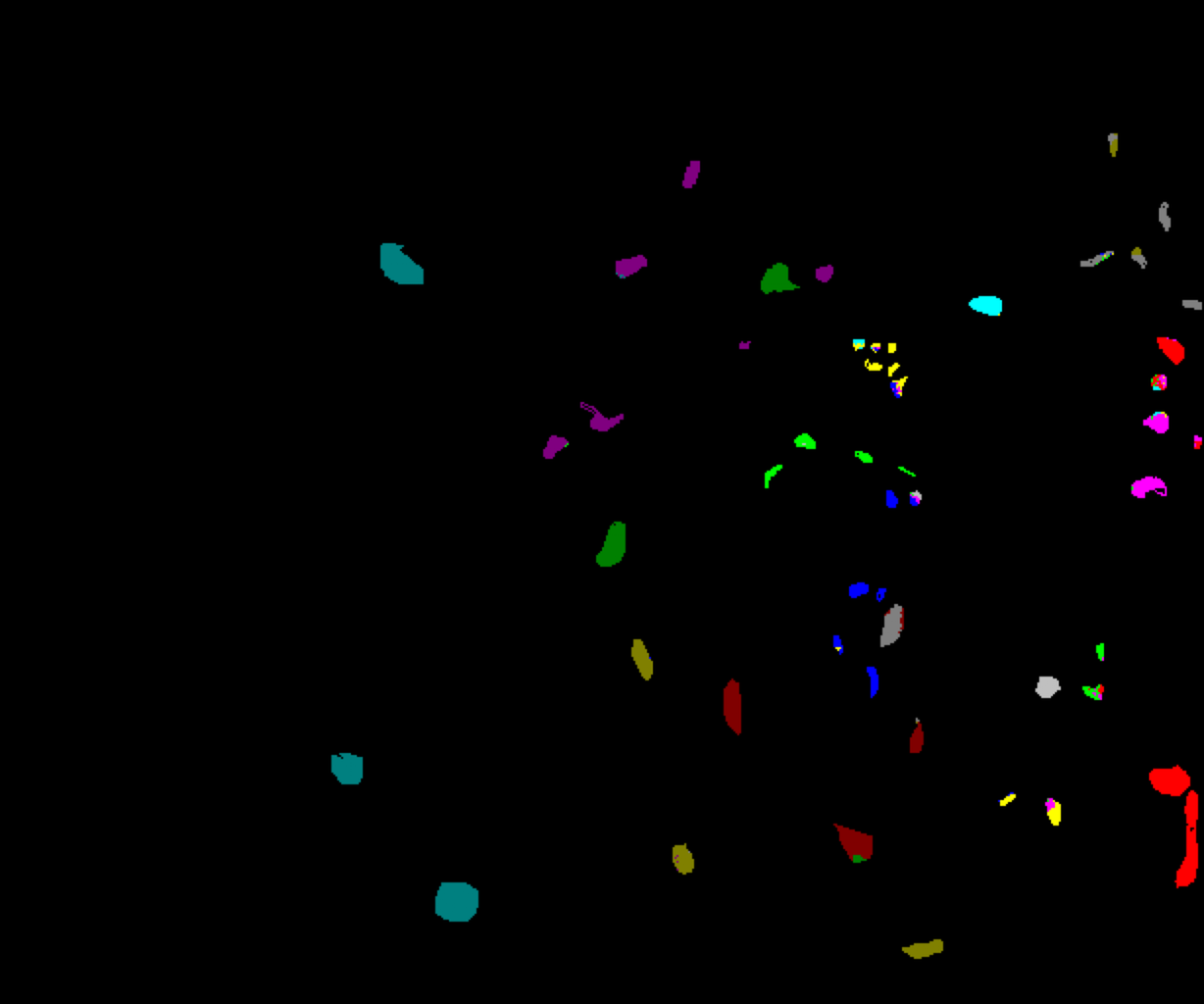}\\[0.2em]
        \scriptsize(h)
    \end{minipage}\hspace{\colsep}
    \begin{minipage}[t]{\colw}\centering
        \includegraphics[width=\linewidth]{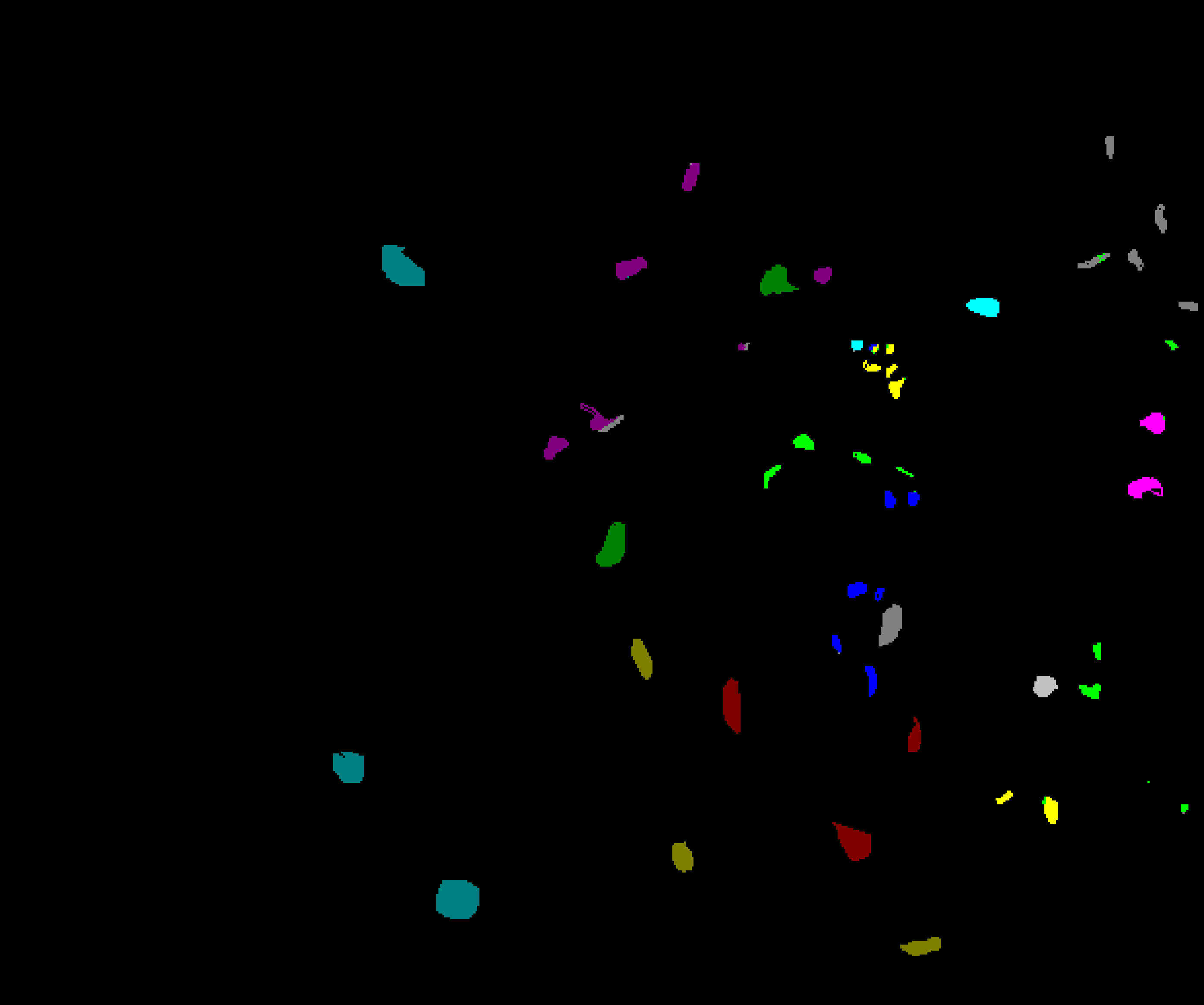}\\[0.2em]
        \scriptsize(i)
    \end{minipage}\hspace{\colsep}
    \begin{minipage}[t]{\colw}\centering
        \includegraphics[width=\linewidth]{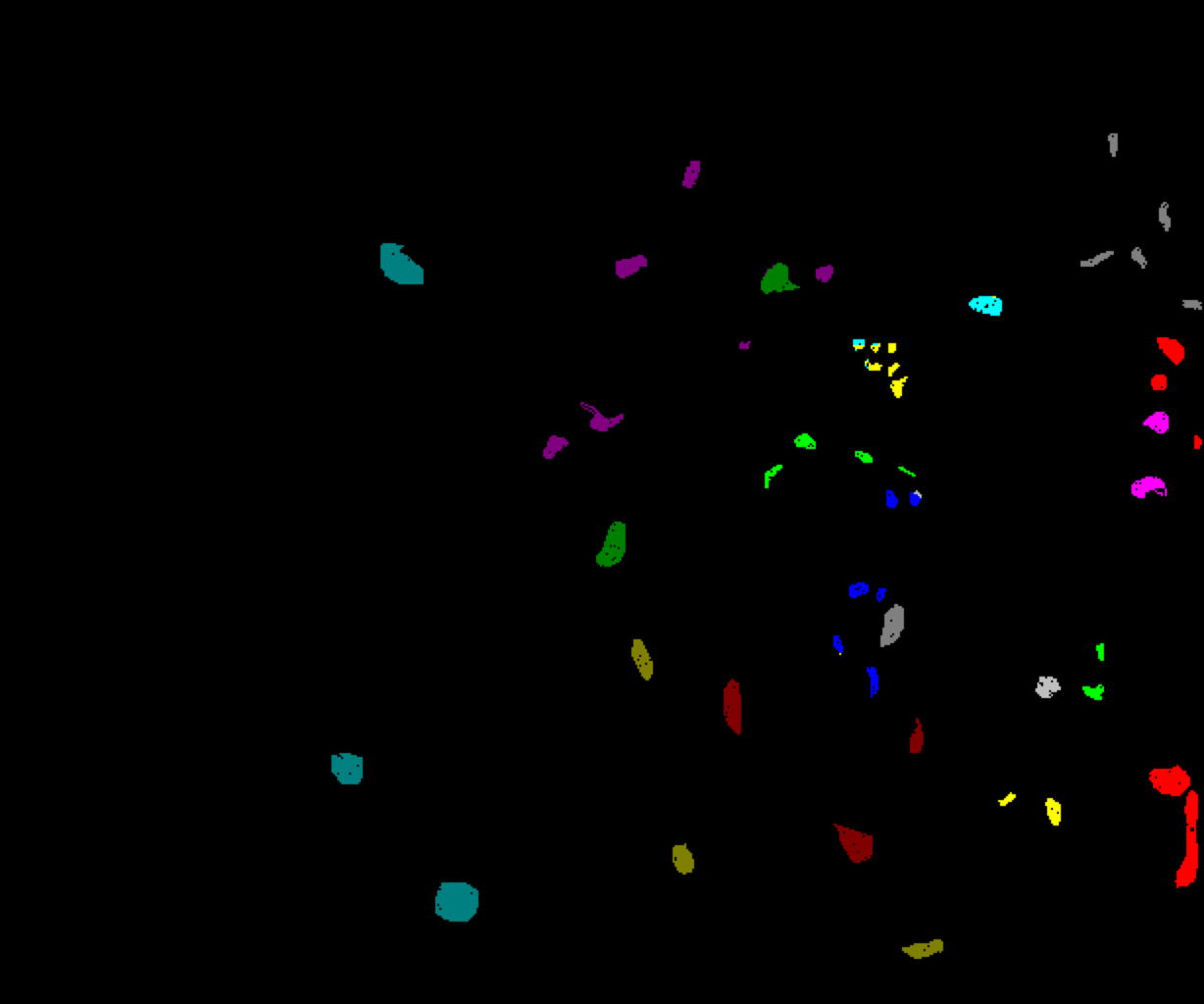}\\[0.2em]
        \scriptsize(j)
    \end{minipage}

    \vspace{0.8em}

    \begin{center}
    \begin{tikzpicture}[
    every node/.style={
        minimum width=2.2cm,
        minimum height=0.65cm,
        font=\small,
        inner sep=0pt,
        outer sep=0pt,
        align=center,
        text width=1.8cm
    }
    ]
    
    %====================
    % 第一行
    %====================
    
    \node[fill={rgb,1:red,0.0;green,0.0;blue,0.0},text=white] at (1.1,0)
    {\textbf{Unlabelled}};
    
    \node[fill={rgb,1:red,1.0;green,0.0;blue,0.0},text=white] at (3.3,0)
    {\textbf{Scrub}};
    
    \node[fill={rgb,1:red,0.0;green,1.0;blue,0.0},text=black] at (5.5,0)
    {\textbf{Swamp}};
    
    \node[fill={rgb,1:red,0.0;green,0.0;blue,1.0},text=white] at (7.7,0)
    {\textbf{Palm}};
    
    \node[fill={rgb,1:red,1.0;green,1.0;blue,0.0},text=black] at (9.9,0)
    {\textbf{Oak}};
    
    \node[fill={rgb,1:red,0.0;green,1.0;blue,1.0},text=black] at (12.1,0)
    {\textbf{Pine}};
    
    \node[fill={rgb,1:red,1.0;green,0.0;blue,1.0},text=white] at (14.3,0)
    {\textbf{Broadleaf}};
    
    %====================
    % 第二行
    %====================
    
    \node[fill={rgb,1:red,0.75;green,0.75;blue,0.75},text=black] at (1.1,-0.65)
    {\textbf{Hardwood}};
    
    \node[fill={rgb,1:red,0.5;green,0.5;blue,0.5},text=white] at (3.3,-0.65)
    {\textbf{Graminoid}};
    
    \node[fill={rgb,1:red,0.5;green,0.0;blue,0.0},text=white] at (5.5,-0.65)
    {\textbf{Spartina}};
    
    \node[fill={rgb,1:red,0.5;green,0.5;blue,0.0},text=white] at (7.7,-0.65)
    {\textbf{Cattail}};
    
    \node[fill={rgb,1:red,0.0;green,0.5;blue,0.0},text=white] at (9.9,-0.65)
    {\textbf{Salt}};
    
    \node[fill={rgb,1:red,0.5;green,0.0;blue,0.5},text=white] at (12.1,-0.65)
    {\textbf{Mud}};
    
    \node[fill={rgb,1:red,0.0;green,0.5;blue,0.5},text=white] at (14.3,-0.65)
    {\textbf{Water}};
    
    \end{tikzpicture}
    \end{center}

    \vspace{0.1em}
    % \tiny(k)

    \caption{Classification maps for the KSC dataset. (a) False color image. (b) Ground-truth. (c) A2S2K. (d) DMSGer. (e) SSTN. (f) CTF-SSCL. (g) DEMAE. (h) RMAE. (i) Semi. (j) Ours. }
    \label{fig:KSC_classification_maps}
\end{figure*}

\begin{figure*}[!t]
    % \raggedright
    % \hspace*{-2em}
    \centering

    %==================== 参数 ====================%
    \newcommand{\colw}{0.097\textwidth}
    \newcommand{\colsep}{-0.007\textwidth}

    %==================== 分类图 ====================%
    \begin{minipage}[t]{\colw}\centering
        \includegraphics[width=0.95\linewidth]{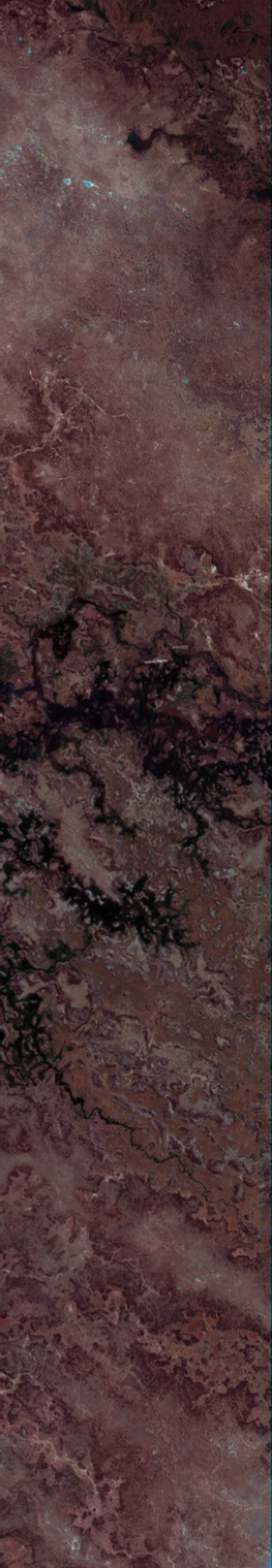}\\[0.2em]
        \scriptsize(a)
    \end{minipage}\hspace{\colsep}
    \begin{minipage}[t]{\colw}\centering
        \includegraphics[width=0.95\linewidth]{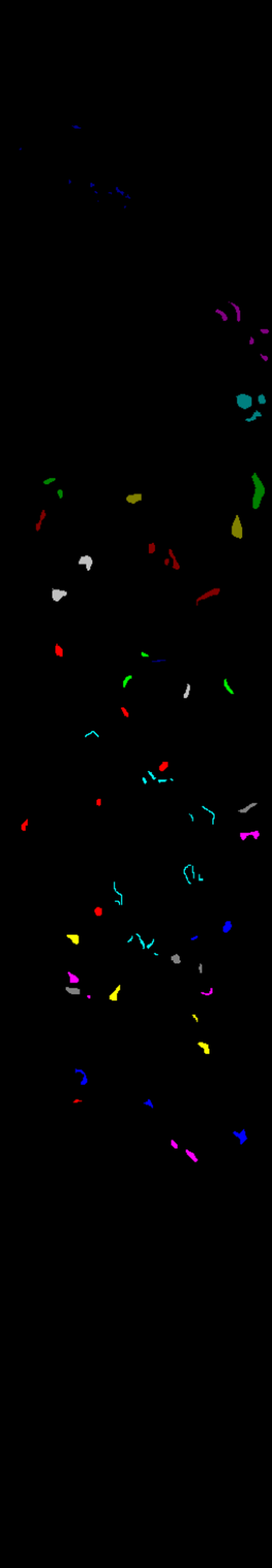}\\[0.2em]
        \scriptsize(b)
    \end{minipage}\hspace{\colsep}
    \begin{minipage}[t]{\colw}\centering
        \includegraphics[width=0.95\linewidth]{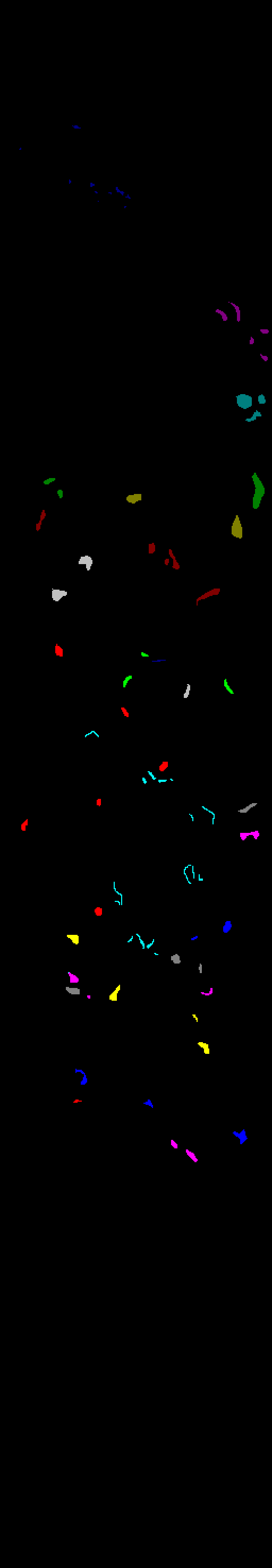}\\[0.2em]
        \scriptsize(c)
    \end{minipage}\hspace{\colsep}
    \begin{minipage}[t]{\colw}\centering
        \includegraphics[width=0.95\linewidth]{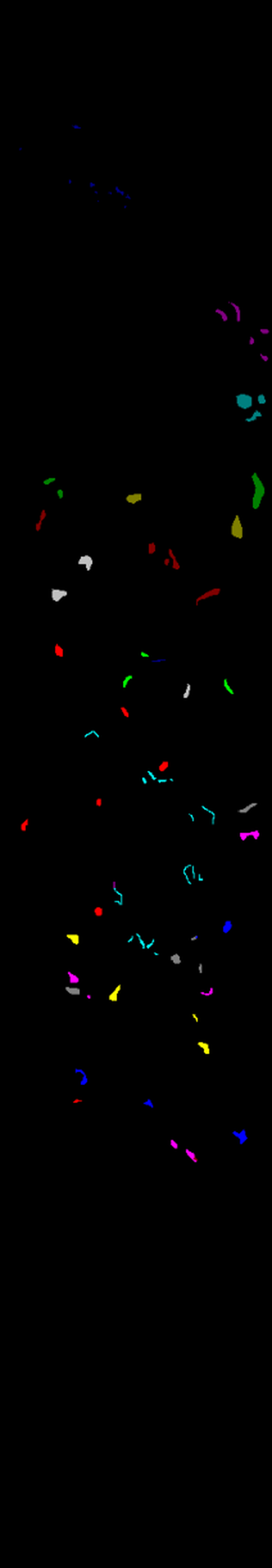}\\[0.2em]
        \scriptsize(d)
    \end{minipage}\hspace{\colsep}
    \begin{minipage}[t]{\colw}\centering
        \includegraphics[width=0.95\linewidth]{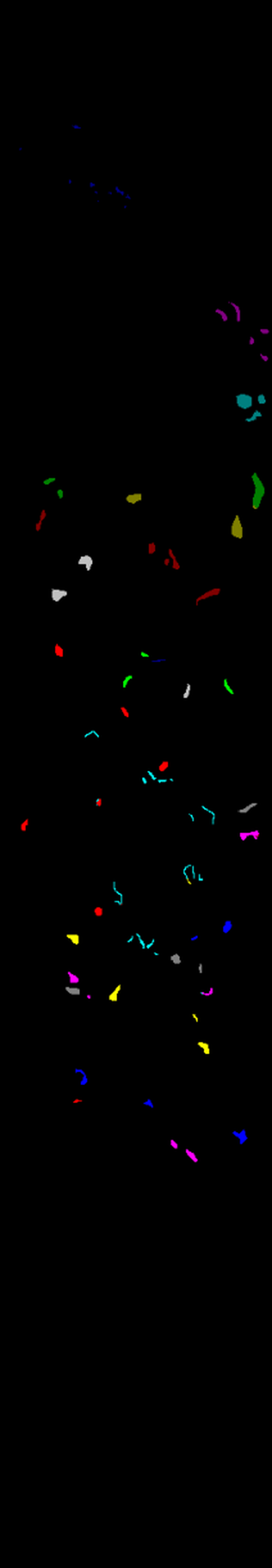}\\[0.2em]
        \scriptsize(e)
    \end{minipage}\hspace{\colsep}
    \begin{minipage}[t]{\colw}\centering
        \includegraphics[width=0.95\linewidth]{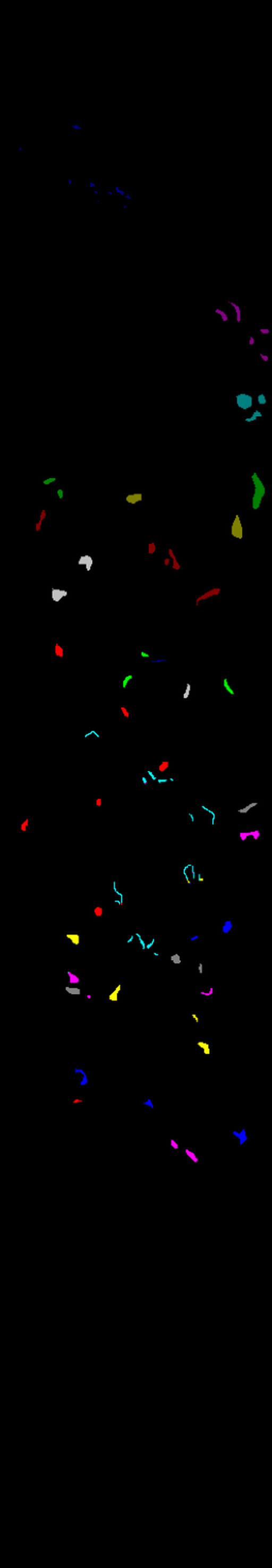}\\[0.2em]
        \scriptsize(f)
    \end{minipage}\hspace{\colsep}
    \begin{minipage}[t]{\colw}\centering
        \includegraphics[width=0.95\linewidth]{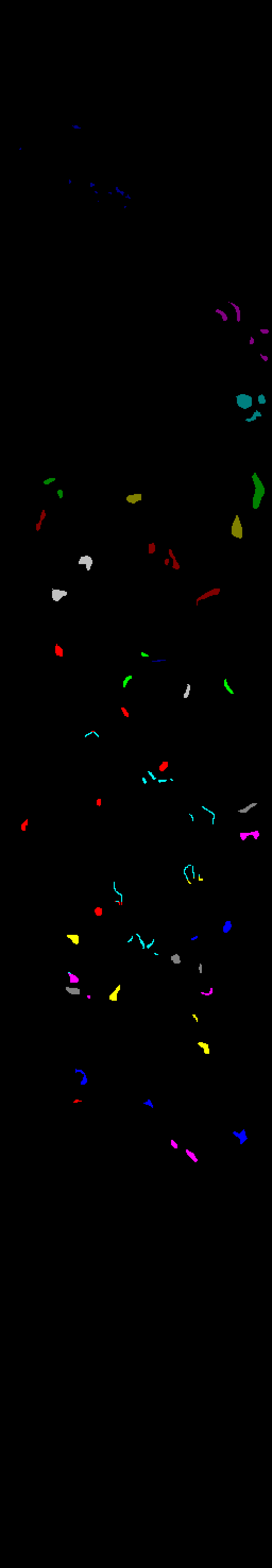}\\[0.2em]
        \scriptsize(g)
    \end{minipage}\hspace{\colsep}
    \begin{minipage}[t]{\colw}\centering
        \includegraphics[width=0.95\linewidth]{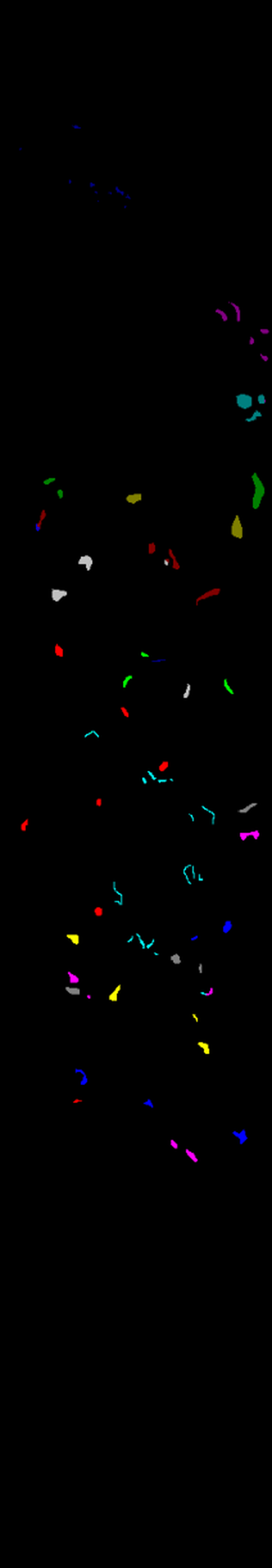}\\[0.2em]
        \scriptsize(h)
    \end{minipage}\hspace{\colsep}
    \begin{minipage}[t]{\colw}\centering
        \includegraphics[width=0.962\linewidth]{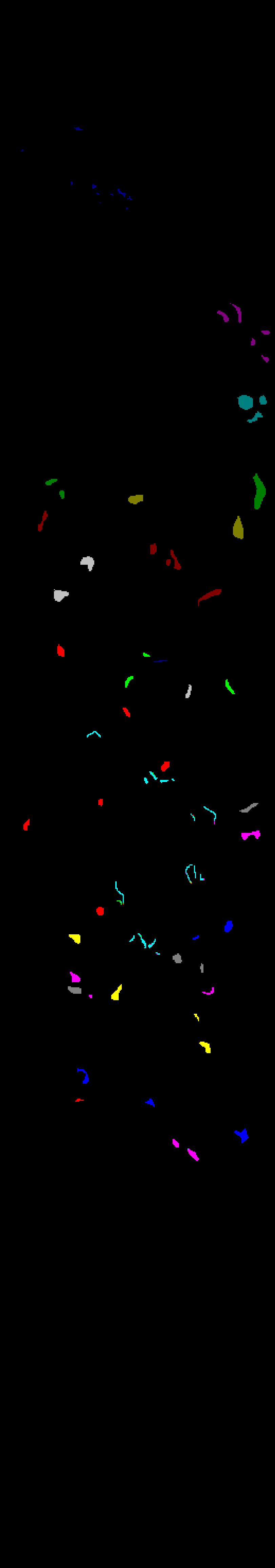}\\[0.2em]
        \scriptsize(i)
    \end{minipage}\hspace{\colsep}
    \begin{minipage}[t]{\colw}\centering
        \includegraphics[width=0.95\linewidth]{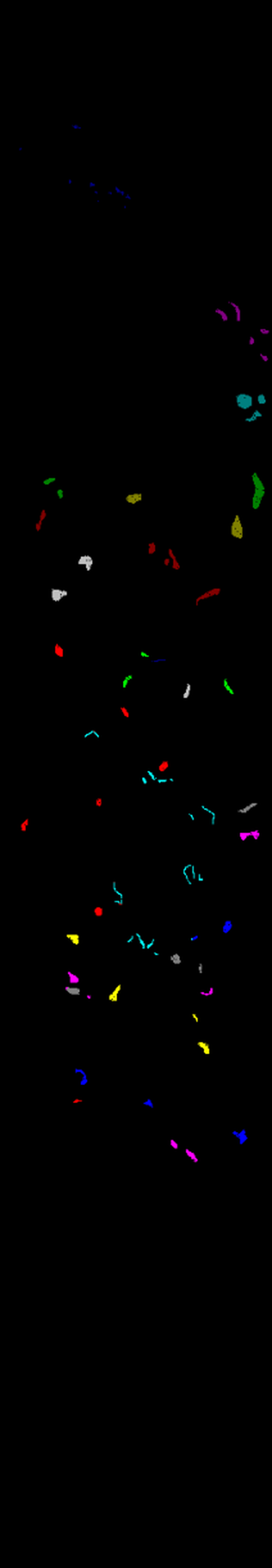}\\[0.2em]
        \scriptsize(j)
    \end{minipage}

    %==================== 横向颜色条 ====================%
    
    \vspace{0.1em}

    % \begin{minipage}{\textwidth}
    % \centering
    % \tiny
    % \setlength{\tabcolsep}{3pt}
    % \renewcommand{\arraystretch}{1.1}
    % \setlength{\tabcolsep}{4pt}
    
    % \begin{tabular}{ccccccccccccccc}
    
    % \textcolor[rgb]{0,0,0}{\rule{4mm}{4mm}} &
    % \textcolor[rgb]{1,0,0}{\rule{4mm}{4mm}} &
    % \textcolor[rgb]{0,1,0}{\rule{4mm}{4mm}} &
    % \textcolor[rgb]{0,0,1}{\rule{4mm}{4mm}} &
    % \textcolor[rgb]{1,1,0}{\rule{4mm}{4mm}} &
    % \textcolor[rgb]{0,1,1}{\rule{4mm}{4mm}} &
    % \textcolor[rgb]{1,0,1}{\rule{4mm}{4mm}} &
    % \textcolor[rgb]{0.75,0.75,0.75}{\rule{4mm}{4mm}} &
    % \textcolor[rgb]{0.5,0.5,0.5}{\rule{4mm}{4mm}} &
    % \textcolor[rgb]{0.5,0,0}{\rule{4mm}{4mm}} &
    % \textcolor[rgb]{0.5,0.5,0}{\rule{4mm}{4mm}} &
    % \textcolor[rgb]{0,0.5,0}{\rule{4mm}{4mm}} &
    % \textcolor[rgb]{0.5,0,0.5}{\rule{4mm}{4mm}} &
    % \textcolor[rgb]{0,0.5,0.5}{\rule{4mm}{4mm}} &
    % \textcolor[rgb]{0,0,0.5}{\rule{4mm}{4mm}} \\
    
    % Unlabelled &
    % Water &
    % Hippo grass &
    % Floodplain grasses 1 &
    % Floodplain grasses 2 &
    % Reeds &
    % Riparian &
    % Firescar &
    % Island &
    % Woodlands &
    % Shrublands &
    % Grasslands &
    % Short Mopane &
    % Mixed Mopane &
    % Exposed Soils \\
    
    % \end{tabular}

    \begin{center}
    \begin{tikzpicture}[
    every node/.style={
        minimum width=2.2cm,
        minimum height=0.7cm,
        font=\footnotesize,
        inner sep=0pt,
        outer sep=0pt,
        align=center,
        text width=1.8cm
    }
    ]
    
    %====================
    % 第一行（8类）
    %====================
    
    \node[fill={rgb,1:red,0;green,0;blue,0},text=white]
    at (1.1,0) {\textbf{Unlabelled}};
    
    \node[fill={rgb,1:red,1;green,0;blue,0},text=white]
    at (3.3,0) {\textbf{Water}};
    
    \node[fill={rgb,1:red,0;green,1;blue,0},text=black]
    at (5.5,0) {\textbf{Hippo grass}};
    
    \node[fill={rgb,1:red,0;green,0;blue,1},text=white]
    at (7.7,0) {\textbf{Floodplain grasses 1}};
    
    \node[fill={rgb,1:red,1;green,1;blue,0},text=black]
    at (9.9,0) {\textbf{Floodplain grasses 2}};
    
    \node[fill={rgb,1:red,0;green,1;blue,1},text=black]
    at (12.1,0) {\textbf{Reeds}};
    
    \node[fill={rgb,1:red,1;green,0;blue,1},text=white]
    at (14.3,0) {\textbf{Riparian}};
    
    \node[fill={rgb,1:red,0.75;green,0.75;blue,0.75},text=black]
    at (16.5,0) {\textbf{Firescar}};
    
    %====================
    % 第二行（7类）
    %====================
    
    \node[fill={rgb,1:red,0.5;green,0.5;blue,0.5},text=white]
    at (1.1,-0.7) {\textbf{Island}};
    
    \node[fill={rgb,1:red,0.5;green,0;blue,0},text=white]
    at (3.3,-0.7) {\textbf{Woodlands}};
    
    \node[fill={rgb,1:red,0.5;green,0.5;blue,0},text=white]
    at (5.5,-0.7) {\textbf{Shrublands}};
    
    \node[fill={rgb,1:red,0;green,0.5;blue,0},text=white]
    at (7.7,-0.7) {\textbf{Grasslands}};
    
    \node[fill={rgb,1:red,0.5;green,0;blue,0.5},text=white]
    at (9.9,-0.7) {\textbf{Short Mopane}};
    
    \node[fill={rgb,1:red,0;green,0.5;blue,0.5},text=white]
    at (12.1,-0.7) {\textbf{Mixed Mopane}};
    
    \node[fill={rgb,1:red,0;green,0;blue,0.5},text=white]
    at (14.3,-0.7) {\textbf{Exposed Soils}};
    
    \end{tikzpicture}
    \end{center}

    \vspace{0.1em}
    % \tiny(k)
    
    % \end{minipage}

    \caption{Classification maps for the Botswana dataset. 
    (a) False color image. 
    (b) Ground-truth. 
    (c) A2S2K. 
    (d) DMSGer. 
    (e) SSTN. 
    (f) CTF-SSCL. 
    (g) DEMAE. 
    (h) RMAE. 
    (i) Semi-HCN. 
    (j) Ours. }
    % (k) Color labels.}
    
    \label{fig:Botswana_classification_maps}

\end{figure*}

\subsection{Experimental setup}
The implementation is built upon the PyTorch framework, with all experiments trained on an NVIDIA RTX 2080 Ti GPU for computational efficiency. A standardized preprocessing pipeline is applied to the input HSI data: linear normalization of reflectance values to [0,1], followed by division into $24 \times 24$ pixel patches that are converted into three-dimensional cubes compatible with the network architecture.

The study adopts balanced class sampling, randomly selecting 10 samples per class for training while reserving the remainder for testing. To ensure robust statistical analysis, all experiments are repeated through 10 independent trials with averaged results reported. The Adam optimizer is configured with an initial learning rate of \(1\times10^{-3}\) for 200 training epochs. Performance evaluation comprehensively measures Overall Accuracy (OA), Average Accuracy (AA), and the Kappa coefficient.

For the superpixel generation module, the number of superpixels is determined by the image size and a predefined pixel number per superpixel to balance spatial granularity and computational cost. In practice, this parameter is set to 200 for Houston2013 and 50 for the other datasets.

\subsection{Performance comparison}

Comparative analyses include two fully-supervised baselines (A2S2K\citep{ref15}, SSTN\citep{ref16}), three semi-supervised models (DMSGer\citep{ref32}, CTF-SSCL\citep{ref33}, Semi-HCN\citep{semi-HCN}), and two few-shot self-supervised methods (DEMAE\citep{ref23}, RMAE\citep{ref24} ). Tables \ref{tab:PaviaU_results}--\ref{tab:Botswana_results} summarize the classification accuracy of various methods, with the corresponding classification maps illustrated in Fig. \ref{fig:PaviaU_classification_maps}--\ref{fig:Botswana_classification_maps}.

On the PaviaU dataset, the proposed method exhibits superior performance, as shown in Table~\ref{tab:PaviaU_results}, attaining 95.21\% OA, 94.46\% AA, and 93.67\% Kappa--surpassing RMAE by 2.53\%, 0.24\%, and 3.21\%, respectively. In terms of per-class accuracy, the proposed model performs better across most land-cover categories, particularly in Meadows, Gravel, and Self-blocking bricks (Classes~2,~3,~and~8). As illustrated in Fig~\ref{fig:PaviaU_classification_maps}, the classification map produced by the proposed method exhibits smoother boundary transitions and more complete regional delineation.

The proposed method achieves notable results on the Houston 2013 dataset (Table~\ref{tab:Houston2013_results}), with OA/AA/Kappa values of 89.77\%/90.87\%/88.94\%, representing significant enhancement of 1.75\%/1.94\%/2.77\% over DEMAE. It exhibits clear superiority in classifying buildings, grass, and low vegetation categories. The visually evident boundary blurring in DEMAE's outputs underscores our method's enhanced boundary definition capacity (Fig~\ref{fig:Houston_classification_maps}).

 The proposed method achieves notable results on the Salinas dataset (see Table~\ref{tab:Salinas_results}), achieving OA/AA/Kappa values of 96.85\%/98.38\%/96.50\%, which represent increases of 0.54\%/2.01\%/0.44\% over the second-best A2S2K. From a qualitative perspective, as shown in Fig~\ref{fig:Salinas_classification_maps}, our method performs noticeably better than A2S2K on the celery and vinyard untrained categories. Moreover, it attains the best results for the soil develop category, as most other methods tend to misclassify pixels from other classes into this category.

As shown in Table~\ref{tab:KSC_results}, the proposed method achieves remarkable performance on the KSC dataset, with OA, AA, and Kappa achieving 99.71\%, 99.48\%, and 99.68\%, respectively--exceeding DEMAE by 0.49\%, 0.38\%, and 0.55\%. The model demonstrates near-perfect accuracy for vegetation, soil, and wetland categories, indicating strong discriminative power and generalization capability.

For the Botswana dataset (see Table~\ref{tab:Botswana_results}), the proposed method achieves 99.19\% OA, 99.22\% AA, and 99.12\% Kappa, outperforming RMAE by 0.36\%, 0.19\%, and 0.39\%. Although SSTN attains near-saturated overall accuracy, its limited capacity for capturing local details results in marginally reduced comprehensive performance. Other competing methods, heavily reliant on massive labeled samples, perform relatively inferior performance under limited annotation conditions. These results further underscore the robustness and superiority of the proposed method.

\begin{table}[!t]
\centering
\caption{Ablation study of key modules on four representative datasets.}
\label{tab:ablation}
\renewcommand{\arraystretch}{1.1}
\setlength{\tabcolsep}{4pt}
\resizebox{\linewidth}{!}{%
\begin{tabular}{c l c c c c}
\hline
Dataset & Metrics & w/o EASLP & w/o DHP & w/o ATSC & Ours \\
\hline
\multirow{3}{*}{PaviaU}
& OA (\%) & 92.66 $\pm$ 2.53 & 94.20 $\pm$ 1.50 & 94.87 $\pm$ 1.32 & \textbf{95.21 $\pm$ 1.17} \\
& AA (\%) & 91.45 $\pm$ 1.90 & 94.22 $\pm$ 0.80 & 93.97 $\pm$ 1.35 & \textbf{94.46 $\pm$ 1.48} \\
& $\kappa\times100$ & 90.09 $\pm$ 3.13 & 92.37 $\pm$ 1.92 & 93.21 $\pm$ 1.75 & \textbf{93.67 $\pm$ 1.54} \\
\hline
\multirow{3}{*}{Houston2013}
& OA (\%) & 88.90 $\pm$ 1.33 & 89.38 $\pm$ 1.64 & 89.16 $\pm$ 1.39 & \textbf{89.77 $\pm$ 1.17} \\
& AA (\%) & 90.17 $\pm$ 0.86 & 90.56 $\pm$ 1.31 & 90.35 $\pm$ 1.18 & \textbf{90.87 $\pm$ 0.98} \\
& $\kappa\times100$ & 88.00 $\pm$ 1.44 & 88.52 $\pm$ 1.78 & 88.28 $\pm$ 1.51 & \textbf{88.94 $\pm$ 1.26} \\
\hline
\multirow{3}{*}{KSC}
& OA (\%) & 99.60 $\pm$ 0.13 & 99.66 $\pm$ 0.14 & 99.65 $\pm$ 0.14 & \textbf{99.71 $\pm$ 0.17} \\
& AA (\%) & 99.43 $\pm$ 0.82 & 99.44 $\pm$ 0.22 & 99.40 $\pm$ 0.25 & \textbf{99.48 $\pm$ 0.30} \\
& $\kappa\times100$ & 99.58 $\pm$ 0.17 & 99.62 $\pm$ 0.16 & 99.60 $\pm$ 0.16 & \textbf{99.68 $\pm$ 0.19} \\
\hline
\multirow{3}{*}{Botswana}
& OA (\%) & 99.14 $\pm$ 0.39 & 99.18 $\pm$ 0.53 & 99.12 $\pm$ 0.46 & \textbf{99.19 $\pm$ 0.40} \\
& AA (\%) & 99.17 $\pm$ 0.45 & 99.20 $\pm$ 0.54 & 99.09 $\pm$ 0.39 & \textbf{99.22 $\pm$ 0.49} \\
& $\kappa\times100$ & 99.07 $\pm$ 0.44 & 99.11 $\pm$ 0.57  & 99.03 $\pm$ 0.36 & \textbf{99.12 $\pm$ 0.44} \\
\hline
\end{tabular}%
}
\end{table}

\subsection{Parameter analysis}

Comprehensive sensitivity experiments are conducted to assess three critical hyperparameters: the number of superpixels, the historical prediction queue's window size, the dynamic historical fusion weight $\alpha$, and the number of training samples.

\begin{figure}[!t]
    \centering
    \includegraphics[width=0.9\linewidth]{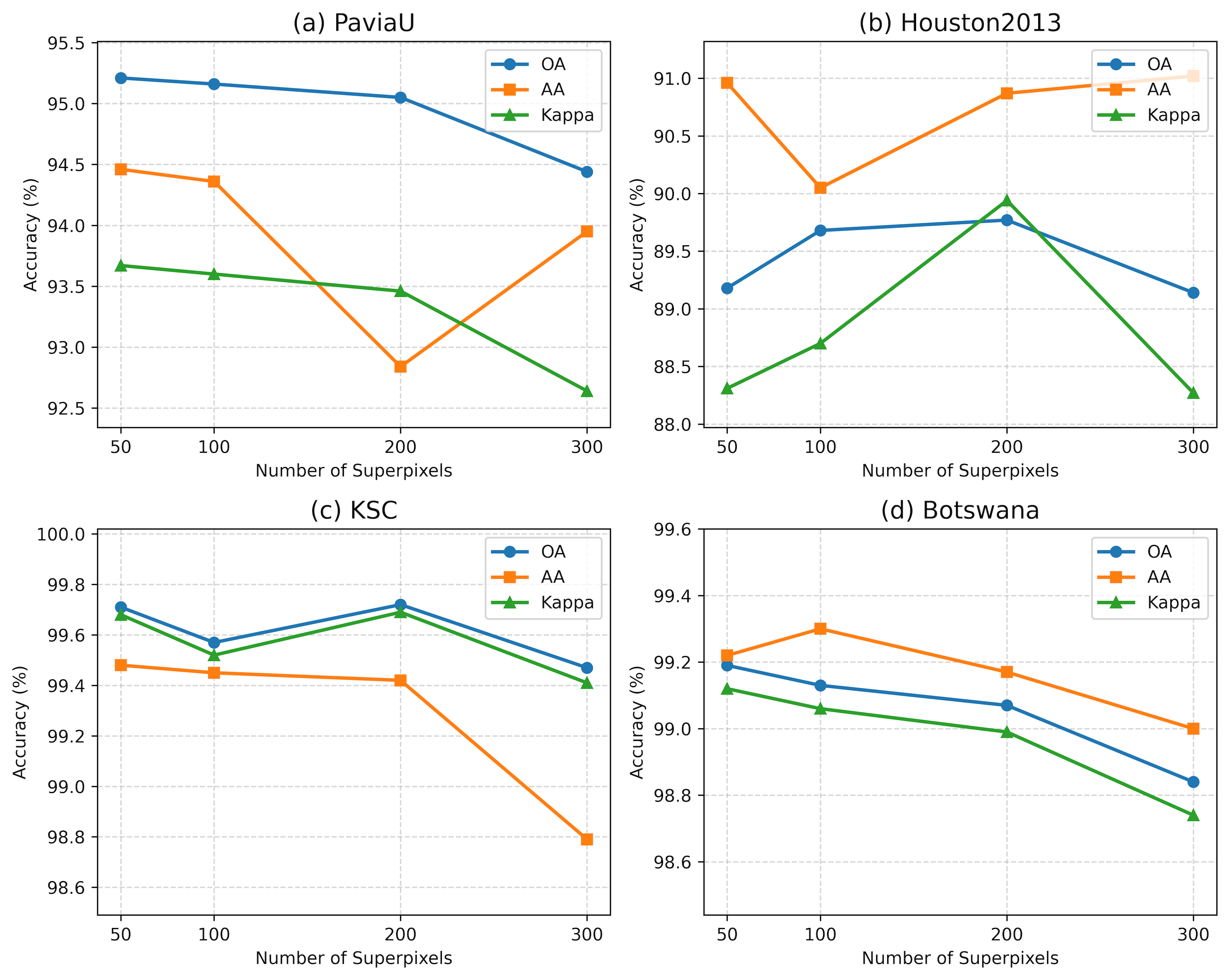}
    \caption{Sensitivity analysis on the number of superpixels across different datasets.}
    \label{fig:Number of Superpixels}
\end{figure}

\textbf{Number of superpixels:} Four candidate values that we tested were 50, 100, 200, and 300. The optimal number of superpixels depends on the characteristics of the dataset. For Houston2013, which contains diverse classes, complex scenes, and sparsely distributed land-cover categories, too few superpixels may cause under-segmentation and merge different classes, degrading fine-grained representation. In contrast, the other datasets contain larger and more homogeneous regions, where excessive superpixel partitioning may disrupt spatial integrity and reduce the effectiveness of spatial information utilization. According to the results in Fig~\ref{fig:Number of Superpixels}, 200 superpixels are selected for Houston2013, whereas 50 superpixels are used for the other datasets.

\begin{figure}[!t]
    \centering
    \includegraphics[width=0.9\linewidth]{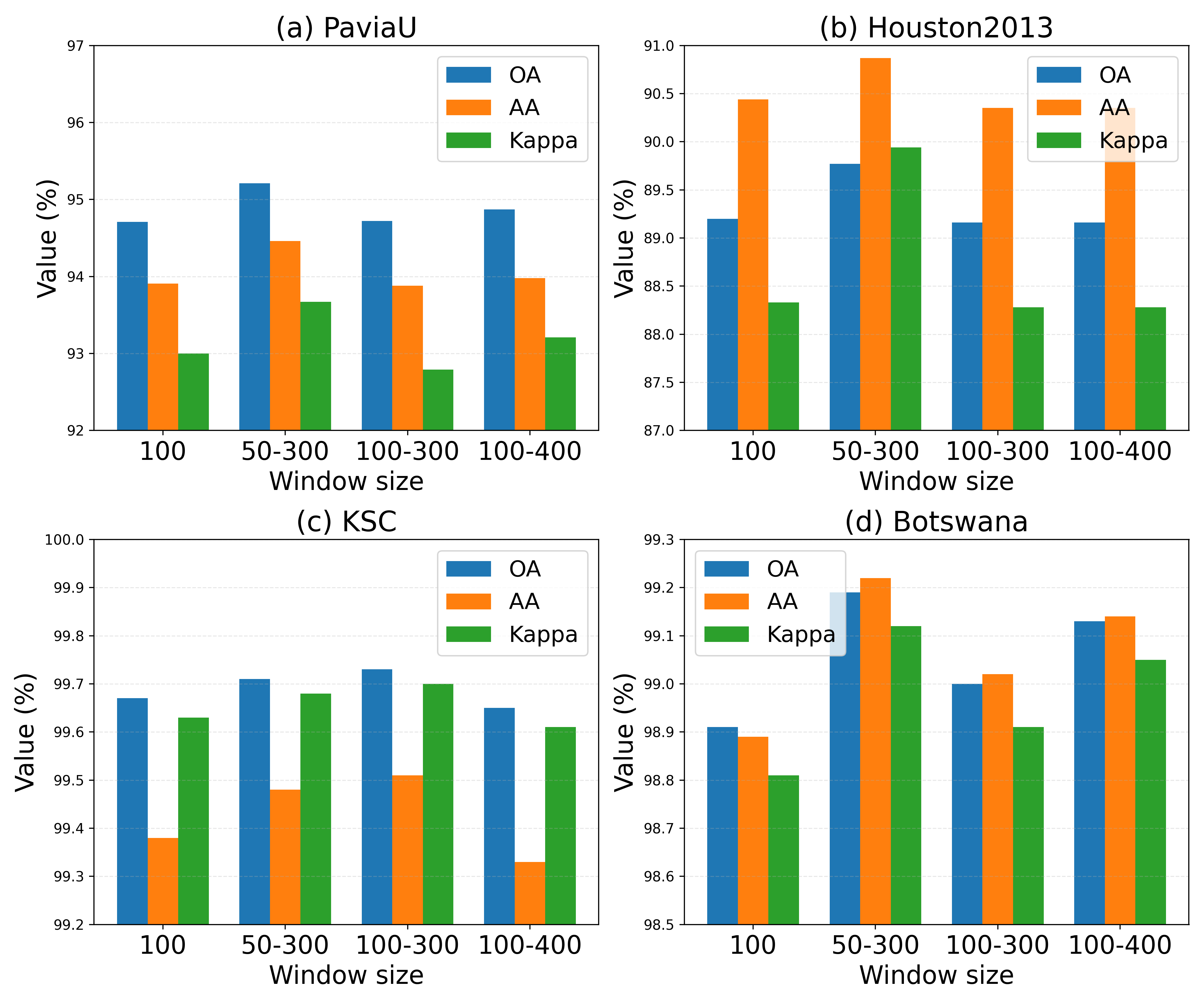}
    \caption{Sensitivity analysis on the historical window size across different datasets.}
    \label{fig:window-size}
\end{figure}

\textbf{Historical window size:} Three schemes are designed: fixed window (100) and dynamically increasing windows (50--300, 100--300, and 100--400). The fixed window evaluates the impact of a constant historical length on pseudo-label stability, whereas the dynamic window accommodates varying requirements across training stages---starting with shorter windows initially and gradually expanding to incorporate more historical information.  
Experimental results show that the dynamic window scheme with range 50--300 achieves optimal performance, indicating that moderate window growth establishes a favorable balance between convergence speed and stability. In comparison, schemes with fixed window or excessive/insufficient growth rate exhibit relatively inferior performance, as shown in Fig~\ref{fig:window-size}.

\begin{figure}[!t]
    \centering
    \includegraphics[width=0.9\linewidth]{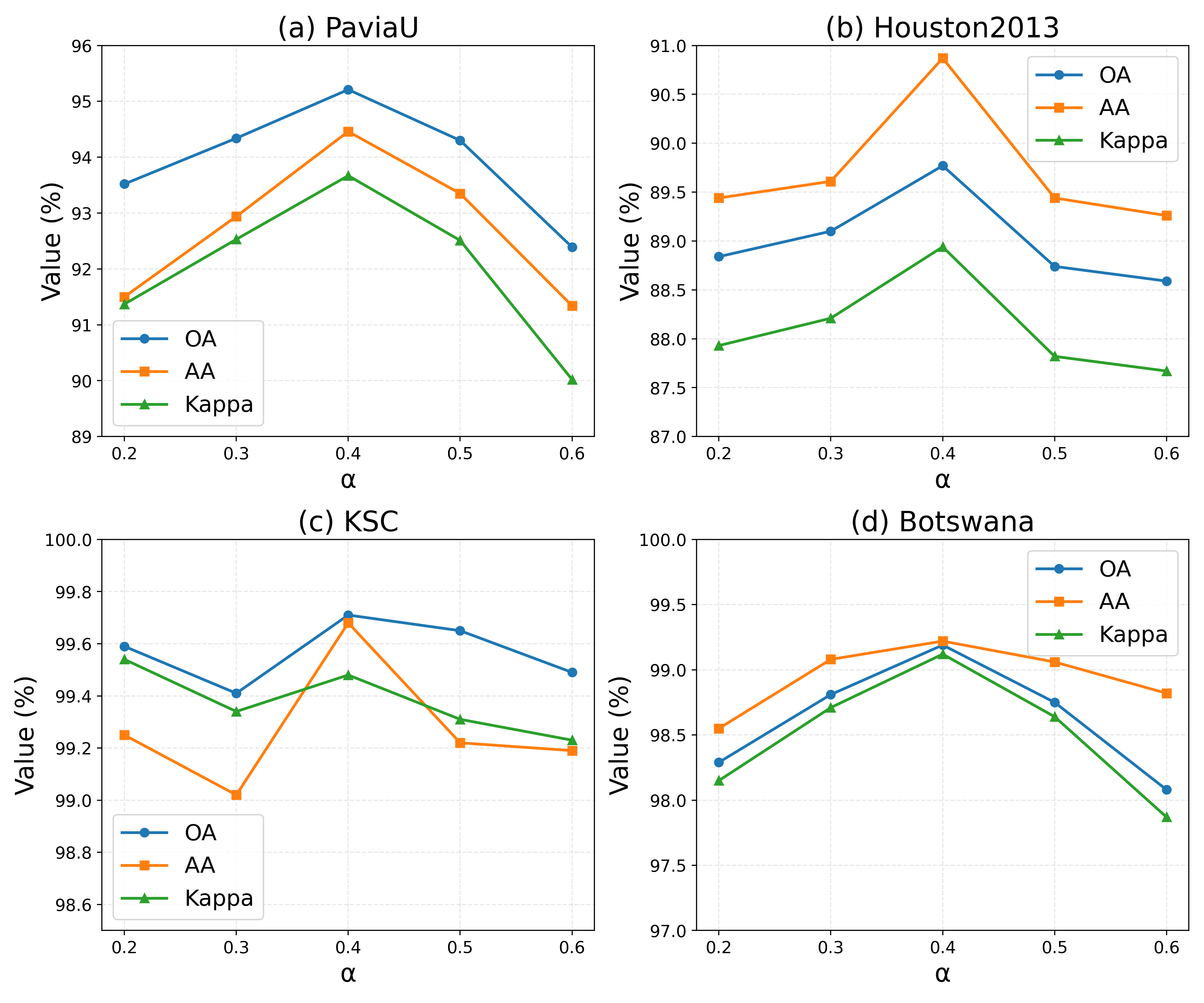}
    \caption{Sensitivity analysis on the dynamic historical fusion weight $\alpha$ across different datasets.}
    \label{fig:alpha}
\end{figure}

\textbf{Dynamic historical fusion weight $\alpha$:} Five ranges are designed: 0.1--0.2, 0.1--0.3, 0.1--0.4, 0.1--0.5, and 0.1--0.6. Results demonstrate that the model achieves optimal performance across all four datasets when the upper bound is set to 0.4. On PaviaU, performance peaks at 95.21\% OA, 94.46\% AA, and 93.67\% Kappa, respectively. Expanding the upper bound to 0.5 or 0.6 causes performance deterioration, with Houston2013 showing significant OA/Kappa declines. This reveals that excessive $\alpha$ values render pseudo-label updates too conservative, resulting in compromised adaptability to new patterns. The 0.1--0.4 range is consequently identified as the optimal balance for effective operation (see Fig~\ref{fig:alpha}). 

\begin{figure*}[!t]
  \centering
  \includegraphics[width=\textwidth]{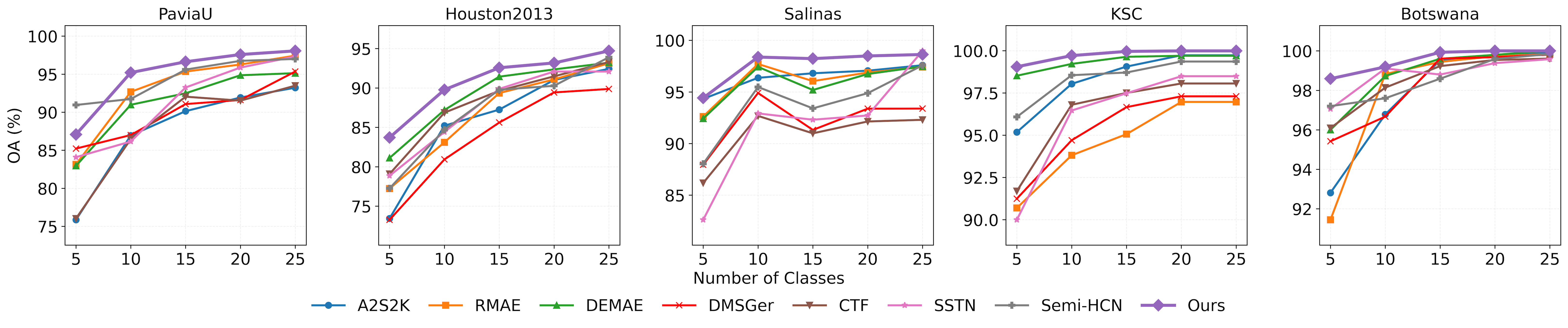}
  \caption{Overall accuracies with respect to different numbers of training samples on five datasets.}
  \label{fig:accuracy_comparison}
\end{figure*}

\textbf{The number of training samples:} A comprehensive scaling analysis is performed using 5-25 labeled samples per class at 5-sample intervals. The results indicate a monotonic improvement in OA with increasing sample size. The proposed method sustains superior performance across all data regimes, confirming its robustness in few-shot scenarios (see Fig.~\ref{fig:accuracy_comparison}).

\begin{figure}[!t]
  \centering
  \includegraphics[width=\linewidth]{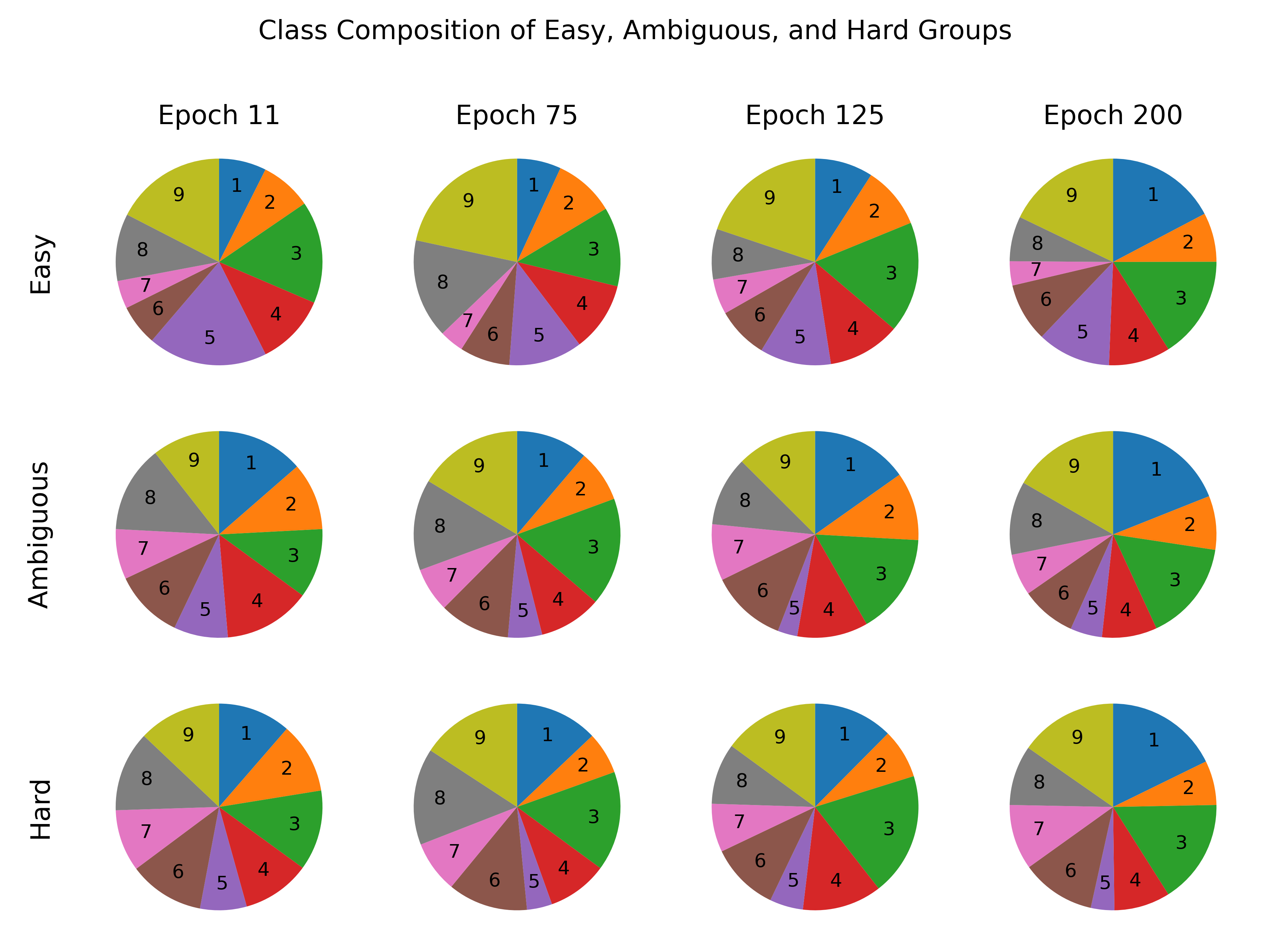}
  \caption{Class-wise composition of easy, ambiguous, and hard sample groups across different training stages on PaviaU dataset.}
  \label{fig:class_pie}
\end{figure}

\textbf{Class-wise distribution analysis of ATSC groups:}
Since the adaptive thresholds $\tau_a$ and $\tau_c$ are updated from batch-level statistics, class-wise population differences may affect the reliability distribution used for sample grouping. Therefore, Fig.~\ref{fig:class_pie} presents the class-wise composition of the easy, ambiguous, and hard groups on the PaviaU dataset, which exhibits notable class distribution disparities, at four representative training stages. It can be observed that Class 2, despite having the largest number of samples, does not occupy a dominant proportion in these groups. In contrast, Class 9, which has the fewest samples, consistently maintains a relatively large proportion across all three groups. These results indicate that the dynamic threshold update process does not exhibit an obvious bias toward majority classes under such imbalanced class distributions.

% illustrates the evolution of easy, ambiguous, and hard sample proportions during training. After the warm-up stage, most unlabeled samples are initially assigned to the hard group because the model predictions are still unreliable. As training progresses, the proportion of easy samples increases rapidly from approximately 35\% to nearly 80\%, while the proportion of hard samples decreases from over 60\% to around 10\%. In the later training stage, the sample distribution gradually stabilizes, with easy samples accounting for the majority of unlabeled samples and both ambiguous and hard samples remaining at relatively low levels. This behavior suggests that an increasing number of unlabeled samples are gradually recognized as reliable and incorporated into supervision, thereby reducing the risk of error accumulation and unstable pseudo-label reinforcement.

% \begin{figure}[!t]
%   \centering
%   \includegraphics[width=\linewidth]{figs/ratio.png}
%   \caption{Evolution of easy, ambiguous, and hard sample proportions during training.}
%   \label{fig:ratio}
% \end{figure}

\begin{figure}[!t]
\centering
\subfloat[]{
    \includegraphics[width=0.15\textwidth]{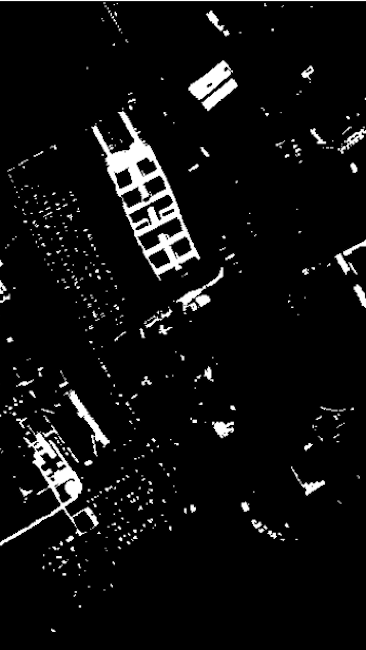}
}
\hspace{-3mm}
\subfloat[]{
    \includegraphics[width=0.15\textwidth]{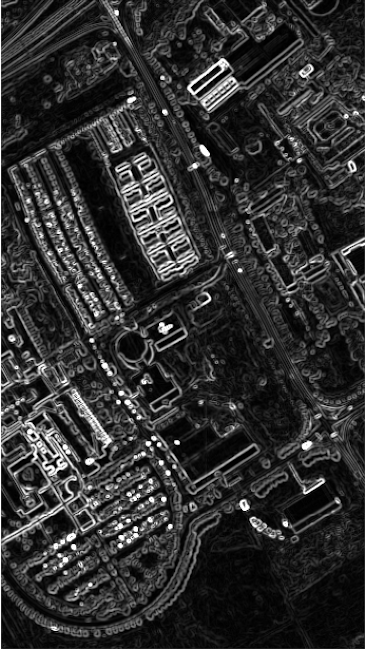}
}
\hspace{-3mm}
\subfloat[]{
    \includegraphics[width=0.15\textwidth]{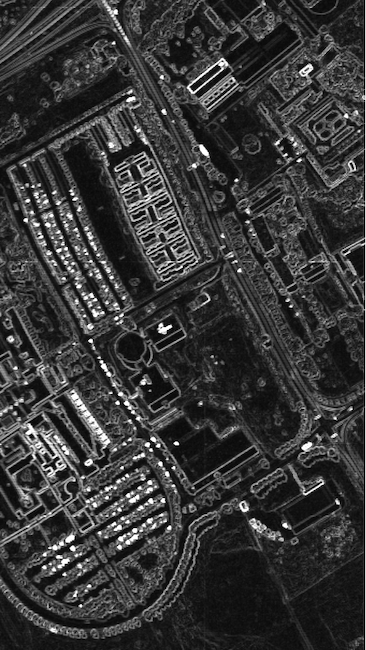}
}
\caption{Comparison of different edge extraction strategies: (a) without edge extraction, (b) the proposed edge extraction strategy based on Sobel filtering of the spectral mean image, and (c) band-wise spectral gradient aggregation.}
\label{fig:edge_extraction_comparison}
\end{figure}

\textbf{Edge extraction strategy analysis:}
To verify our Sobel-based edge extraction strategy, we compare three variants: no edge extraction, the adopted Sobel filtering on the spectral mean image, and band-wise spectral gradient aggregation. The band-wise spectral gradient aggregation is conceptually similar to~\cite{chu2022hyperspectral}, which emphasizes computing gradient information across spectral bands rather than relying only on a single grayscale representation. As shown in (see Fig.~\ref{fig:edge_extraction_comparison}), the variant without edge extraction produces sparse and fragmented responses, making it difficult to delineate continuous object boundaries. In contrast, our Sobel-based strategy captures more complete structural boundaries, confirming the necessity of explicit edge modeling in EASLP. Moreover, although the Sobel-based and band-wise spectral gradient edge maps exhibit minor local differences, they remain highly consistent in terms of overall boundary distribution. The mean absolute difference (MAD) between the two edge maps is only 0.0229, suggesting that the spectral-mean-based Sobel operator effectively preserves the dominant boundary information required by EASLP. These results support the effectiveness and suitability of the adopted spectral-mean-based Sobel operator for hyperspectral image classification.

\subsection{Ablation experiments}

Three ablation experiments are designed to validate the effectiveness of the modules. As shown in Table~\ref{tab:ablation}, results across four datasets demonstrate component contributions, with peak performances marked in bold.

\textbf{EASLP}: Removing the EASLP module (w/o EASLP) causes noticeable performance degradation across all datasets. The most pronounced drop occurs on PaviaU (OA: 95.21\% $\rightarrow$ 92.66\%), followed by 1--2\% decreases on Houston2013 and variable declines on KSC/Botswana. This consistent degradation confirms the critical role of spatial structural constraints, revealing the insufficiency of spectral features alone.

\textbf{DHP}: Ablating the DHP module (w/o DHP) causes marginal but consistent degradation in OA/AA/Kappa. While its quantitative improvement is less dramatic than other components, it fundamentally improves training stability by leveraging historical information to suppress pseudo-label fluctuations and ensure smoother convergence. Although the performance improvement brought by this module is less pronounced than that of the previous two modules, it significantly enhanced the robustness of training. With the incorporation of historical information, the temporal fluctuation of pseudo-labels is reduced, leading to more stable model convergence. This capability proves especially valuable in few-shot or high-noise scenarios by effectively mitigating error accumulation. 

Since the KSC and Botswana datasets with ten labeled samples per class have already reached near-saturation, we additionally conducted ablation experiments on these two datasets using five labeled samples per class \cite{Le2021AlertTrap}. As shown in Table~\ref{tab:ablation_fivelabel}, results show that our proposed method achieves more noticeable performance improvements under this more challenging low-label setting, further confirming the effectiveness of the DHP and EASLP modules.

% \textcolor{red}{It is worth noting that on near-saturated datasets such as KSC and Botswana, where the baseline performance is already extremely high, the marginal contribution of DHP becomes less apparent when evaluated solely by OA and Kappa. Nevertheless, its effect on training robustness and pseudo-label consistency remains beneficial, suggesting that more challenging evaluation settings may be required to fully reveal its advantages.}

\textbf{ATSC}: We remove the ATSC module (w/o ATSC), relying solely on confidence scores from fused historical and current predictions. A fixed threshold of 0.95 is applied for binary categorization of all unlabeled samples. Results show quantifiable performance deterioration across all four datasets--approximately 1\% OA/AA drops on PaviaU and Houston2013, with KSC/Botswana showing smaller declines but increased standard deviation, indicating pseudo-label instability. The Count-Gap mechanism proves essential for consistency-based sample stratification and robust pseudo-labeling.

\begin{table}[htbp]
\centering
\caption{Ablation results on KSC and Botswana datasets using five labeled samples per class.}
\label{tab:ablation_fivelabel}

\begin{tabular}{ccccc}
\toprule
Dataset & Metrics & w/o EASLP & w/o DHP & Ours \\
\midrule

\multirow{3}{*}{KSC}
& OA (\%) & 98.06 & 98.48 & 99.05 \\
& AA (\%) & 96.86 & 97.88 & 98.16 \\
& $\kappa \times 100$ & 97.83 & 98.31 & 98.94 \\

\midrule

\multirow{3}{*}{Botswana}
& OA (\%) & 96.85 & 97.99 & 98.59 \\
& AA (\%) & 97.26 & 97.28 & 98.67 \\
& $\kappa \times 100$ & 96.59 & 97.82 & 98.56 \\

\bottomrule
\end{tabular}

\end{table}

%\clearpage
\section{CONCLUSION}\label{conclusion}

% The proposed semi-supervised framework for hyperspectral image classification seamlessly integrates spatial prior information with dynamic learning mechanism, achieving substantial gains in feature extraction and classification performance. Confronting the dual challenges of high annotation costs and limited sample availability, the incorporated EASLP module successfully alleviates label diffusion originating from superpixel segmentation, consequently strengthening classification robustness in boundary regions. Furthermore, the DHP method enhances temporal consistency and noise robustness by smoothing pseydo-label fluctuations. The ATSC strategy improves pseudo-label quality and learning efficiency through hierarchical sample utilization. By synergistically integrating DHP and ATSC, the proposed DREPL framework achieves spatiotemporal synchronization, thereby boosting classification performance.

The proposed semi-supervised hyperspectral classification framework integrates edge-aware modeling with a dynamically reliable pseudo-labeling strategy, effectively addressing two key issues: label diffusion and pseudo-label instability. To mitigate label diffusion caused by superpixel segmentation, the EASLP module introduces boundary constraints into the similarity matrix and adaptively adjusts the contribution of neighboring pixels based on boundary strength, thereby suppressing the spread of labels into incorrect regions.During the pseudo-label consistency optimization stage, the DREPL strategy dynamically combines historical predictions with current predictions to improve consistency. At the same time, by incorporating confidence and the Count-Gap metric, unlabeled samples are automatically categorized into easy, ambiguous, and hard groups, with different optimization strategies applied to each. In this way, pseudo-label stability is enhanced jointly from both spatial and temporal perspectives.

%Results validate the efficacy of our framework in few-shot HSI classification with robust performance. 
The results validate the effectiveness of our framework in few-shot hyperspectral image classification, achieving an average improvement of 7.00\% in Overall Accuracy (OA), 5.60\% in Average Accuracy (AA), and 8.00\% in Kappa on five benchmark datasets. However, the current study is limited to the commonly adopted single-scene setting, and the cross-scene and cross-sensor generalization capability of the proposed framework remains unexplored. Future work will focus on improving cross-domain transferability and robustness for multi-scene hyperspectral image classification.
% However, the proposed method is inherently constrained by single-scene applicability and unaddressed domain adaptation needs. Subsequent investigations should prioritize cross-domain generalization and transferabiliy enhancements for multi-scene applications.

\section*{Declarations}
% All manuscripts must contain the following sections under the heading 'Declarations'.

%If any of the sections are not relevant to your manuscript, please include the heading and write 'Not applicable' for that section.

\paragraph{Acknowledgements} 
This work was supported by the Basic Scientific Research Project of the Education Department of Liaoning Province (Key Project of Independent Topic Selection, 2024), Grant No. 524053218.

% \paragraph*{Funding}
% %information that explains whether and by whom the research was supported
% Text text text text text text text text text text text text text text text text text text text text text text text text text text text text text text text text text text text text text text.

% \paragraph*{Conflicts of interest/Competing interests}
% %include appropriate disclosures
% Text text text text text text text text text text text text text text text text text text text text text text text text text text text text text text text text text text text text text text.

\paragraph{Availability of data and material} 
%data transparency
All hyperspectral datasets used in this paper are public standard open datasets, which are not original data produced by the authors. 

\paragraph{Code availability} 
%software application or custom code
The source code used in this study is not publicly available. Researchers with legitimate academic research purposes may obtain the code from the corresponding author upon reasonable request for non-commercial academic use.

\paragraph{Authors' contributions} 
%optional: please review the submission guidelines from the journal whether statements are mandatory
Conceptualization, Q.M.and Q.F.; Methodology, Q.M.; Software, Q.M.and T.L.; Validation, T.L.and S.Z.; Formal analysis, Q.M.; Investigation, Q.M.and T.L.; Resources, L.F.and W.Y.; Writing – original draft, Q.M.; Writing – review \& editing, F.Q.and T.L.; Visualization, Q.M.; Supervision, L.F.and W.Y.; Project administration, L.F.and W.Y.; Funding acquisition, Q.F.. All authors have read and agreed to the published version of the manuscript.

% Put your bibliogrpahy settings here
\bibliographystyle{spbasic}      % basic style, author-year citations
\bibliography{bibliography}   % name your BibTeX data base
\end{document}